\documentclass[10pt,twocolumn,letterpaper]{article}

\usepackage{cvpr}              
\definecolor{cvprblue}{rgb}{0.21,0.49,0.74}
\usepackage{algorithm}
\usepackage{algpseudocode}
\usepackage{amsfonts}
\usepackage{amssymb}
\usepackage{bm}
\usepackage{ulem}
\usepackage{float}
\usepackage{array}
\usepackage{enumitem}

\usepackage{amsmath}
\usepackage{graphicx}
\usepackage{multirow}
\usepackage{subcaption}
\usepackage[pagebackref]{hyperref}

\newcommand{\bd}{\bm{d}}
\newcommand{\bg}{\bm{g}}

\newcommand{\bW}{\bm{W}}
\newcommand{\bx}{\bm{x}}

\newcommand{\bnu}{\bm{\nu}}

\newcommand{\by}{\bm{y}}
\newcommand{\bI}{\bm{I}}
\newcommand{\bA}{\bm{A}}

\newcommand{\bJ}{\bm{J}}

\newcommand{\calA}{\mathcal{A}}

\newcommand{\calL}{\mathcal{L}}

\newcommand{\calN}{\mathcal{N}}

\newcommand{\calW}{\mathcal{W}}

\newcommand{\rmd}{\mathrm{d}}
\newcommand{\rmT}{\mathrm{T}}

\newcommand{\bbR}{\mathbb{R}}

\newcommand{\btheta}{\bm{\theta}}

\def\paperID{NULL} 
\def\confName{CVPR}
\def\confYear{2026}

\title{Outlier-Robust Diffusion Solvers for Inverse Problems}

\author{Yang Zheng$^{1}$ \quad Jiahua Liu$^{1}$ \\ Tongyao Pang$^{2}$\quad Wen Li$^{1}$ \quad Zhaoqiang Liu$^{1,}$\thanks{Corresponding author.}\\
$^{1}$School of Computer Science and Engineering, University of Electronic Science and Technology of China \\ $^{2}$Yau Mathematical Sciences Center, Tsinghua University\\
{\tt\small 202511081645@std.uestc.edu.cn} \quad {\tt\small 202421081114@std.uestc.edu.cn} \\{\tt\small typang@tsinghua.edu.cn} \quad {\tt\small liwenbnu@gmail.com} \quad {\tt\small zqliu12@gmail.com}}

\begin{document}
\maketitle
\begin{abstract}
Methods based on diffusion models (DMs) for solving inverse problems (IPs) have recently achieved remarkable performance. However, DM-based methods typically struggle against outliers, which are common in real-world measurements. In this work, to tackle IPs with outliers, we first refine the measurement via explicit noise estimation to mitigate the effect of noise. Subsequently, we formulate an iteratively reweighted least squares objective based on the Huber loss to address the outliers. We propose a method utilizing gradient descent to approximately solve the corresponding optimization problem for the robust objective. To avoid delicate tuning of the learning rate required by the gradient descent method, we further employ the conjugate gradient method with an efficient strategy for updating. Extensive experiments on multiple image datasets for linear and nonlinear tasks under various conditions demonstrate that our proposed methods exhibit robustness to outliers and outperform recent DM-based methods in most cases. The code is available at \url{https://github.com/StarNextDay/Robust-DAPS.git}.
\end{abstract}

\section{Introduction}
Inverse problems (IPs) encompass a broad class of tasks focused on estimating underlying signals from degraded and noisy observations. IPs are crucial in numerous domains, including audio signal processing~\cite{saito2023unsupervised, moliner2023solving}, remote sensing~\cite{twomey2019introduction}, and image restoration~\cite{candes2005decoding, wu2019deep}. The typical objective of an IP is to recover an unknown signal $\bm{x}_{0}^{\ast} \in \mathbb{R}^{n}$ from a noisy observation $\bm{y} \in \mathbb{R}^{m}$, modeled as \cite{Fou13, saharia2022palette}:
\begin{equation}
\label{eq:y=ax_n}
    \by = \mathcal{A}(\bm{x}_{0}^{\ast}) + \bm{\nu},
\end{equation}
where $\mathcal{A}: \mathbb{R}^{n} \to \mathbb{R}^{m}$ denotes a forward operator and $\bm{\nu} \in \mathbb{R}^{m}$ represents unknown stochastic noise.

Recent advances in diffusion models (DMs) have catalyzed the development of a wide range of DM-based approaches for solving IPs. These methods can be broadly classified into two paradigms. The first paradigm comprises methods tailored to specific IPs, such as super-resolution~\cite{gao2023implicit, shang2024resdiff}, inpainting~\cite{corneanu2024latentpaint, lugmayr2022repaint}, and deblurring~\cite{ren2023multiscale, sanghvi2025kernel}, which are trained in a supervised fashion using paired datasets of degraded observations and corresponding ground-truth signals. While these methods often produce high-quality reconstructions, their dependence on task-specific training and large volumes of labeled data introduces computational overhead and limits their generalizability. The second paradigm circumvents these limitations by leveraging pretrained DMs to solve IPs without requiring additional fine-tuning~\cite{chung2022mr, song2024solving, song2021solving, chung2023parallel, kawar2022ddrm, wang2022zero, song2023pigdm, dou2024diffusion, wang2024dmplug, murata2023gibbsddrm}. These methods implicitly assume that the target signal is close to the generative manifold learned by a pretrained DM and exploit the learned prior of the DM to facilitate signal reconstruction. The methods of the second paradigm have recently gained popularity as they do not require retraining for specific tasks and avoid the need for paired datasets.

However, recent DM-based methods typically only consider how to address the noise contained in the observation, but neglect the effect of outliers. In practical scenarios, the observation is often affected not only by stochastic Gaussian noise but also by outliers, such as those encountered in image deblurring tasks~\cite{pan2016robust, dong2017blind, chen2020oid, dong2021deep}, and other IPs~\cite{jalal2021robust, dalalyan2019outlier, song2024robust}. To model these outliers, we employ the arbitrary corruption model proposed in~\cite{jalal2021robust, song2024robust}. This model stipulates that a fraction $\rho \in [0, 1)$ of the observation are arbitrarily replaced by an outlier vector $\bm{\xi} \in \bbR^{m}$, such that:
\begin{equation}
y_i = \begin{cases}
\xi_{i}, & \text{if } i \in \mathcal{C} \\
\left(\calA(\bx^{\ast}_{0})\right)_{i} + \nu_{i}, & \text{if } i \notin \mathcal{C}
\end{cases},
\end{equation}
where $\mathcal{C}$ is the unknown set of corrupted indices and its elements are randomly selected from the indices of the measurement $\by$ with the probability $\rho$. $\xi_{i}$ denotes the $i$-th entry of the outlier vector. Motivated by the limitations of recent DM-based methods on IPs with outliers, this work proposes robust DM-based methods for IPs with outliers.

\subsection{Related work}
We discuss related work in two parts: 1) Inverse problems with diffusion models and 2) methods for outlier handling.

\textbf{Inverse problems with diffusion models:} Since the pioneering advancements in DMs~\cite{sohl2015deep, ho2020denoising, song2021score}, extensive research has aimed to enhance both the computational efficiency and the quality of generated outputs~\cite{song2020denoising, lu2022dpm, karras2022elucidating, zhao2024unipc}. Furthermore, DMs have emerged as powerful tools for addressing IPs~\cite{chung2023dps, mardani2023variational, song2023pigdm, wang2024dmplug, li2024decoupled, moufad2024variational, dou2025hybrid, zheng2025integrating}, achieving remarkable performance across diverse applications.

DM-based methods for IPs reconstruct underlying signals through different frameworks. For instance, MCG~\cite{chung2022mcg} and DPS~\cite{chung2023dps} employ Tweedie's formula to estimate the reconstructed signal, supplementing this with gradient-based posterior corrections. DiffPIR~\cite{zhu2023denoising} adopts a proximal update scheme to approximate the conditional posterior mean. DCPS~\cite{janati2024divide} introduces divide-and-conquer posterior sampling to leverage the inherent structure of DMs. RED-diff~\cite{mardani2023variational} leverages variational inference by introducing a tractable surrogate distribution to approximate the real posterior. DAPS~\cite{zhang2024daps} decouples the diffusion sampling trajectory to enable early-stage error correction.

Although recent DM-based approaches yield promising results for IPs under Gaussian noise, these methods often struggle to accurately recover the underlying signal when the measurements contain outliers, highlighting the need for robust DM-based methods against IPs with outliers.

\textbf{Methods for outlier handling: } Outliers are common in real-world scenarios due to faulty sensor readings or brief interference during data transmission~\cite{jalal2021robust, song2024robust, yang2026outlier}. The central idea behind methods for outlier handling lies in constructing a robust loss function to mitigate the influence of the outliers. To achieve this mitigation, various robust methods have been proposed.

One class of methods achieves robustness through a sample selection strategy, such as trimmed loss (TL), which explicitly discards the data points that contribute the highest loss~\cite{liu2019high, liu2020high, shen2019learning}. Another class involves robust estimators, such as the median-of-means (MOM) method, which leverages the resistance of median to extreme values~\cite{jalal2021robust, laforgue2021generalization}. MOM estimates the expected risk by partitioning the data into blocks and taking the median of the block means, thereby filtering out groups that are heavily biased by outliers. However, both TL and MOM methods discard the information of data points flagged as potential outliers, thus failing to fully utilize the information in the measurement data. Alternatively, the Huber loss achieves robustness by leveraging all measurement information and differentially penalizing the data points~\cite{dalalyan2019outlier, song2024robust, d2021consistent}. Specifically, the Huber loss achieves robustness by applying a quadratic penalty to data points with small discrepancies and a linear penalty to those with large discrepancies, thus limiting the loss contribution of outliers without discarding any measurement information.

Additionally, methods for outlier handling typically incorporate prior knowledge regarding the underlying signal. This is often achieved either through manually designed regularization terms~\cite{cho2011handling, loh2015regularized, pan2016robust, song2024robust} or by leveraging generative models, such as generative adversarial models~\cite{jalal2021robust, wang2020further}. Given the strong generative capabilities of DMs, in this paper, we utilize pre-trained DMs coupled with methods for outlier handling to solve IPs contaminated with outliers.
\subsection{Contributions}
Our main contributions are summarized as follows:
\begin{itemize}
\item We first introduce an optimization objective employing the Huber loss coupled with explicit noise estimation to mitigate the influence of outliers. We approximately solve the corresponding optimization problem via gradient descent, and the approach is referred to as Robust-GD.
\item To circumvent the need for delicate learning rate tuning in Robust-GD, we substitute gradient descent with the conjugate gradient method and introduce an efficient update strategy tailored for this optimization.The corresponding approach is referred to as Robust-CG.
\item We conduct extensive empirical evaluations across diverse image datasets and under various conditions. The results show that our algorithms achieve superior performance compared to recent DM-based methods in most cases, confirming the effectiveness of our methods.
\end{itemize}

\begin{figure*}[!htbp]
    \centering
    \includegraphics[width=0.9\linewidth]{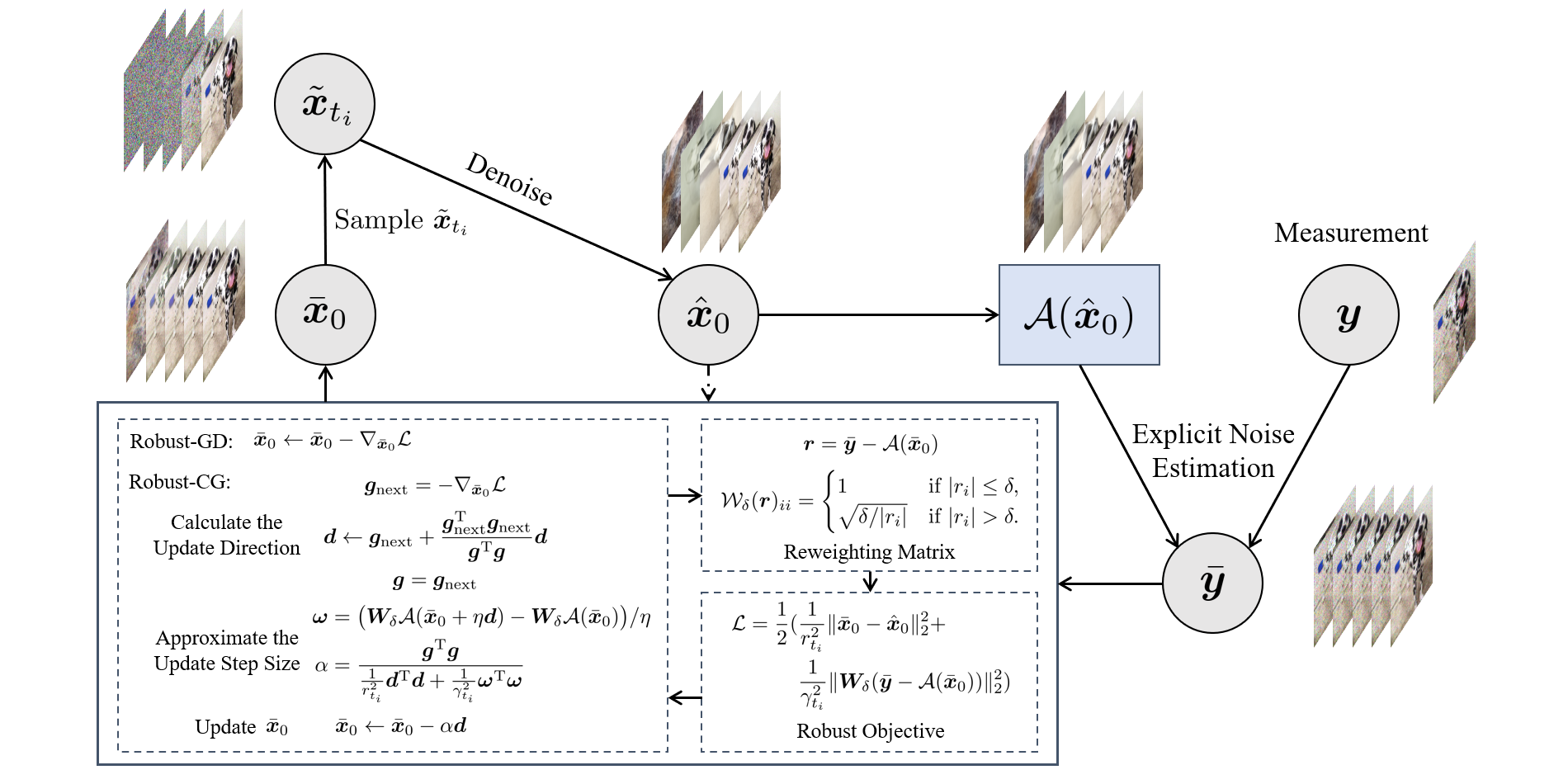}
    \vspace{-1em}
    \caption{%
        Overview of our proposed Robust-GD and Robust-CG methods. At each timestep $t_{i}$, we first calculate the signal estimate $\hat{\bx}_{0}$. Next, we refine the measurement $\by$ to $\bar{\by}$ using explicit noise estimation. We then formulate a robust objective function designed to handle outliers and approximately solve the corresponding optimization problem using either the gradient descent method (Robust-GD) or the conjugate gradient method (Robust-CG). After obtaining the result $\bar{\bx}_{0}$, we sample $\tilde{\bx}_{t_{i-1}} \sim \mathcal{N}(\alpha_{t_{i-1}}\bar{\bx}_{0}, \sigma_{t_{i}}^{2}\bI_{n})$. After $N$ iterations, we use $\tilde{\bx}_{t_{0}}$ as the final reconstructed image.
    }
    \label{fig:pipeline}
    \vspace{-1em}
\end{figure*}

\section{Preliminaries}
DM comprises a forward stochastic process that gradually corrupts data into noise and a corresponding reverse process that aims to reconstruct the data from the noise. The forward dynamics are described by the following stochastic differential equation (SDE):
\begin{equation}\label{eq:forward_process}
    \rmd \bm{x}_{t} = f(t)\,\bm{x}_{t}\,\rmd t + g(t)\,\rmd \bm{w}_t,
    \quad \bm{x}_0 \sim p_0,
\end{equation}
where $f(t)$ and $g(t)$ are time-dependent drift and diffusion coefficients, and $\bm{w}_t \in \bbR^{n}$ is a standard Wiener process. Let $p_t$ be the marginal distribution of $\bm{x}_t$. For $t \in [0,T]$, the conditional distribution of $\bm{x}_t|\bm{x}_0$ is Gaussian with $\bm{x}_t|\bm{x}_{0} \sim \mathcal{N}(\alpha_t \bm{x}_0, \sigma_t^2 \bI_{n})$, where $\alpha_t$ and $\sigma_t$ are differentiable, non-negative and monotonic functions with bounded derivatives, such that the signal-to-noise ratio $\alpha_t^2/\sigma_t^2$ is decreasing in $t$. Functions $\alpha_t$ and $\sigma_t$ are chosen that the marginal distribution at the terminal time $p_{T}$ approximates $\mathcal{N}(\bm{0}, \tilde{\sigma}^2 \bI_{n})$ for some $\tilde{\sigma} > 0$.

To match the SDE in Eq.~\eqref{eq:forward_process} with the conditional distribution of $\bx_{t}|\bx_{0}$ specified above, the coefficients $f(t)$ and $g(t)$ need to satisfy the following~\cite{lu2022dpm}:
\begin{equation}
\label{eq:relation_f_alpha}
f(t) = \frac{\rmd \log \alpha_t}{\rmd t},
\quad
g^2(t) = \frac{\rmd \sigma_t^2}{\rmd t} - 2\,\frac{\rmd \log \alpha_t}{\rmd t}\,\sigma_t^2.
\end{equation}

Following~\cite{song2021score}, the forward SDE admits a corresponding time-reversed diffusion process as follows:
\begin{equation}
\label{eq:reverse_SDE}
\rmd \bm{x}_t = 
\bigl(f(t)\,\bm{x}_t - g^2(t)\,\nabla_{\bm{x}_{t}} \log p_t(\bm{x}_t)\bigr)\,\rmd t
+ g(t)\,\rmd \bar{\bm{w}}_t,
\end{equation}
where $\bar{\bm{w}}_t$ denotes a reverse-time Wiener process and $\nabla_{\bm{x_{t}}} \log p_t$ is the score function.

As shown in~\cite{song2021score}, a corresponding deterministic process with the same marginal distribution is defined by the following ordinary differential equation (ODE):
\begin{equation}
\label{eq:reverse_ODE}
\rmd \bm{x}_t = 
\bigl(f(t)\,\bm{x}_t - \tfrac{1}{2} g^2(t)\,\nabla_{\bm{x}_{t}} \log p_t(\bm{x}_t)\bigr)\,\rmd t.
\end{equation}
Sampling can thus be performed via numerical solvers of this ODE, given that the unknown score function $\nabla_{\bx_{t}}\log p_{t}$ is approximated by a certain neural network function. Concretely, the score function $\nabla_{\bx_{t}} \log p_{t}(\bx_{t})$ can be approximated by $(\alpha_{t} \bx_{\btheta}(\bx_{t}, t)-\bx_t )/\sigma_{t}^{2}$~\cite{kingma2021variational,lu2022dpmp}, where $\bx_{\btheta}(\bx_{t}, t)$ is a data prediction network.  

Let the time interval $[0, T]$ be partitioned as $0 = t_{0} < t_{1} < \cdots  < t_{N} = T$. Then, for instance, the DDIM~\cite{song2020denoising} sampling scheme can be regarded as a first-order numerical solution to Eq.~\eqref{eq:reverse_ODE}, and can be expressed as follows:
\begin{equation}
    \tilde{\bx}_{t_{i-1}} = \frac{\sigma_{t_{i-1}}}{\sigma_{t_{i}}} \tilde{\bx}_{t_{i}} + \sigma_{t_{i-1}} \left( \frac{\alpha_{t_{i-1}}}{\sigma_{t_{i-1}}} - \frac{\alpha_{t_{i}}}{\sigma_{t_{i}}} \right) \bx_{\btheta}(\tilde{\bx}_{t_{i}}, t_{i}).
\end{equation}

\section{Methods}
This section introduces our proposed methods, referred to as Robust-GD and Robust-CG, two novel DM-based approaches for solving IPs with outliers. In Section~\ref{sec:noise_estimation}, we present the concept of explicit noise estimation to mitigate the effect of noise in the measurement. In Section~\ref{sec:huber_loss}, we formulate an iteratively reweighted least squares objective based on the Huber loss to address the outliers. We approximately solve the corresponding optimization problem via gradient descent, which we summarize as the Robust-GD algorithm. To avoid the delicate fine-tuning of the learning rate required by Robust-GD, in Section~\ref{sec:conjugate}, we replace the gradient descent method with the conjugate gradient method. We also propose an efficient updating strategy, summarizing the complete algorithm as Robust-CG. We visualize the entire sampling process of our proposed Robust-GD and Robust-CG methods in Figure~\ref{fig:pipeline}.

\subsection{Explicit noise estimation}
\label{sec:noise_estimation}
Following previous work~\cite{zhu2023denoising, zhang2024daps}, the estimation of the underlying signal at time step $t$ using DMs is typically formulated as the following optimization problem:
\begin{equation}
\min_{\bar{\bx}_{0}}\frac{1}{2r_{t}^{2}} \|\bar{\bx}_{0} - \hat{\bx}_{0}(\tilde{\bx}_{t}, t)\|_{2}^{2} + \lambda \|\by - \mathcal{A}(\bar{\bx}_{0})\|_{2}^{2},
\end{equation}
where $\hat{\bx}_{0}(\tilde{\bx}_{t}, t)$ estimates the underlying signal given the noisy latent variable $\tilde{\bx}_{t}$. Here, $r_{t} > 0$ is a time-dependent hyperparameter, and $\lambda > 0$ is a hyperparameter related to the measurement noise. Various DM-based approaches utilize different strategies for estimating the underlying signal. For instance, DiffPIR uses Tweedie's formula to obtain the estimate $\hat{\bx}_{0}(\tilde{\bx}_{t}, t) = \bx_{\btheta}(\tilde{\bx}_{t}, t)$, whereas DAPS obtains $\hat{\bm{x}}_{0}(\tilde{\bm{x}}_{t}, t)$ through a multi-step reverse process starting from time $t$. In this paper, our approach is similar to DAPS, as we also obtain $\hat{\bm{x}}_{0}(\tilde{\bm{x}}_{t}, t)$ through a multi-step reverse process starting from time $t$.

Recent DM-based methods typically perform well when the measurements are only slightly perturbed, for example, under Gaussian noise with a low noise level. However, the work \cite{chen2025robust} suggests that the second squared $\ell_2$ fidelity term, $\|\by - \mathcal{A}(\bar{\bx}_{0})\|_{2}^{2}$, may be problematic when the measurement $\by$ is heavily contaminated, as the large discrepancy between the noisy measurement $\by$ and the noiseless measurement $\mathcal{A}(\bx^{\ast}_{0})$ can destroy the reconstruction process. Hence, we consider refining the measurement $\by$. We first mitigate the effect of noise. Instead of training an additional model to generate a pseudo condition as in~\cite{chen2025robust}, we explicitly estimate the additive noise $\bnu$ to refine the measurement. Assuming the additive noise follows $\calN(\bm{0}, \sigma^{2}\bI_{m})$, we formulate the following optimization problem:
\begin{equation}
\label{eq:tdne}
\min_{\bar{\bx}_{0}, \bar{\bnu}}\frac{1}{2r_{t}^{2}} \|\bar{\bx}_{0} - \hat{\bx}_{0}(\tilde{\bx}_{t}, t)\|_{2}^{2} + \frac{1}{2\sigma^{2}}\|\bar{\bnu}\|_{2}^{2} + \frac{1}{2\gamma_{t}^{2}} \|\by - \mathcal{A}(\bar{\bx}_{0}) - \bar{\bnu}\|_{2}^{2},
\end{equation}
where $\gamma_{t}$ is a time-dependent hyperparameter that balances the guidance strength of the fidelity term $\|\by - \mathcal{A}(\bar{\bx}_{0}) - \bar{\bnu}\|_{2}^{2}$. As suggested in DAPS~\cite{zhang2024daps}, the sensitivity of the estimation $\hat{\bx}_{0}(\tilde{\bx}_{t}, t)$ to its noisy latent input $\tilde{\bx}_t$ is time-dependent. Since $\tilde{\bx}_t$ is derived from the result of the optimization problem in the previous step, noise accumulated in the intermediate estimate $\tilde{\bx}_0$ can affect the estimation $\hat{\bx}_{0}(\tilde{\bx}_{t}, t)$ in subsequent steps. Specifically, $\hat{\bx}_{0}(\tilde{\bx}_{t}, t)$ is robust to perturbations in $\tilde{\bx}_t$ during early timesteps, allowing for stronger guidance from the measurement $\bar{\by}$ (i.e., a smaller $\gamma_t$) and tolerance of potential noise. Conversely, in later timesteps, $\hat{\bx}_{0}(\tilde{\bx}_{t}, t)$ becomes more sensitive, necessitating a larger hyperparameter $\gamma_t$ to ensure that the estimated $\bar{\bx}_0$ remains consistent with the estimation $\hat{\bx}_{0}(\tilde{\bx}_{t}, t)$, thereby preserving the naturalness of the restoration. We set the hyperparameter $\gamma_{t} = 1/\sigma_{t}$ empirically, where $1/\sigma_{t}$ satisfies the requirement of having a low value in early timesteps and a high value in later timesteps.

We then proceed to solve the optimization problem in Eq.~\eqref{eq:tdne} iteratively. Since we aim to refine the measurement, we first estimate the additive noise as:
\begin{align}
\tilde{\bnu} &= \arg\min_{\bar{\bnu}}\frac{1}{2\sigma^{2}}\|\bar{\bnu}\|_{2}^{2} + \frac{1}{2\gamma_{t}^{2}} \|\by - \mathcal{A}(\hat{\bx}_{0}(\tilde{\bx}_{t}, t)) - \bar{\bnu}\|_{2}^{2}\\
\label{eq:tdne_noise}
&= \frac{\sigma^{2}}{\gamma_{t}^{2}+\sigma^{2}}\left(\by - \mathcal{A}(\hat{\bx}_{0}(\tilde{\bx}_{t}, t))\right).
\end{align}
Substituting Eq.~\eqref{eq:tdne_noise} into Eq.~\eqref{eq:tdne} to refine the measurement $\by$, the optimization problem for $\bar{\bx}_{0}$ becomes:
\begin{equation}
\min_{\bar{\bx}_{0}}\frac{1}{2}\left(\frac{1}{r_{t}^{2}} \|\bar{\bx}_{0} - \hat{\bx}_{0}(\tilde{\bx}_{t}, t)\|_{2}^{2} + \frac{1}{\gamma_{t}^{2}} \|\bar{\by} - \mathcal{A}(\bar{\bx}_{0})\|_{2}^{2}\right),
\end{equation}
where $\bar{\by} = \frac{1}{\gamma_{t}^{2}+\sigma^{2}}(\gamma_{t}^{2}\by + \sigma^{2}\calA(\hat{\bx}_{0}(\tilde{\bx}_{t}, t)))$.

\subsection{Robust objective for outliers}
\label{sec:huber_loss}
To further address IPs with outliers, we follow~\cite{song2024robust} to replace the squared $\ell_2$ fidelity term, $\|\bar{\by} - \mathcal{A}(\bar{\bx}_{0})\|_{2}^{2}$, with a sum of element-wise Huber loss functions. For the threshold $\delta > 0$, the univariate Huber loss operator $\mathcal{H}_{\delta}(\cdot)$ is defined as:
\begin{equation}
\mathcal{H}_{\delta}(r) = \begin{cases} r^{2}, & \text{if } |r| \le \delta, \\ 2\delta |r| - \delta^{2}, & \text{if } |r| > \delta. \end{cases}
\end{equation}
The robust data fidelity term is applied to the residual vector $\bm{r} = \bar{\by} - \mathcal{A}(\bar{\bx}_{0})$ and is defined by summing the Huber loss over all entries:
\begin{equation}
\mathcal{H}_{\delta}(\bar{\by} - \mathcal{A}(\bar{\bx}_{0})) = \sum_{i=1}^{m} \mathcal{H}_{\delta} ((\bar{\by} - \mathcal{A}(\bar{\bx}_{0}))_{i}).
\end{equation}
We follow~\cite{kummerle2021iteratively} and express the Huber loss term $\mathcal{H}_{\delta}(\bar{\by} - \mathcal{A}(\bar{\bx}_{0}))$ in a quadratic form, $\tilde{\mathcal{H}}_{\delta}(\bar{\by} - \mathcal{A}(\bar{\bx}_{0}))$. This new form shares the same gradient with respect to $\bar{\bx}_{0}$ as the original term and enables the use of an iteratively reweighted least squares scheme:
\begin{equation}
\tilde{\mathcal{H}}_{\delta}(\bar{\by} - \mathcal{A}(\bar{\bx}_{0})) = \|\mathcal{W}_{\delta}(\bar{\by}-\mathcal{A}(\bar{\bx}_{0}))(\bar{\by} - \mathcal{A}(\bar{\bx}_{0}))\|^{2}_{2},
\end{equation}
where $\mathcal{W}_{\delta}(\cdot)$ is a diagonal operator defined as:
\begin{equation}
\label{eq:W_sigma}
\mathcal{W}_{\delta}(\bm{r})_{ii} = \begin{cases} 1, & \text{if } |r_{i}| \le \delta, \\ \sqrt{\delta/|r_{i}|}, & \text{if } |r_{i}| > \delta. \end{cases}
\end{equation}
Note that although $\mathcal{W}_{\delta}(\bm{r})$ is computed using the current estimate $\bar{\bx}_{0}$, it is detached from the gradient computation with respect to $\bar{\bx}_{0}$ to ensure $\nabla_{\bar{\bx}_{0}}\mathcal{H}_{\delta}(\bar{\by} - \mathcal{A}(\bar{\bx}_{0}))=\nabla_{\bar{\bx}_{0}}\tilde{\mathcal{H}}_{\delta}(\bar{\by} - \mathcal{A}(\bar{\bx}_{0}))$. The modified optimization objective incorporates the Huber loss to achieve robustness against outliers and is given by:
\begin{equation}
\label{eq:huber_loss}
\min_{\bar{\bx}_{0}}\frac{1}{2} \left(\frac{1}{r_{t}^{2}}\|\bar{\bx}_{0} - \hat{\bx}_{0}(\tilde{\bx}_{t}, t)\|_{2}^{2} + \frac{1}{\gamma_{t}^{2}}\|\bm{W}_{\delta}(\bar{\by} - \mathcal{A}(\bar{\bx}_{0}))\|^{2}_{2}\right),
\end{equation}
where we use $\bm{W}_{\delta} \in \bbR^{m \times m}$ to denote $\mathcal{W}_{\delta}(\bar{\by}-\mathcal{A}(\bar{\bx}_{0}))$ for brevity. We can approximately solve the optimization problem in Eq.~\eqref{eq:huber_loss} via gradient descent, initialized at $\tilde{\bx}_{0}$ and using an empirically set learning rate $\eta_{x}$:\footnote{Experiments with varying learning rate $\eta_x$ for Robust-GD revealed that the performance of Robust-GD is sensitive to the selection of the learning rate. The detailed results are presented in the supplementary material.}
\begin{align}
\label{eq:loss_function}
\bar{\bx}_{0}^{(j+1)} = \bar{\bx}_{0}^{(j)} - \frac{\eta_{x}}{2}\nabla_{\bar{\bx}_{0}^{(j)}}\big(&\frac{1}{r_{t}^{2}}\|\bar{\bx}_{0}^{(j)} - \hat{\bx}_{0}(\tilde{\bx}_{t}, t)\|_{2}^{2} + \\ 
\nonumber &\frac{1}{\gamma_{t}^{2}}\|\bm{W}_{\delta}^{(j)}(\bar{\by} - \mathcal{A}(\bar{\bx}_{0}^{(j)}))\|^{2}_{2}\big).
\end{align}
The corresponding algorithm is presented in Algorithm~\ref{alg:robust-gd}.
\begin{algorithm}[!htbp]
\begin{small}
\begin{algorithmic}[1]
\Require Data prediction network $\bx_{\btheta}(\cdot,\cdot)$, measurement vector $\by$, forward operator $\calA(\cdot)$, noise schedule $\{\sigma_{t_{i}}\}_{i = 0}^{N}$, $\{\alpha_{t_{i}}\}_{i = 0}^{N}$, time dependent parameters $\{r_{t_{i}}\}_{i=0}^{N}$, $\{\gamma_{t_{i}}\}_{i=0}^{N}$, Gaussian noise level $\sigma$, learning rate $\eta_{x}$, total number of sample steps and iterations $N$, $J$, the Huber loss threshold parameter $\delta$
\State Sample $\tilde{\bx}_{t_{N}} \sim \calN(\bm{0}, \bI_{n})$
\For{$i = N, \dots, 1$}
\State $\bar{\bx}_{0}^{(0)} = \hat{\bx}_{0} = \hat{\bx}_{0}(\tilde{\bx}_{t_{i}}, t_{i})$
\State $\bar{\by} = \frac{1}{\gamma_{t_{i}}^{2}+\sigma^{2}}(\gamma_{t_{i}}^{2}\by + \sigma^{2}\calA(\hat{\bx}_{0}))$
\For{$j = 0, \dots, J-1$}
\State $\bW_{\delta}^{(j)} = \calW_{\delta}(\bar{\by}- \calA(\bar{\bx}_{0}^{(j)}))$
\State $\calL = \frac{1}{2}\big(\frac{1}{r_{t_{i}}^{2}}\|\bar{\bx}_{0}^{(j)} - \hat{\bx}_{0}\|_{2}^{2} + \frac{1}{\gamma_{t_{i}}^{2}} \|\bW_{\delta}^{(j)}(\bar{\by} - \calA(\bar{\bx}^{(j)}_{0}))\|^{2}_{2}\big)$
\State $\bar{\bx}_{0}^{(j+1)} = \bar{\bx}_{0}^{(j)} - \eta_{x}\nabla_{\bar{\bx}_{0}^{(j)}} \mathcal{L}$
\EndFor
\State Sample $\tilde{\bx}_{t_{i-1}} \sim \calN(\alpha_{t_{i-1}}\bar{\bx}_{0}^{(J)}, \sigma_{t_{i-1}}^{2}\bI_{n})$
\EndFor
\State \Return $\tilde{\bx}_{t_{0}}$
\end{algorithmic}
\end{small}
\caption{Robust Diffusion Solver using Gradient Descent Method (Robust-GD)}
\label{alg:robust-gd}
\end{algorithm}

\subsection{Solve via conjugate gradient method}
\label{sec:conjugate}
Using gradient descent to solve Eq.~\eqref{eq:huber_loss} typically requires delicate tuning of the learning rate. We instead employ the conjugate gradient (CG) method to mitigate the issue. At the $j$-th iteration, given the current gradient $\bg_{j}$, the update direction $\bd_{j}$, and the current result $\bar{\bx}_{0}^{(j)}$, the next step is to calculate the update step size $\alpha_{j}$. This step is formulated as the line search problem:
\begin{align}
\label{eq:alpha_loss}
\alpha_{j} = \arg\min_{\alpha}
\frac{1}{2}\big(&\frac{1}{r_{t}^{2}}\|(\bar{\bx}_{0}^{(j)} + \alpha\bd_{j}) - \hat{\bx}_{0}(\tilde{\bx}_{t}, t)\|_{2}^{2} + \\
\nonumber &\frac{1}{\gamma_{t}^{2}}\|\bm{W}_{\delta}^{(j)}(\bar{\by} - \mathcal{A}(\bar{\bx}_{0}^{(j)} + \alpha\bd_{j}))\|^{2}_{2}\big).
\end{align}
When the forward operator is linear, specifically, the forward operator is $\mathcal{A}(\bx) = \bA\bx$, where $\bA \in \mathbb{R}^{m\times n}$ is the measurement matrix, the objective function is quadratic, which allows us to efficiently obtain the closed-form optimal update step size $\alpha_{j}$ as:
\begin{equation}
\alpha_{j} = \frac{\bg_{j}^{\rmT}\bd_{j}}{\frac{1}{r_{t}^{2}}\bd_{j}^{\rmT}\bd_{j} + \frac{1}{\gamma_{t}^{2}}(\bW_{\delta}^{(j)}\bA\bd_{j})^{\rmT}(\bW_{\delta}^{(j)}\bA\bd_{j})}.
\end{equation}
Furthermore, we adapt the $\bg_{j}^{\rm T}\bg_{j}$ form for the numerator, which is commonly used in non-linear conjugate gradient methods and is known to be effective in line searches~\cite{nocedal2006numerical}. Empirically, this choice also shows superior performance compared to using the standard $\bg_{j}^{\rmT}\bd_{j}$ term,\footnote{Empirical results demonstrate that Robust-CG utilizing the conjugate condition $\bg_{j}^{\rmT}\bg_{j}$ achieves superior performance compared to the version employing $\bg_{j}^{\rmT}\bd_{j}$. Detailed experimental results are presented in the supplementary material.} yielding the closed-form expression for the optimal step size $\alpha_j$:
\begin{equation}
\label{eq:cg_gg}
\alpha_{j} = \frac{\bg_{j}^{\rmT}\bg_{j}}{\frac{1}{r_{t}^{2}}\bd_{j}^{\rmT}\bd_{j} + \frac{1}{\gamma_{t}^{2}}(\bW_{\delta}^{(j)}\bA\bd_{j})^{\rmT}(\bW_{\delta}^{(j)}\bA\bd_{j})}.
\end{equation}
For the nonlinear IP, we cannot directly calculate the optimal step size. We first linearize the nonlinear operator $\mathcal{A}(\bar{\bx}_{0}^{(j)} + \alpha \bd_{j})$ via a first-order Taylor expansion around the current estimate $\bar{\bx}_0^{(j)}$:
\begin{equation}
\mathcal{A}(\bar{\bx}_{0}^{(j)} + \alpha\bd_{j}) \approx \mathcal{A}(\bar{\bx}_{0}^{(j)}) + \alpha\bJ(\bar{\bx}_0^{(j)})\bd_{j},
\end{equation}
where $\bJ(\bar{\bx}_{0}^{(j)})$ is the Jacobian matrix of $\mathcal{A}(\cdot)$ evaluated at $\bar{\bx}_{0}^{(j)}$. By substituting this linear approximation, the optimal step size is obtained in the same form as the linear case:
\begin{equation}
\alpha_{j} = \frac{\bg_{j}^{\rmT}\bg_{j}}{\frac{1}{r_{t}^{2}}\bd_{j}^{\rmT}\bd_{j} + \frac{1}{\gamma_{t}^{2}}(\bW_{\delta}^{(j)}\bJ(\bar{\bx}_0^{(j)})\bd_{j})^{\rmT}(\bW_{\delta}^{(j)}\bJ(\bar{\bx}_0^{(j)})\bd_{j})}.
\end{equation}
However, calculating $\bJ(\bar{\bx}_0^{(j)})$ explicitly is also computationally expensive. To avoid explicitly computing the Jacobian matrix $\bJ(\bar{\bx}_0^{(j)})$, we utilize a finite difference approximation for the Jacobian-vector product $\bJ(\bar{\bx}_0^{(j)})\bd_{j}$~\cite{martens2010deep, kelley1995iterative}:
\begin{equation}
\bJ(\bar{\bx}_0^{(j)})\bd_{j}\approx (\mathcal{A}(\bar{\bx}_0^{(j)} + \eta\bd_{j}) - \mathcal{A}(\bar{\bx}_0^{(j)}))/{\eta},
\end{equation}
where $\eta$ is a small, experimentally determined hyperparameter. By substituting this approximation into the derivative of Eq.~\eqref{eq:alpha_loss} and setting it to zero, we obtain the calculation for $\alpha_{j}$ under nonlinear scenarios:
\begin{align}
\bm{\omega}_{j} &= \left(\bW_{\delta}^{(j)}\mathcal{A}(\bar{\bx}_{0}^{(j)} +\eta\bd_{j}) - \bW_{\delta}^{(j)}\mathcal{A}(\bar{\bx}_{0}^{(j)})\right) /\eta,\\
\alpha_{j} &= \frac{\bg_{j}^{\rmT}\bg_{j}}{\frac{1}{r_{t}^{2}}\bd_{j}^{\rmT}\bd_{j} + \frac{1}{\gamma_{t}^{2}}\bm{\omega}_{j}^{\rmT}\bm{\omega}_{j}}.
\end{align}
To ensure the conjugacy property for the search direction, we employ the Fletcher-Reeves formula~\cite{jiang2019improved}. The complete Robust-CG algorithm is detailed in Algorithm~\ref{alg:robust}.

\begin{algorithm}[!htbp]
\begin{small}
\begin{algorithmic}[1]
\Require Data prediction network $\bx_{\btheta}(\cdot,\cdot)$, measurement vector $\by$, forward operator $\calA(\cdot)$, noise schedule $\{\sigma_{t_{i}}\}_{i = 0}^{N}$, $\{\alpha_{t_{i}}\}_{i = 0}^{N}$, time dependent parameters $\{r_{t_{i}}\}_{i=0}^{N}$, $\{\gamma_{t_{i}}\}_{i=0}^{N}$, Gaussian noise level $\sigma$, finite-difference approximation parameter $\eta$, total number of sample steps and iterations $N$, $J$, the Huber loss threshold parameter $\delta$
\State Sample $\tilde{\bx}_{t_{N}} \sim \calN(\bm{0}, \bI_{n})$
\For{$i = N, \dots, 1$}
\State $\bar{\bx}_{0}^{(0)} =\hat{\bx}_{0} = \hat{\bx}_{0}(\tilde{\bx}_{t_{i}}, t_{i})$
\State $\bar{\by} = \frac{1}{\gamma_{t_{i}}^{2}+\sigma^{2}}(\gamma_{t_{i}}^{2}\by + \sigma^{2}\calA(\hat{\bx}_{0}))$
\State $\bW_{\delta}^{(0)} = \calW_{\delta}(\bar{\by}- \calA(\bar{\bx}_{0}^{(0)}))$
\State $\calL = \frac{1}{2}\big(\frac{1}{r_{t_{i}}^{2}}\|\bar{\bx}_{0}^{(0)} - \hat{\bx}_{0}\|_{2}^{2} + \frac{1}{\gamma_{t_{i}}^{2}} \|\bW_{\delta}^{(0)}(\bar{\by} - \calA(\bar{\bx}^{(0)}_{0}))\|^{2}_{2}\big)$
\State $\bd_{0} = \bg_{0} = -\nabla_{\bar{\bx}^{(0)}_{0}}\calL$
\For{$j = 0, \dots, J-1$}
\State $\bm{\omega}_{j} = \big(\bW_{\delta}^{(j)}\mathcal{A}(\bar{\bx}_{0}^{(j)} +\eta\bd_{j}) - \bW_{\delta}^{(j)}\mathcal{A}(\bar{\bx}_{0}^{(j)})\big) /\eta$
\State $\alpha_{j} = (\bg_{j}^{\rmT}\bg_{j})/\big(\frac{1}{r_{t_{i}}^{2}}\bd_{j}^{\rmT}\bd_{j} + \frac{1}{\gamma_{t_{i}}^{2}}\bm{\omega}_{j}^{\rmT}\bm{\omega}_{j}\big)$
\State $\bar{\bx}_{0}^{(j+1)} = \bar{\bx}_{0}^{(j)} + \alpha_{j}\bd_{j}$
\State $\bW_{\delta}^{(j+1)} = \calW_{\delta}(\bar{\by}- \calA(\bar{\bx}_{0}^{(j+1)}))$
\State $\calL = \frac{1}{2}(\frac{1}{r_{t_{i}}^{2}}\|\bar{\bx}_{0}^{(j+1)} - \hat{\bx}_{0}\|_{2}^{2} + \frac{1}{\gamma_{t_{i}}^{2}} \|\bW_{\delta}^{(j+1)}(\bar{\by} - \calA(\bar{\bx}^{(j+1)}_{0}))\|^{2}_{2})$
\State $\bg_{j+1} = -\nabla_{\bar{\bx}_{0}^{(j+1)}}\calL$
\State $\bd_{j+1} = \bg_{j+1} + \frac{\bg_{j+1}^{\rmT}\bg_{j+1}}{\bg_{j}^{\rmT}\bg_{j}}\bd_{j}$
\EndFor
\State Sample $\tilde{\bx}_{t_{i-1}} \sim \calN(\alpha_{t_{i-1}}\bar{\bx}_{0}^{(J)}, \sigma_{t_{i-1}}^{2}\bI_{n})$
\EndFor
\State \Return $\tilde{\bx}_{t_{0}}$
\end{algorithmic}
\end{small}
\caption{Robust Diffusion Solver using Conjugate Gradient Method (Robust-CG)}
\label{alg:robust}
\end{algorithm}

\section{Experimental results}
\label{sec:exp}
We evaluate the two proposed Robust-GD and Robust-CG methods on three $256 \times 256$ validation datasets, namely CelebA~\cite{liu2015faceattributes}, FFHQ~\cite{karras2019style} and ImageNet~\cite{deng2009imagenet}.\footnote{Due to the page limit, the experimental results on FFHQ are presented in the supplementary material.} From each dataset, we randomly sample 100 validation images for image reconstruction, following the experimental setting adopted in recent DM-based methods for IPs~\cite{zhu2023denoising, zhang2024daps, wang2024dmplug}. To assess reconstruction quality and demonstrate that our methods balance the perception-distortion tradeoff~\cite{blau2018perception}, we report both distortion and perceptual metrics. Specifically, we use the peak signal-to-noise ratio (PSNR) and structural similarity index (SSIM) to evaluate distortion, and learned perceptual image patch similarity (LPIPS)~\cite{zhang2018unreasonable} to measure perceptual quality. We also report the Fr\'echet Inception Distance (FID)~\cite{heusel2017gans} for the main experiments to quantify perceptual quality. We use \textbf{bold} to indicate the best performance, \underline{underline} for the second-best.

For linear IPs, we compare our methods with baseline DM-based methods, including DPS~\cite{chung2023dps}, DiffPIR~\cite{zhu2023denoising}, DCPS~\cite{janati2024divide}, RED-diff~\cite{mardani2023variational}, and DAPS~\cite{zhang2024daps}. For nonlinear deblurring tasks, we compare our methods with DPS, RED-Diff, and DAPS. Experiments on all main tasks are conducted under Gaussian noise with the noise level $\sigma = 0.05$. To validate the robustness of our methods under different contamination levels, we set the contamination factor $\rho = 0.02$ or $0.10$ following~\cite{jalal2021robust, song2024robust}. For all tasks, we set the value of all entries of the outlier vector $\xi_{i}$ to $-1$, which is consistent with the measurement boundary $\by \in [-1, 1]$. All experiments are run on a single NVIDIA GeForce RTX 4090 GPU.

\subsection{Inpainting and super-resolution}
\label{sec:exp_ip_sp}
\begin{figure}[!htbp]
    \centering
    \vspace{-1.0em}
    \includegraphics[width=0.875\linewidth]{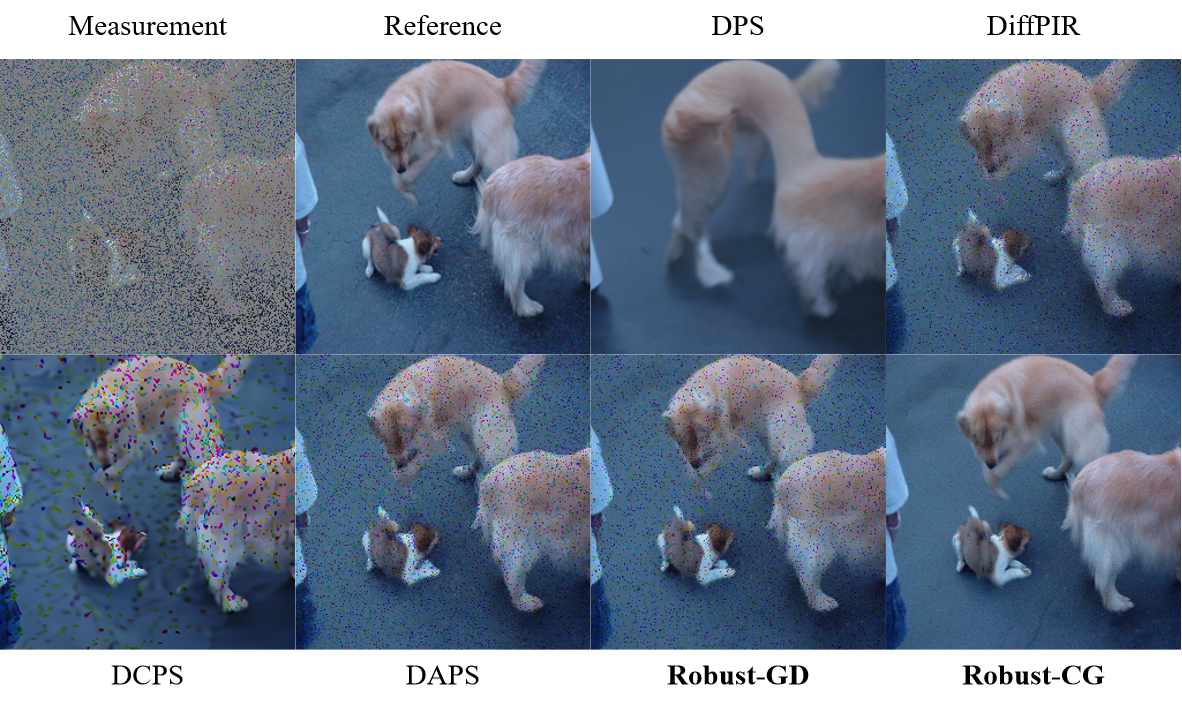}
    \vspace{-1.0em}
    \caption{%
    Visualization results of our methods and other DM-based approaches for the inpainting task, with Gaussian noise ($\sigma = 0.05$) and a contamination fraction of $\rho = 0.10$.
    }
    \vspace{-1.0em}
    \label{fig:exp_ip}
\end{figure}

\begin{figure}[!htbp]
    \centering
    \vspace{-1.0em}
    \includegraphics[width=0.875\linewidth]{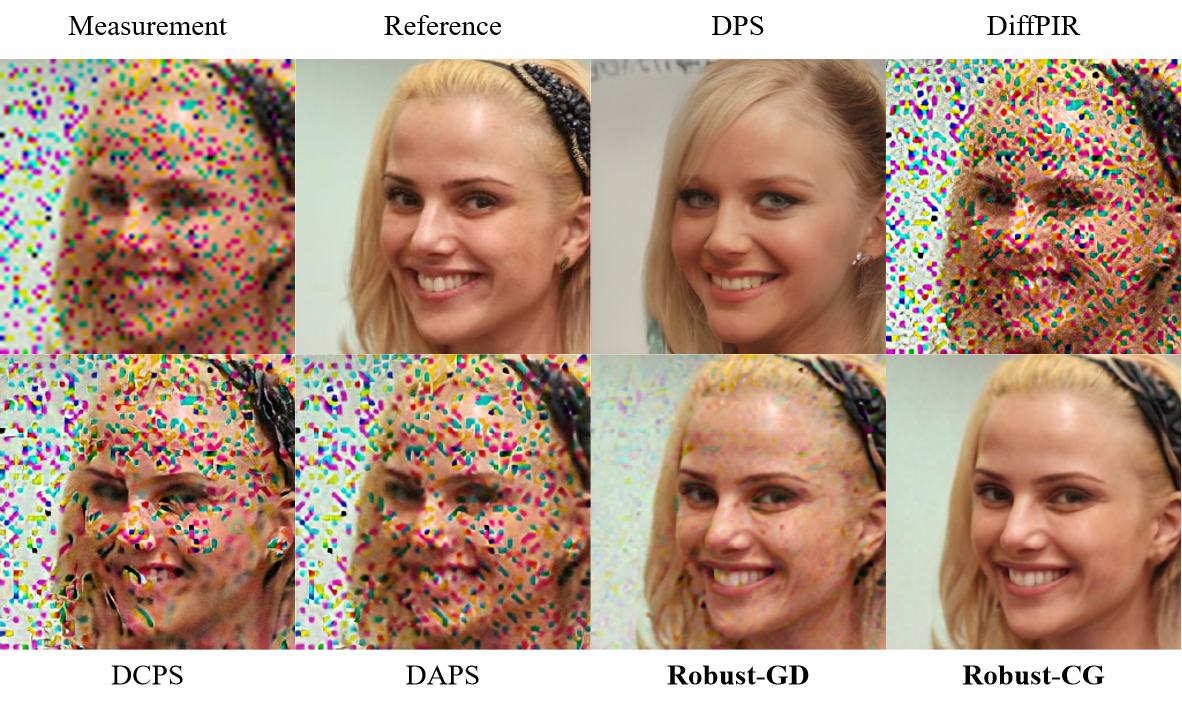}
    \vspace{-1.0em}
    \caption{%
    Visualization results of our methods and other DM-based approaches for the super-resolution task, with Gaussian noise ($\sigma = 0.05$) and a contamination fraction of $\rho = 0.10$.
    }
    \vspace{-1.0em}
    \label{fig:exp_sp}
\end{figure}

For linear IPs, we first evaluate the performance of our methods on two tasks, namely the inpainting (random $70\%$ masking) and super-resolution ($4\times$ downscaling) tasks. The settings for both tasks follow those established in previous works~\cite{chung2023dps, zhang2024daps, wang2024dmplug}. For the inpainting task, the threshold for Huber loss $\delta$ is set to $0.01$ in Robust-GD and $0.02$ in Robust-CG. For the super-resolution task, $\delta$ is set to $0.02$ in Robust-GD and $0.005$ in Robust-CG. The finite difference approximation parameter $\eta$ in Robust-CG is set to $10^{-3}$ for both tasks.\footnote{More detailed setup for the two proposed algorithms is presented in the supplementary material.} The results presented in Table~\ref{tab:ip_sp} demonstrate that Robust-CG performs the best across almost all metrics. Robust-GD also performs well compared to other recent DM-based methods and exhibits robustness against outliers. We visualize the experimental results for the inpainting and super-resolution tasks in Figures~\ref{fig:exp_ip} and~\ref{fig:exp_sp}, respectively. The results demonstrate that our approaches yield better performance against outliers and reconstruct images that are visually more similar to the reference image.

\begin{table*}[ht]
\begin{center}
\renewcommand{\arraystretch}{0.9}
\vspace{-1.5em}
\resizebox{\textwidth}{!}{
\setlength{\tabcolsep}{0.8mm}{
\begin{tabular}{c c c c c c c c c c c c c c c c c c}
\hline
\multicolumn{2}{c}{}
&\multicolumn{4}{c}{\scriptsize{\textbf{CelebA ($256 \times 256$)}}}
&\multicolumn{4}{c}{\scriptsize{\textbf{ImageNet ($256 \times 256$)}}}
&\multicolumn{4}{c}{\scriptsize{\textbf{CelebA ($256 \times 256$)}}}
&\multicolumn{4}{c}{\scriptsize{\textbf{ImageNet ($256 \times 256$)}}}
\\
\cmidrule(lr){3-6}
\cmidrule(lr){7-10}
\cmidrule(lr){11-14}
\cmidrule(lr){15-18}
\multicolumn{2}{c}{}
&\scriptsize{PSNR$\uparrow$}
&\scriptsize{SSIM$\uparrow$}
&\scriptsize{LPIPS$\downarrow$}
&\scriptsize{FID$\downarrow$}
&\scriptsize{PSNR$\uparrow$}
&\scriptsize{SSIM$\uparrow$}
&\scriptsize{LPIPS$\downarrow$}
&\scriptsize{FID$\downarrow$}
&\scriptsize{PSNR$\uparrow$}
&\scriptsize{SSIM$\uparrow$}
&\scriptsize{LPIPS$\downarrow$}
&\scriptsize{FID$\downarrow$}
&\scriptsize{PSNR$\uparrow$}
&\scriptsize{SSIM$\uparrow$}
&\scriptsize{LPIPS$\downarrow$}
&\scriptsize{FID$\downarrow$}
\\
\hline
\scriptsize{$\rho$}
&\scriptsize{Methods}
&\multicolumn{8}{c}{\scriptsize{\textbf{Super-resolution ($4 \times$)}}}
&\multicolumn{8}{c}{\scriptsize{\textbf{Inpainting (random $70\%$)}}}
\\
\cline{1-18}
\multirow{7}{*}{\scriptsize{0.02}}
&\scriptsize{DPS}
&\scriptsize{23.99 \tiny{$\pm$ 1.40}}%
&\scriptsize{0.655 \tiny{$\pm$ 0.075}}%
&\scriptsize{0.184 \tiny{$\pm$ 0.057}}%
&\scriptsize{\underline{66.96}}
&\scriptsize{20.86 \tiny{$\pm$ 3.48}}%
&\scriptsize{0.495 \tiny{$\pm$ 0.189}}%
&\scriptsize{\underline{0.447} \tiny{$\pm$ 0.154}}%
&\scriptsize{181.15}
&\scriptsize{26.68 \tiny{$\pm$ 1.32}}%
&\scriptsize{0.748 \tiny{$\pm$ 0.058}}%
&\scriptsize{0.148 \tiny{$\pm$ 0.048}}%
&\scriptsize{63.49}
&\scriptsize{23.49 \tiny{$\pm$ 3.48}}%
&\scriptsize{0.620 \tiny{$\pm$ 0.171}}%
&\scriptsize{0.361 \tiny{$\pm$ 0.155}}%
&\scriptsize{144.56}
\\
&\scriptsize{DiffPIR}
&\scriptsize{20.75 \tiny{$\pm$ 1.68}}%
&\scriptsize{0.561 \tiny{$\pm$ 0.049}}%
&\scriptsize{0.486 \tiny{$\pm$ 0.073}}%
&\scriptsize{239.75}
&\scriptsize{18.66 \tiny{$\pm$ 2.24}}%
&\scriptsize{0.389 \tiny{$\pm$ 0.087}}%
&\scriptsize{0.591 \tiny{$\pm$ 0.094}}%
&\scriptsize{248.36}
&\scriptsize{25.83 \tiny{$\pm$ 1.40}}%
&\scriptsize{0.695 \tiny{$\pm$ 0.039}}%
&\scriptsize{0.316 \tiny{$\pm$ 0.088}}%
&\scriptsize{150.54}
&\scriptsize{23.49 \tiny{$\pm$ 2.65}}%
&\scriptsize{0.651 \tiny{$\pm$ 0.064}}%
&\scriptsize{0.325 \tiny{$\pm$ 0.104}}%
&\scriptsize{133.38}
\\
&\scriptsize{DCPS}
&\scriptsize{20.64 \tiny{$\pm$ 2.09}}%
&\scriptsize{0.591 \tiny{$\pm$ 0.081}}%
&\scriptsize{0.370 \tiny{$\pm$ 0.103}}%
&\scriptsize{193.03}
&\scriptsize{18.00 \tiny{$\pm$ 2.71}}%
&\scriptsize{0.395 \tiny{$\pm$ 0.144}}%
&\scriptsize{0.613 \tiny{$\pm$ 0.143}}%
&\scriptsize{238.36}
&\scriptsize{\underline{30.70} \tiny{$\pm$ 1.37}}%
&\scriptsize{\textbf{0.865} \tiny{$\pm$ 0.031}}%
&\scriptsize{0.103 \tiny{$\pm$ 0.037}}%
&\scriptsize{68.34}
&\scriptsize{23.63 \tiny{$\pm$ 4.11}}%
&\scriptsize{\underline{0.727} \tiny{$\pm$ 0.110}}%
&\scriptsize{0.244 \tiny{$\pm$ 0.102}}%
&\scriptsize{117.91}
\\
&\scriptsize{RED-diff} 
&\scriptsize{22.23 \tiny{$\pm$ 1.76}}%
&\scriptsize{0.679 \tiny{$\pm$ 0.028}}%
&\scriptsize{0.472 \tiny{$\pm$ 0.067}}%
&\scriptsize{192.54}
&\scriptsize{20.81 \tiny{$\pm$ 2.38}}%
&\scriptsize{0.594 \tiny{$\pm$ 0.077}}%
&\scriptsize{0.544 \tiny{$\pm$ 0.086}}%
&\scriptsize{229.77}
&\scriptsize{24.69 \tiny{$\pm$ 1.15}}%
&\scriptsize{0.656 \tiny{$\pm$ 0.030}}%
&\scriptsize{0.392 \tiny{$\pm$ 0.086}}%
&\scriptsize{152.93}
&\scriptsize{22.49 \tiny{$\pm$ 2.37}}%
&\scriptsize{0.599 \tiny{$\pm$ 0.047}}%
&\scriptsize{0.423 \tiny{$\pm$ 0.095}}%
&\scriptsize{185.72}
\\
&\scriptsize{DAPS} 
&\scriptsize{22.51 \tiny{$\pm$ 1.98}}%
&\scriptsize{0.685 \tiny{$\pm$ 0.048}}%
&\scriptsize{0.403 \tiny{$\pm$ 0.087}}%
&\scriptsize{236.29}
&\scriptsize{20.97 \tiny{$\pm$ 1.90}}%
&\scriptsize{\underline{0.633} \tiny{$\pm$ 0.062}}%
&\scriptsize{0.519 \tiny{$\pm$ 0.110}}%
&\scriptsize{241.92}
&\scriptsize{25.57 \tiny{$\pm$ 1.98}}%
&\scriptsize{0.691 \tiny{$\pm$ 0.048}}%
&\scriptsize{0.298 \tiny{$\pm$ 0.087}}%
&\scriptsize{195.00}
&\scriptsize{23.20 \tiny{$\pm$ 1.90}}%
&\scriptsize{0.645 \tiny{$\pm$ 0.062}}%
&\scriptsize{0.277 \tiny{$\pm$ 0.110}}%
&\scriptsize{182.83}
\\
&\scriptsize{\textbf{Robust-GD}} 
&\scriptsize{\underline{29.41} \tiny{$\pm$ 1.46}}%
&\scriptsize{\underline{0.803} \tiny{$\pm$ 0.035}}%
&\scriptsize{\underline{0.145} \tiny{$\pm$ 0.043}}%
&\scriptsize{72.29}
&\scriptsize{\underline{24.02} \tiny{$\pm$ 3.26}}%
&\scriptsize{0.623 \tiny{$\pm$ 0.088}}%
&\scriptsize{0.484 \tiny{$\pm$ 0.130}}%
&\scriptsize{\underline{165.10}}
&\scriptsize{30.47 \tiny{$\pm$ 1.18}}%
&\scriptsize{0.818 \tiny{$\pm$ 0.021}}%
&\scriptsize{\underline{0.089} \tiny{$\pm$ 0.027}}%
&\scriptsize{\underline{62.62}}
&\scriptsize{\underline{24.50} \tiny{$\pm$ 2.95}}%
&\scriptsize{0.710 \tiny{$\pm$ 0.071}}%
&\scriptsize{\underline{0.211} \tiny{$\pm$ 0.099}}%
&\scriptsize{\underline{83.10}}
\\
&\scriptsize{\textbf{Robust-CG}} 
&\scriptsize{\textbf{29.67} \tiny{$\pm$ 1.62}}%
&\scriptsize{\textbf{0.831} \tiny{$\pm$ 0.040}}%
&\scriptsize{\textbf{0.125} \tiny{$\pm$ 0.038}}%
&\scriptsize{\textbf{64.04}}
&\scriptsize{\textbf{24.90} \tiny{$\pm$ 4.02}}%
&\scriptsize{\textbf{0.697} \tiny{$\pm$ 0.129}}%
&\scriptsize{\textbf{0.375} \tiny{$\pm$ 0.133}}%
&\scriptsize{\textbf{146.05}}
&\scriptsize{\textbf{31.20} \tiny{$\pm$ 1.52}}%
&\scriptsize{\underline{0.855} \tiny{$\pm$ 0.023}}%
&\scriptsize{\textbf{0.071} \tiny{$\pm$ 0.021}}%
&\scriptsize{\textbf{51.42}}
&\scriptsize{\textbf{26.55} \tiny{$\pm$ 3.96}}%
&\scriptsize{\textbf{0.785} \tiny{$\pm$ 0.076}}%
&\scriptsize{\textbf{0.120} \tiny{$\pm$ 0.040}}%
&\scriptsize{\textbf{55.24}}
\\
\cline{1-18}
\multirow{7}{*}{\scriptsize{0.10}}
&\scriptsize{DPS}
&\scriptsize{20.95 \tiny{$\pm$ 1.40}}%
&\scriptsize{0.607 \tiny{$\pm$ 0.079}}%
&\scriptsize{\underline{0.217} \tiny{$\pm$ 0.061}}%
&\scriptsize{\underline{70.76}}
&\scriptsize{18.78 \tiny{$\pm$ 2.85}}%
&\scriptsize{0.450 \tiny{$\pm$ 0.188}}%
&\scriptsize{\underline{0.496} \tiny{$\pm$ 0.146}}%
&\scriptsize{\underline{192.57}}
&\scriptsize{22.29 \tiny{$\pm$ 1.28}}%
&\scriptsize{0.679 \tiny{$\pm$ 0.068}}%
&\scriptsize{0.185 \tiny{$\pm$ 0.055}}%
&\scriptsize{\underline{68.02}}
&\scriptsize{\underline{20.19} \tiny{$\pm$ 2.59}}%
&\scriptsize{\underline{0.538} \tiny{$\pm$ 0.179}}%
&\scriptsize{\underline{0.459} \tiny{$\pm$ 0.165}}%
&\scriptsize{177.36}
\\
&\scriptsize{DiffPIR}
&\scriptsize{14.76 \tiny{$\pm$ 1.71}}%
&\scriptsize{0.290 \tiny{$\pm$ 0.085}}%
&\scriptsize{0.671 \tiny{$\pm$ 0.081}}%
&\scriptsize{328.01}
&\scriptsize{13.40 \tiny{$\pm$ 2.56}}%
&\scriptsize{0.203 \tiny{$\pm$ 0.092}}%
&\scriptsize{0.769 \tiny{$\pm$ 0.104}}%
&\scriptsize{302.80}
&\scriptsize{19.72 \tiny{$\pm$ 1.80}}%
&\scriptsize{0.438 \tiny{$\pm$ 0.096}}%
&\scriptsize{0.599 \tiny{$\pm$ 0.142}}%
&\scriptsize{191.35}
&\scriptsize{18.90 \tiny{$\pm$ 2.39}}%
&\scriptsize{0.433 \tiny{$\pm$ 0.111}}%
&\scriptsize{0.561 \tiny{$\pm$ 0.168}}%
&\scriptsize{193.35}
\\
&\scriptsize{DCPS}
&\scriptsize{15.74 \tiny{$\pm$ 1.51}}%
&\scriptsize{0.371 \tiny{$\pm$ 0.103}}%
&\scriptsize{0.578 \tiny{$\pm$ 0.093}}%
&\scriptsize{317.65}
&\scriptsize{14.87 \tiny{$\pm$ 2.38}}%
&\scriptsize{0.324 \tiny{$\pm$ 0.101}}%
&\scriptsize{0.714 \tiny{$\pm$ 0.092}}%
&\scriptsize{284.75}
&\scriptsize{\underline{24.26} \tiny{$\pm$ 1.58}}%
&\scriptsize{\underline{0.807} \tiny{$\pm$ 0.039}}%
&\scriptsize{\underline{0.136} \tiny{$\pm$ 0.042}}%
&\scriptsize{78.22}
&\scriptsize{16.59 \tiny{$\pm$ 3.13}}%
&\scriptsize{0.411 \tiny{$\pm$ 0.156}}%
&\scriptsize{0.644 \tiny{$\pm$ 0.206}}%
&\scriptsize{212.46}
\\
&\scriptsize{RED-diff} 
&\scriptsize{16.10 \tiny{$\pm$ 1.92}}%
&\scriptsize{0.449 \tiny{$\pm$ 0.068}}%
&\scriptsize{0.639 \tiny{$\pm$ 0.083}}%
&\scriptsize{264.86}
&\scriptsize{18.96 \tiny{$\pm$ 2.31}}%
&\scriptsize{0.412 \tiny{$\pm$ 0.086}}%
&\scriptsize{0.610 \tiny{$\pm$ 0.147}}%
&\scriptsize{271.98}
&\scriptsize{19.86 \tiny{$\pm$ 1.67}}%
&\scriptsize{0.439 \tiny{$\pm$ 0.081}}%
&\scriptsize{0.623 \tiny{$\pm$ 0.134}}%
&\scriptsize{182.13}
&\scriptsize{15.99 \tiny{$\pm$ 2.54}}%
&\scriptsize{0.425 \tiny{$\pm$ 0.076}}%
&\scriptsize{0.672 \tiny{$\pm$ 0.088}}%
&\scriptsize{243.29}
\\
&\scriptsize{DAPS} 
&\scriptsize{16.63 \tiny{$\pm$ 2.45}}%
&\scriptsize{0.428 \tiny{$\pm$ 0.103}}%
&\scriptsize{0.582 \tiny{$\pm$ 0.101}}%
&\scriptsize{333.06}
&\scriptsize{15.84 \tiny{$\pm$ 2.17}}%
&\scriptsize{0.433 \tiny{$\pm$ 0.079}}%
&\scriptsize{0.660 \tiny{$\pm$ 0.114}}%
&\scriptsize{274.08}
&\scriptsize{19.15 \tiny{$\pm$ 2.45}}%
&\scriptsize{0.437 \tiny{$\pm$ 0.103}}%
&\scriptsize{0.596 \tiny{$\pm$ 0.101}}%
&\scriptsize{261.25}
&\scriptsize{18.02 \tiny{$\pm$ 2.17}}%
&\scriptsize{0.408 \tiny{$\pm$ 0.079}}%
&\scriptsize{0.567 \tiny{$\pm$ 0.114}}%
&\scriptsize{104.99}
\\
&\scriptsize{\textbf{Robust-GD}} 
&\scriptsize{\underline{26.52} \tiny{$\pm$ 1.27}}%
&\scriptsize{\underline{0.716} \tiny{$\pm$ 0.042}}%
&\scriptsize{0.264 \tiny{$\pm$ 0.068}}%
&\scriptsize{110.21}
&\scriptsize{\underline{20.48} \tiny{$\pm$ 2.17}}%
&\scriptsize{\underline{0.488} \tiny{$\pm$ 0.079}}%
&\scriptsize{0.649 \tiny{$\pm$ 0.114}}%
&\scriptsize{233.46}
&\scriptsize{23.28 \tiny{$\pm$ 0.87}}%
&\scriptsize{0.575 \tiny{$\pm$ 0.057}}%
&\scriptsize{0.374 \tiny{$\pm$ 0.086}}%
&\scriptsize{169.96}
&\scriptsize{19.02 \tiny{$\pm$ 2.49}}%
&\scriptsize{0.453 \tiny{$\pm$ 0.120}}%
&\scriptsize{0.491 \tiny{$\pm$ 0.185}}%
&\scriptsize{\underline{171.40}}
\\
&\scriptsize{\textbf{Robust-CG}} 
&\scriptsize{\textbf{28.96} \tiny{$\pm$ 1.56}}%
&\scriptsize{\textbf{0.819} \tiny{$\pm$ 0.041}}%
&\scriptsize{\textbf{0.129} \tiny{$\pm$ 0.039}}%
&\scriptsize{\textbf{67.41}}
&\scriptsize{\textbf{24.28} \tiny{$\pm$ 3.80}}%
&\scriptsize{\textbf{0.671} \tiny{$\pm$ 0.128}}%
&\scriptsize{\textbf{0.416} \tiny{$\pm$ 0.136}}%
&\scriptsize{\textbf{159.32}}
&\scriptsize{\textbf{29.74} \tiny{$\pm$ 1.29}}%
&\scriptsize{\textbf{0.809} \tiny{$\pm$ 0.022}}%
&\scriptsize{\textbf{0.093} \tiny{$\pm$ 0.029}}%
&\scriptsize{\textbf{58.48}}
&\scriptsize{\textbf{25.57} \tiny{$\pm$ 3.60}}%
&\scriptsize{\textbf{0.720} \tiny{$\pm$ 0.076}}%
&\scriptsize{\textbf{0.143} \tiny{$\pm$ 0.046}}%
&\scriptsize{\textbf{62.32}}
\\
\hline
\end{tabular}
}
}
\end{center}
\vspace{-1.5em}
\caption{(\textbf{Linear IPs}) \textbf{Super-resolution ($4\times$)}, \textbf{inpainting (random $70\%$)} with additive Gaussian noise ($\sigma = 0.05$) and contamination fraction $\rho = 0.02$ or $0.10$.}
\label{tab:ip_sp}
\vspace{-1.5em}
\end{table*}

\subsection{Linear image deblurring}
\label{sec:exp_linear_deblur}
\begin{figure}[htbp]
    \centering
    \vspace{-1.0em}
    \includegraphics[width=0.875\linewidth]{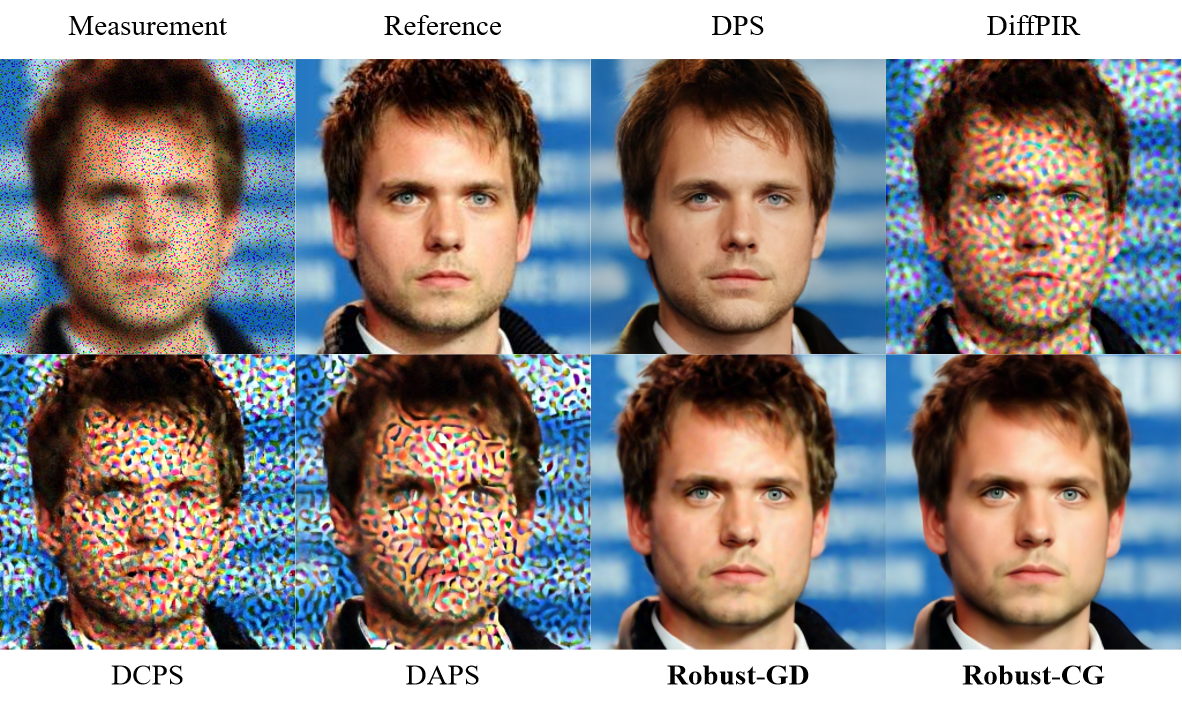}
    \vspace{-1.0em}
    \caption{%
    Visualization results of our methods and other DM-based approaches for the Gaussian deblurring task, with Gaussian noise ($\sigma = 0.05$) and a contamination fraction of $\rho = 0.10$.
    }
    \label{fig:exp_gaussian_deblur}
    \vspace{-1.0em}
\end{figure}

\begin{figure}[htbp]
    \centering
    \vspace{-1.0em}
    \includegraphics[width=0.875\linewidth]{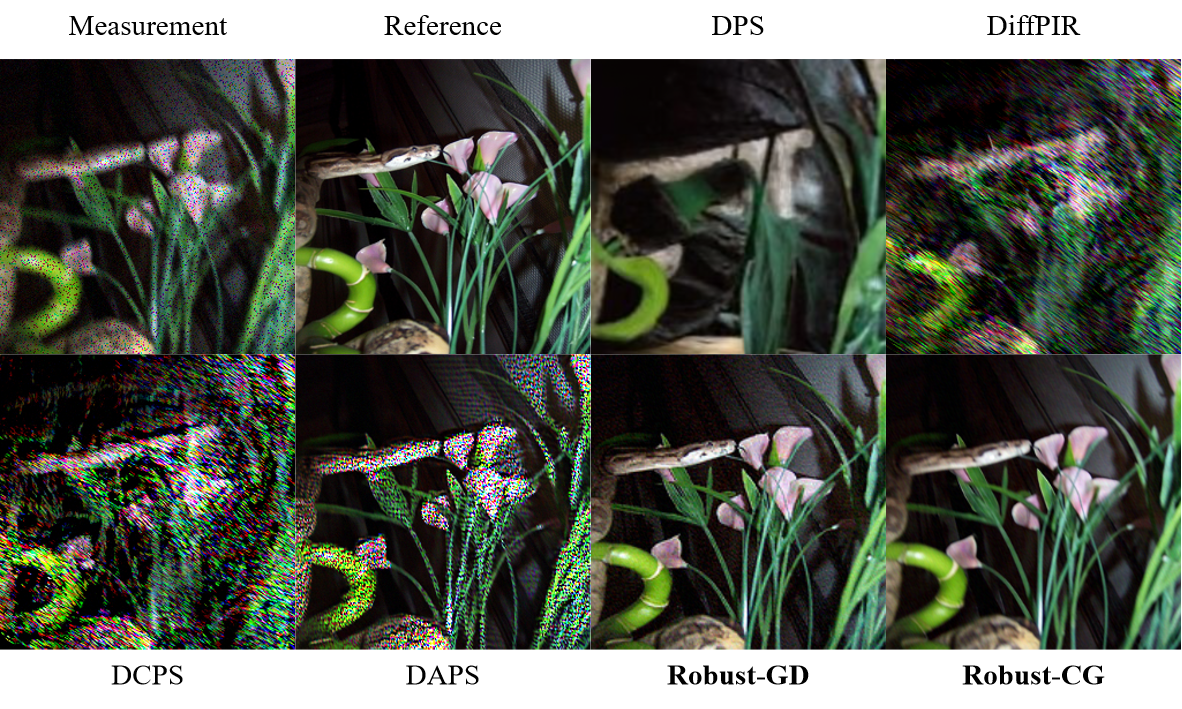}
    \vspace{-1.0em}
    \caption{%
    Visualization results of our methods and other DM-based approaches for the motion deblurring task, with Gaussian noise ($\sigma = 0.05$) and a contamination fraction of $\rho = 0.10$.
    }
    \label{fig:exp_motion_deblur}
    \vspace{-1.0em}
\end{figure}

We also evaluate the performance of our methods on two deblurring tasks for linear IPs, namely Gaussian deblurring and motion deblurring. For Gaussian and motion deblurring, we used kernels of size $61 \times 61$ with standard deviations of $3.0$ and $0.5$, respectively. All task settings follow those established in previous works~\cite{chung2023dps, zhang2024daps}. For both Robust-GD and Robust-CG, the threshold $\delta$ is set to $0.02$ and the finite difference approximation parameter $\eta$ in Robust-CG is set to $10^{-4}$. The results are presented in Table~\ref{tab:linear_deblur}, demonstrating that our methods achieve the best performance across almost all metrics. We also visualize the experimental results for the Gaussian deblurring and motion deblurring tasks in Figures~\ref{fig:exp_gaussian_deblur} and~\ref{fig:exp_motion_deblur}, respectively. Compared to other methods, the results demonstrate that our approaches yield better performance against outliers and produce images that retain more details and are visually more similar to the reference images.

\begin{table*}[ht]
\begin{center}
\vspace{-1.5em}
\renewcommand{\arraystretch}{0.9}
\resizebox{\textwidth}{!}{
\setlength{\tabcolsep}{0.85mm}{
\begin{tabular}{c c c c c c c c c c c c c c c c c c}
\hline
\multicolumn{2}{c}{}
&\multicolumn{4}{c}{\scriptsize{\textbf{CelebA ($256 \times 256$)}}}
&\multicolumn{4}{c}{\scriptsize{\textbf{ImageNet ($256 \times 256$)}}}
&\multicolumn{4}{c}{\scriptsize{\textbf{CelebA ($256 \times 256$)}}}
&\multicolumn{4}{c}{\scriptsize{\textbf{ImageNet ($256 \times 256$)}}}
\\
\cmidrule(lr){3-6}
\cmidrule(lr){7-10}
\cmidrule(lr){11-14}
\cmidrule(lr){15-18}
\multicolumn{2}{c}{}
&\scriptsize{PSNR$\uparrow$}
&\scriptsize{SSIM$\uparrow$}
&\scriptsize{LPIPS$\downarrow$}
&\scriptsize{FID$\downarrow$}
&\scriptsize{PSNR$\uparrow$}
&\scriptsize{SSIM$\uparrow$}
&\scriptsize{LPIPS$\downarrow$}
&\scriptsize{FID$\downarrow$}
&\scriptsize{PSNR$\uparrow$}
&\scriptsize{SSIM$\uparrow$}
&\scriptsize{LPIPS$\downarrow$}
&\scriptsize{FID$\downarrow$}
&\scriptsize{PSNR$\uparrow$}
&\scriptsize{SSIM$\uparrow$}
&\scriptsize{LPIPS$\downarrow$}
&\scriptsize{FID$\downarrow$}
\\
\hline
\scriptsize{$\rho$}
&\scriptsize{Methods}
&\multicolumn{8}{c}{\scriptsize{\textbf{Gaussian deblurring}}}
&\multicolumn{8}{c}{\scriptsize{\textbf{Motion deblurring}}}
\\
\cline{1-18}
\multirow{7}{*}{\scriptsize{0.02}}
&\scriptsize{DPS}
&\scriptsize{25.14 \tiny{$\pm$ 1.48}}%
&\scriptsize{0.680 \tiny{$\pm$ 0.074}}%
&\scriptsize{0.152 \tiny{$\pm$ 0.044}}%
&\scriptsize{\textbf{62.67}}
&\scriptsize{21.82 \tiny{$\pm$ 3.73}}%
&\scriptsize{0.532 \tiny{$\pm$ 0.184}}%
&\scriptsize{\textbf{0.358} \tiny{$\pm$ 0.137}}%
&\scriptsize{\textbf{139.39}}
&\scriptsize{23.44 \tiny{$\pm$ 1.43}}%
&\scriptsize{0.633 \tiny{$\pm$ 0.078}}%
&\scriptsize{0.182 \tiny{$\pm$ 0.053}}%
&\scriptsize{\underline{68.37}}
&\scriptsize{20.34 \tiny{$\pm$ 3.31}}%
&\scriptsize{0.480 \tiny{$\pm$ 0.181}}%
&\scriptsize{0.425 \tiny{$\pm$ 0.142}}%
&\scriptsize{180.41}
\\
&\scriptsize{DiffPIR}
&\scriptsize{22.45 \tiny{$\pm$ 1.00}}%
&\scriptsize{0.482 \tiny{$\pm$ 0.074}}%
&\scriptsize{0.493 \tiny{$\pm$ 0.084}}%
&\scriptsize{211.40}
&\scriptsize{19.85 \tiny{$\pm$ 1.90}}%
&\scriptsize{0.367 \tiny{$\pm$ 0.101}}%
&\scriptsize{0.676 \tiny{$\pm$ 0.109}}%
&\scriptsize{250.67}
&\scriptsize{20.48 \tiny{$\pm$ 0.95}}%
&\scriptsize{0.362 \tiny{$\pm$ 0.070}}%
&\scriptsize{0.608 \tiny{$\pm$ 0.082}}%
&\scriptsize{274.28}
&\scriptsize{17.13 \tiny{$\pm$ 1.58}}%
&\scriptsize{0.250 \tiny{$\pm$ 0.091}}%
&\scriptsize{0.703 \tiny{$\pm$ 0.139}}%
&\scriptsize{286.64}
\\
&\scriptsize{DCPS}
&\scriptsize{22.45 \tiny{$\pm$ 1.83}}%
&\scriptsize{0.563 \tiny{$\pm$ 0.097}}%
&\scriptsize{0.307 \tiny{$\pm$ 0.087}}%
&\scriptsize{172.98}
&\scriptsize{18.49 \tiny{$\pm$ 2.62}}%
&\scriptsize{0.356 \tiny{$\pm$ 0.144}}%
&\scriptsize{0.593 \tiny{$\pm$ 0.146}}%
&\scriptsize{224.90}
&\scriptsize{21.65 \tiny{$\pm$ 3.99}}%
&\scriptsize{0.585 \tiny{$\pm$ 0.184}}%
&\scriptsize{0.351 \tiny{$\pm$ 0.240}}%
&\scriptsize{148.53}
&\scriptsize{14.22 \tiny{$\pm$ 3.16}}%
&\scriptsize{0.245 \tiny{$\pm$ 0.148}}%
&\scriptsize{0.765 \tiny{$\pm$ 0.238}}%
&\scriptsize{265.70}
\\
&\scriptsize{RED-diff} 
&\scriptsize{27.99 \tiny{$\pm$ 1.56}}%
&\scriptsize{0.751 \tiny{$\pm$ 0.051}}%
&\scriptsize{0.273 \tiny{$\pm$ 0.067}}%
&\scriptsize{85.72}
&\scriptsize{23.80 \tiny{$\pm$ 3.68}}%
&\scriptsize{0.603 \tiny{$\pm$ 0.136}}%
&\scriptsize{0.558 \tiny{$\pm$ 0.144}}%
&\scriptsize{169.24}
&\scriptsize{24.80 \tiny{$\pm$ 1.42}}%
&\scriptsize{0.573 \tiny{$\pm$ 0.066}}%
&\scriptsize{0.381 \tiny{$\pm$ 0.086}}%
&\scriptsize{150.30}
&\scriptsize{21.66 \tiny{$\pm$ 3.03}}%
&\scriptsize{0.459 \tiny{$\pm$ 0.095}}%
&\scriptsize{0.565 \tiny{$\pm$ 0.095}}%
&\scriptsize{269.02}
\\
&\scriptsize{DAPS} 
&\scriptsize{23.73 \tiny{$\pm$ 2.11}}%
&\scriptsize{0.586 \tiny{$\pm$ 0.098}}%
&\scriptsize{0.362 \tiny{$\pm$ 0.092}}%
&\scriptsize{202.42}
&\scriptsize{20.35 \tiny{$\pm$ 2.11}}%
&\scriptsize{0.450 \tiny{$\pm$ 0.098}}%
&\scriptsize{0.619 \tiny{$\pm$ 0.092}}%
&\scriptsize{251.12}
&\scriptsize{23.74 \tiny{$\pm$ 3.15}}%
&\scriptsize{0.621 \tiny{$\pm$ 0.121}}%
&\scriptsize{0.330 \tiny{$\pm$ 0.173}}%
&\scriptsize{\textbf{63.44}}
&\scriptsize{19.27 \tiny{$\pm$ 3.15}}%
&\scriptsize{0.447 \tiny{$\pm$ 0.121}}%
&\scriptsize{0.544 \tiny{$\pm$ 0.173}}%
&\scriptsize{232.58}
\\
&\scriptsize{\textbf{Robust-GD}} 
&\scriptsize{\textbf{30.00} \tiny{$\pm$ 1.75}}%
&\scriptsize{\textbf{0.825} \tiny{$\pm$ 0.044}}%
&\scriptsize{\textbf{0.132} \tiny{$\pm$ 0.040}}%
&\scriptsize{\underline{63.44}}
&\scriptsize{\textbf{25.96} \tiny{$\pm$ 3.90}}%
&\scriptsize{\textbf{0.679} \tiny{$\pm$ 0.120}}%
&\scriptsize{\underline{0.424} \tiny{$\pm$ 0.131}}%
&\scriptsize{\underline{159.84}}
&\scriptsize{\textbf{29.87} \tiny{$\pm$ 1.86}}%
&\scriptsize{\textbf{0.801} \tiny{$\pm$ 0.049}}%
&\scriptsize{\textbf{0.116} \tiny{$\pm$ 0.045}}%
&\scriptsize{81.80}
&\scriptsize{\textbf{25.80} \tiny{$\pm$ 3.79}}%
&\scriptsize{\textbf{0.696} \tiny{$\pm$ 0.106}}%
&\scriptsize{\textbf{0.300} \tiny{$\pm$ 0.139}}%
&\scriptsize{\textbf{115.26}}
\\
&\scriptsize{\textbf{Robust-CG}} 
&\scriptsize{\underline{29.48} \tiny{$\pm$ 1.81}}%
&\scriptsize{\underline{0.820} \tiny{$\pm$ 0.051}}%
&\scriptsize{\underline{0.150} \tiny{$\pm$ 0.047}}%
&\scriptsize{63.50}
&\scriptsize{\underline{24.79} \tiny{$\pm$ 4.34}}%
&\scriptsize{\underline{0.673} \tiny{$\pm$ 0.155}}%
&\scriptsize{0.458 \tiny{$\pm$ 0.162}}%
&\scriptsize{169.11}
&\scriptsize{\underline{29.27} \tiny{$\pm$ 2.36}}%
&\scriptsize{\underline{0.793} \tiny{$\pm$ 0.065}}%
&\scriptsize{\underline{0.136} \tiny{$\pm$ 0.057}}%
&\scriptsize{87.09}
&\scriptsize{\underline{24.94} \tiny{$\pm$ 4.13}}%
&\scriptsize{\underline{0.665} \tiny{$\pm$ 0.124}}%
&\scriptsize{\underline{0.376} \tiny{$\pm$ 0.154}}%
&\scriptsize{\underline{152.85}}
\\
\cline{1-18}
\multirow{7}{*}{\scriptsize{0.10}}
&\scriptsize{DPS}
&\scriptsize{22.06 \tiny{$\pm$ 1.42}}%
&\scriptsize{0.646 \tiny{$\pm$ 0.077}}%
&\scriptsize{0.178 \tiny{$\pm$ 0.051}}%
&\scriptsize{\underline{63.49}}
&\scriptsize{19.70 \tiny{$\pm$ 2.89}}%
&\scriptsize{0.490 \tiny{$\pm$ 0.185}}%
&\scriptsize{\textbf{0.405} \tiny{$\pm$ 0.131}}%
&\scriptsize{\textbf{160.82}}
&\scriptsize{20.90 \tiny{$\pm$ 1.37}}%
&\scriptsize{0.596 \tiny{$\pm$ 0.082}}%
&\scriptsize{0.208 \tiny{$\pm$ 0.058}}%
&\scriptsize{\textbf{68.96}}
&\scriptsize{18.60 \tiny{$\pm$ 2.80}}%
&\scriptsize{0.437 \tiny{$\pm$ 0.186}}%
&\scriptsize{0.455 \tiny{$\pm$ 0.130}}%
&\scriptsize{192.53}
\\
&\scriptsize{DiffPIR}
&\scriptsize{18.07 \tiny{$\pm$ 1.02}}%
&\scriptsize{0.355 \tiny{$\pm$ 0.083}}%
&\scriptsize{0.622 \tiny{$\pm$ 0.083}}%
&\scriptsize{308.93}
&\scriptsize{16.60 \tiny{$\pm$ 1.67}}%
&\scriptsize{0.274 \tiny{$\pm$ 0.098}}%
&\scriptsize{0.740 \tiny{$\pm$ 0.088}}%
&\scriptsize{292.01}
&\scriptsize{17.11 \tiny{$\pm$ 0.93}}%
&\scriptsize{0.259 \tiny{$\pm$ 0.071}}%
&\scriptsize{0.712 \tiny{$\pm$ 0.078}}%
&\scriptsize{340.27}
&\scriptsize{14.59 \tiny{$\pm$ 1.52}}%
&\scriptsize{0.180 \tiny{$\pm$ 0.081}}%
&\scriptsize{0.781 \tiny{$\pm$ 0.117}}%
&\scriptsize{317.89}
\\
&\scriptsize{DCPS}
&\scriptsize{15.46 \tiny{$\pm$ 1.65}}%
&\scriptsize{0.302 \tiny{$\pm$ 0.124}}%
&\scriptsize{0.575 \tiny{$\pm$ 0.095}}%
&\scriptsize{331.91}
&\scriptsize{12.89 \tiny{$\pm$ 2.54}}%
&\scriptsize{0.193 \tiny{$\pm$ 0.123}}%
&\scriptsize{0.702 \tiny{$\pm$ 0.092}}%
&\scriptsize{305.81}
&\scriptsize{12.60 \tiny{$\pm$ 3.45}}%
&\scriptsize{0.174 \tiny{$\pm$ 0.176}}%
&\scriptsize{0.927 \tiny{$\pm$ 0.292}}%
&\scriptsize{338.22}
&\scriptsize{10.29 \tiny{$\pm$ 1.92}}%
&\scriptsize{0.112 \tiny{$\pm$ 0.084}}%
&\scriptsize{0.953 \tiny{$\pm$ 0.183}}%
&\scriptsize{332.85}
\\
&\scriptsize{RED-diff} 
&\scriptsize{22.96 \tiny{$\pm$ 1.70}}%
&\scriptsize{0.626 \tiny{$\pm$ 0.076}}%
&\scriptsize{0.391 \tiny{$\pm$ 0.090}}%
&\scriptsize{139.05}
&\scriptsize{20.76 \tiny{$\pm$ 2.68}}%
&\scriptsize{0.497 \tiny{$\pm$ 0.118}}%
&\scriptsize{0.645 \tiny{$\pm$ 0.129}}%
&\scriptsize{219.67}
&\scriptsize{17.36 \tiny{$\pm$ 2.52}}%
&\scriptsize{0.311 \tiny{$\pm$ 0.109}}%
&\scriptsize{0.836 \tiny{$\pm$ 0.185}}%
&\scriptsize{303.61}
&\scriptsize{16.71 \tiny{$\pm$ 3.08}}%
&\scriptsize{0.270 \tiny{$\pm$ 0.116}}%
&\scriptsize{0.807 \tiny{$\pm$ 0.171}}%
&\scriptsize{347.46}
\\
&\scriptsize{DAPS} 
&\scriptsize{16.18 \tiny{$\pm$ 2.26}}%
&\scriptsize{0.356 \tiny{$\pm$ 0.122}}%
&\scriptsize{0.561 \tiny{$\pm$ 0.104}}%
&\scriptsize{297.13}
&\scriptsize{14.17 \tiny{$\pm$ 2.26}}%
&\scriptsize{0.253 \tiny{$\pm$ 0.122}}%
&\scriptsize{0.720 \tiny{$\pm$ 0.104}}%
&\scriptsize{308.08}
&\scriptsize{14.99 \tiny{$\pm$ 3.01}}%
&\scriptsize{0.331 \tiny{$\pm$ 0.130}}%
&\scriptsize{0.718 \tiny{$\pm$ 0.178}}%
&\scriptsize{292.81}
&\scriptsize{12.45 \tiny{$\pm$ 3.01}}%
&\scriptsize{0.226 \tiny{$\pm$ 0.130}}%
&\scriptsize{0.823 \tiny{$\pm$ 0.178}}%
&\scriptsize{305.09}
\\
&\scriptsize{\textbf{Robust-GD}} 
&\scriptsize{\underline{29.27} \tiny{$\pm$ 1.68}}%
&\scriptsize{\underline{0.817} \tiny{$\pm$ 0.044}}%
&\scriptsize{\textbf{0.138} \tiny{$\pm$ 0.041}}%
&\scriptsize{65.59}
&\scriptsize{\textbf{24.90} \tiny{$\pm$ 3.71}}%
&\scriptsize{\underline{0.660} \tiny{$\pm$ 0.114}}%
&\scriptsize{0.453 \tiny{$\pm$ 0.130}}%
&\scriptsize{171.81}
&\scriptsize{\textbf{29.44} \tiny{$\pm$ 1.76}}%
&\scriptsize{\textbf{0.789} \tiny{$\pm$ 0.048}}%
&\scriptsize{\textbf{0.123} \tiny{$\pm$ 0.046}}%
&\scriptsize{\underline{84.79}}
&\scriptsize{\textbf{25.27} \tiny{$\pm$ 3.55}}%
&\scriptsize{\textbf{0.669} \tiny{$\pm$ 0.103}}%
&\scriptsize{\textbf{0.330} \tiny{$\pm$ 0.134}}%
&\scriptsize{\textbf{128.60}}
\\
&\scriptsize{\textbf{Robust-CG}} 
&\scriptsize{\textbf{29.38} \tiny{$\pm$ 1.78}}%
&\scriptsize{\textbf{0.819} \tiny{$\pm$ 0.051}}%
&\scriptsize{\underline{0.151} \tiny{$\pm$ 0.047}}%
&\scriptsize{\textbf{62.52}}
&\scriptsize{\underline{24.73} \tiny{$\pm$ 4.29}}%
&\scriptsize{\textbf{0.672} \tiny{$\pm$ 0.155}}%
&\scriptsize{\underline{0.450} \tiny{$\pm$ 0.163}}%
&\scriptsize{\underline{167.56}}
&\scriptsize{\underline{28.91} \tiny{$\pm$ 2.16}}%
&\scriptsize{\underline{0.777} \tiny{$\pm$ 0.062}}%
&\scriptsize{\underline{0.144} \tiny{$\pm$ 0.057}}%
&\scriptsize{91.22}
&\scriptsize{\underline{24.57} \tiny{$\pm$ 3.80}}%
&\scriptsize{\underline{0.638} \tiny{$\pm$ 0.119}}%
&\scriptsize{\underline{0.401} \tiny{$\pm$ 0.151}}%
&\scriptsize{\underline{165.45}}
\\
\hline
\end{tabular}
}}
\end{center}
\vspace{-1.5em}
\caption{(\textbf{Linear IPs}) \textbf{Gaussian deblurring} and \textbf{motion deblurring} with additive Gaussian noise ($\sigma = 0.05$) and contamination fraction $\rho = 0.02$ or $0.10$.}
\label{tab:linear_deblur}
\vspace{-1.5em}
\end{table*}

\subsection{Nonlinear deblurring}
\label{sec:exp_nonlinear_deblur}
For nonlinear deblurring, we utilize the learned blurring operators from~\cite{tran2021explore} with a known Gaussian-shaped kernel, following the setting in~\cite{chung2023dps, wang2024dmplug}. The corresponding threshold parameter $\delta$ in Robust-GD and Robust-CG is set to $0.01$. The finite difference approximation parameter $\eta$ in Robust-CG is set to $0.0001$. The results summarized in Table~\ref{tab:non} demonstrate that our Robust-GD and Robust-CG methods achieve the best performance across all tasks, highlighting the effectiveness of our methods. 



\begin{table}[!htbp]
\begin{center}
\renewcommand{\arraystretch}{0.9}
\resizebox{0.5\textwidth}{!}{
\setlength{\tabcolsep}{0.85mm}{
\begin{tabular}{c c c c c c c c c c}
\hline
\multicolumn{2}{c}{}
&\multicolumn{4}{c}{\scriptsize{\textbf{CelebA ($256 \times 256$)}}}
&\multicolumn{4}{c}{\scriptsize{\textbf{ImageNet ($256 \times 256$)}}}
\\
\cline{1-10}
\scriptsize{$\rho$}
&\scriptsize{Methods}
&\scriptsize{PSNR$\uparrow$}
&\scriptsize{SSIM$\uparrow$}
&\scriptsize{LPIPS$\downarrow$}
&\scriptsize{FID$\downarrow$}
&\scriptsize{PSNR$\uparrow$}
&\scriptsize{SSIM$\uparrow$}
&\scriptsize{LPIPS$\downarrow$}
&\scriptsize{FID$\downarrow$}
\\
\cline{1-10}
\multirow{5}{*}{\scriptsize{0.02}}
&\scriptsize{DPS}
&\scriptsize{23.42 \tiny{$\pm$ 1.54}}%
&\scriptsize{0.632 \tiny{$\pm$ 0.076}}%
&\scriptsize{0.203 \tiny{$\pm$ 0.055}}%
&\scriptsize{\underline{69.71}}
&\scriptsize{21.01 \tiny{$\pm$ 3.53}}%
&\scriptsize{0.493 \tiny{$\pm$ 0.174}}%
&\scriptsize{0.439 \tiny{$\pm$ 0.148}}%
&\scriptsize{\underline{191.88}}
\\
&\scriptsize{RED-diff}
&\scriptsize{22.49 \tiny{$\pm$ 1.61}}%
&\scriptsize{0.490 \tiny{$\pm$ 0.064}}%
&\scriptsize{0.568 \tiny{$\pm$ 0.116}}%
&\scriptsize{169.33}
&\scriptsize{18.53 \tiny{$\pm$ 2.31}}%
&\scriptsize{0.289 \tiny{$\pm$ 0.067}}%
&\scriptsize{0.727 \tiny{$\pm$ 0.111}}%
&\scriptsize{305.67}
\\
&\scriptsize{DAPS} 
&\scriptsize{20.90 \tiny{$\pm$ 1.81}}%
&\scriptsize{0.428 \tiny{$\pm$ 0.113}}%
&\scriptsize{0.658 \tiny{$\pm$ 0.177}}%
&\scriptsize{237.47}
&\scriptsize{20.22 \tiny{$\pm$ 1.81}}%
&\scriptsize{0.457 \tiny{$\pm$ 0.113}}%
&\scriptsize{0.530 \tiny{$\pm$ 0.177}}%
&\scriptsize{202.20}
\\
&\scriptsize{\textbf{Robust-GD}} 
&\scriptsize{\textbf{28.25} \tiny{$\pm$ 1.55}}%
&\scriptsize{\textbf{0.766} \tiny{$\pm$ 0.037}}%
&\scriptsize{\textbf{0.136} \tiny{$\pm$ 0.040}}%
&\scriptsize{\textbf{68.55}}
&\scriptsize{\textbf{26.40} \tiny{$\pm$ 3.61}}%
&\scriptsize{\textbf{0.741} \tiny{$\pm$ 0.084}}%
&\scriptsize{\textbf{0.214} \tiny{$\pm$ 0.100}}%
&\scriptsize{\textbf{83.52}}
\\
&\scriptsize{\textbf{Robust-CG}} 
&\scriptsize{\underline{26.89} \tiny{$\pm$ 1.59}}%
&\scriptsize{\underline{0.733} \tiny{$\pm$ 0.055}}%
&\scriptsize{\underline{0.192} \tiny{$\pm$ 0.060}}%
&\scriptsize{85.30}
&\scriptsize{\underline{23.66} \tiny{$\pm$ 3.80}}%
&\scriptsize{\underline{0.634} \tiny{$\pm$ 0.137}}%
&\scriptsize{\underline{0.411} \tiny{$\pm$ 0.147}}%
&\scriptsize{201.28}
\\
\cline{1-10}
\multirow{5}{*}{\scriptsize{0.10}}
&\scriptsize{DPS}
&\scriptsize{21.26 \tiny{$\pm$ 1.45}}%
&\scriptsize{0.609 \tiny{$\pm$ 0.077}}%
&\scriptsize{0.211 \tiny{$\pm$ 0.052}}%
&\scriptsize{\textbf{70.11}}
&\scriptsize{19.19 \tiny{$\pm$ 2.80}}%
&\scriptsize{0.455 \tiny{$\pm$ 0.179}}%
&\scriptsize{0.463 \tiny{$\pm$ 0.147}}%
&\scriptsize{\underline{183.46}}
\\
&\scriptsize{RED-diff}
&\scriptsize{16.37 \tiny{$\pm$ 1.96}}%
&\scriptsize{0.274 \tiny{$\pm$ 0.088}}%
&\scriptsize{0.922 \tiny{$\pm$ 0.167}}%
&\scriptsize{288.09}
&\scriptsize{14.95 \tiny{$\pm$ 2.39}}%
&\scriptsize{0.166 \tiny{$\pm$ 0.087}}%
&\scriptsize{0.998 \tiny{$\pm$ 0.185}}%
&\scriptsize{336.14}
\\
&\scriptsize{DAPS} 
&\scriptsize{15.89 \tiny{$\pm$ 2.17}}%
&\scriptsize{0.232 \tiny{$\pm$ 0.110}}%
&\scriptsize{1.036 \tiny{$\pm$ 0.174}}%
&\scriptsize{439.06}
&\scriptsize{15.21 \tiny{$\pm$ 2.17}}%
&\scriptsize{0.228 \tiny{$\pm$ 0.110}}%
&\scriptsize{0.947 \tiny{$\pm$ 0.174}}%
&\scriptsize{316.56}
\\
&\scriptsize{\textbf{Robust-GD}} 
&\scriptsize{\textbf{27.06} \tiny{$\pm$ 1.29}}%
&\scriptsize{\underline{0.711} \tiny{$\pm$ 0.037}}%
&\scriptsize{\textbf{0.156} \tiny{$\pm$ 0.044}}%
&\scriptsize{\underline{80.23}}
&\scriptsize{\textbf{24.55} \tiny{$\pm$ 3.07}}%
&\scriptsize{\textbf{0.636} \tiny{$\pm$ 0.084}}%
&\scriptsize{\textbf{0.253} \tiny{$\pm$ 0.093}}%
&\scriptsize{\textbf{108.10}}
\\
&\scriptsize{\textbf{Robust-CG}} 
&\scriptsize{\underline{26.80} \tiny{$\pm$ 1.54}}%
&\scriptsize{\textbf{0.728} \tiny{$\pm$ 0.058}}%
&\scriptsize{\underline{0.188} \tiny{$\pm$ 0.061}}%
&\scriptsize{89.73}
&\scriptsize{\underline{23.67} \tiny{$\pm$ 3.75}}%
&\scriptsize{\underline{0.632} \tiny{$\pm$ 0.135}}%
&\scriptsize{\underline{0.404} \tiny{$\pm$ 0.143}}%
&\scriptsize{191.30}
\\
\hline
\end{tabular}
}
}
\end{center}
\vspace{-1.25em}
\caption{\textbf{Nonlinear deblurring} with additive Gaussian noise ($\sigma = 0.05$) and contamination fraction $\rho = 0.02$ or $0.10$.}
\label{tab:non}
\vspace{-1.25em}
\end{table}

\subsection{High Gaussian noisy condition}
We follow~\cite{kawar2022ddrm, chen2025robust, shoushtari2025unsupervised} and validate the robustness of our proposed Robust-GD and Robust-CG methods under high Gaussian noisy conditions. We conduct experiments on the CelebA dataset, using 100 randomly selected validation images for super-resolution and inpainting tasks. We set the noise level at $\sigma = 0.5$ as in~\cite{shoushtari2025unsupervised}, and the contamination fraction at $\rho = 0.02$. The settings for these two tasks are identical to those detailed in Section~\ref{sec:exp_ip_sp}. The results presented in Table~\ref{tab:high_noise} demonstrate that our methods achieve the best performance compared to all other DM-based approaches, confirming the robustness of our methods against high Gaussian noise.

\begin{table}[!htbp]
\begin{center}
\renewcommand{\arraystretch}{0.9}
\resizebox{0.5\textwidth}{!}{
\setlength{\tabcolsep}{0.85mm}{
\begin{tabular}{c c c c c c c}
\hline
\multicolumn{1}{c}{}
&\multicolumn{3}{c}{\scriptsize{\textbf{Super-resolution ($4\times$)}}}
&\multicolumn{3}{c}{\scriptsize{\textbf{Inpainting (random $70\%$)}}}
\\
\cline{2-7}
\scriptsize{Methods}
&\scriptsize{PSNR$\uparrow$}
&\scriptsize{SSIM$\uparrow$}
&\scriptsize{LPIPS$\downarrow$}
&\scriptsize{PSNR$\uparrow$}
&\scriptsize{SSIM$\uparrow$}
&\scriptsize{LPIPS$\downarrow$}
\\
\cline{1-7}
\scriptsize{DPS}
&\scriptsize{20.55}%
&\scriptsize{0.546}%
&\scriptsize{\underline{0.246}}%
&\scriptsize{21.90}%
&\scriptsize{0.604}%
&\scriptsize{0.212}%
\\
\scriptsize{DiffPIR}
&\scriptsize{15.68}%
&\scriptsize{0.192}%
&\scriptsize{0.643}%
&\scriptsize{20.31}%
&\scriptsize{0.471}%
&\scriptsize{0.382}%
\\
\scriptsize{DCPS}
&\scriptsize{18.77}%
&\scriptsize{0.476}%
&\scriptsize{0.315}%
&\scriptsize{22.54}%
&\scriptsize{0.620}%
&\scriptsize{0.202}%
\\
\scriptsize{RED-diff} 
&\scriptsize{12.88}%
&\scriptsize{0.109}%
&\scriptsize{0.793}%
&\scriptsize{17.47}%
&\scriptsize{0.186}%
&\scriptsize{0.873}%
\\
\scriptsize{DAPS} 
&\scriptsize{14.29}%
&\scriptsize{0.143}%
&\scriptsize{0.784}%
&\scriptsize{16.99}%
&\scriptsize{0.181}%
&\scriptsize{0.901}%
\\
\scriptsize{\textbf{Robust-GD}} 
&\scriptsize{\underline{22.16}}%
&\scriptsize{\underline{0.590}}%
&\scriptsize{0.279}%
&\scriptsize{\underline{23.91}}%
&\scriptsize{\textbf{0.690}}%
&\scriptsize{\textbf{0.179}}%
\\
\scriptsize{\textbf{Robust-CG}} 
&\scriptsize{\textbf{23.04}}%
&\scriptsize{\textbf{0.632}}%
&\scriptsize{\textbf{0.245}}%
&\scriptsize{\textbf{24.62}}%
&\scriptsize{\underline{0.689}}%
&\scriptsize{\underline{0.200}}%
\\
\hline
\end{tabular}
}
}
\end{center}
\vspace{-1.25em}
\caption{\textbf{Super-resolution ($4\times$)} and \textbf{inpainting (random $70\%$)} with additive Gaussian noise ($\sigma = 0.5$) and contamination fraction $\rho = 0.02$.}
\label{tab:high_noise}
\vspace{-1.25em}
\end{table}

\subsection{Ablation study}
\label{sec:ablation}
We validate the effect of the Huber loss threshold parameter $\delta$ in our algorithms and the finite-difference approximation parameter $\eta$ in the Robust-CG method. Additionally, we present the performance of Robust-GD with varying learning rates $\eta_{x}$ and compare the use of $\bm{g}^{\rm T}\bm{g}$ versus $\bm{g}^{\rm T}\bm{d}$ terms for calculating $\alpha_j$ in Robust-CG (referencing Eq.~\eqref{eq:cg_gg}) in the supplementary material.
\subsubsection{Huber loss threshold parameter}
\label{sec:abl_huber}
To evaluate the effect of the Huber loss threshold parameter $\delta$, we test the Robust-CG method on 100 validation images from the CelebA dataset for Gaussian and motion deblurring tasks, using a contamination factor of $\rho = 0.10$ and a Gaussian noise level of $\sigma = 0.05$. The corresponding experimental results for Robust-GD are presented in the supplementary material. We choose the value of $\delta$ within $\{0.005, 0.01, 0.02, 0.04\}$. The results presented in Table~\ref{tab:weight} indicate that Robust-CG performs similarly across these different threshold values, demonstrating the robustness of our method to the selection of the threshold parameter.

\begin{table}[!htbp]
\begin{center}
\renewcommand{\arraystretch}{0.9}
\resizebox{0.5\textwidth}{!}{
\setlength{\tabcolsep}{0.85mm}{
\begin{tabular}{c c c c c c c}
\hline
\multicolumn{1}{c}{}
&\multicolumn{3}{c}{\scriptsize{\textbf{Gaussian Deblurring}}}
&\multicolumn{3}{c}{\scriptsize{\textbf{Motion Deblurring}}}
\\
\cline{2-7}
\scriptsize{Threshold $\delta$}
&\scriptsize{PSNR$\uparrow$}
&\scriptsize{SSIM$\uparrow$}
&\scriptsize{LPIPS$\downarrow$}
&\scriptsize{PSNR$\uparrow$}
&\scriptsize{SSIM$\uparrow$}
&\scriptsize{LPIPS$\downarrow$}
\\
\cline{1-7}
\scriptsize{0.005}
&\scriptsize{28.75}%
&\scriptsize{0.804}%
&\scriptsize{0.164}%
&\scriptsize{27.91}%
&\scriptsize{0.786}%
&\scriptsize{0.161}%
\\
\scriptsize{0.010}
&\scriptsize{29.04}%
&\scriptsize{0.811}%
&\scriptsize{0.157}%
&\scriptsize{28.55}%
&\scriptsize{0.793}%
&\scriptsize{0.146}%
\\
\scriptsize{0.020}
&\scriptsize{29.38}%
&\scriptsize{0.819}%
&\scriptsize{0.151}%
&\scriptsize{28.91}%
&\scriptsize{0.777}%
&\scriptsize{0.144}%
\\
\scriptsize{0.040} 
&\scriptsize{29.64}%
&\scriptsize{0.823}%
&\scriptsize{0.145}%
&\scriptsize{27.67}%
&\scriptsize{0.681}%
&\scriptsize{0.220}%
\\
\hline
\end{tabular}
}
}
\end{center}
\vspace{-1.25em}
\caption{Performance of Robust-CG with different value of the Huber loss threshold $\delta$ on \textbf{Gaussian deblurring} and \textbf{motion deblurring} with additive Gaussian noise ($\sigma = 0.05$) and contamination fraction $0.10$.}
\label{tab:weight}
\vspace{-1.25em}
\end{table}

\subsubsection{Finite-difference approximation parameter}
We validate the effect of the finite-difference approximation parameter $\eta$ in the Robust-CG method on two tasks, namely Gaussian deblurring and the nonlinear deblurring task. We use 100 validation images from the CelebA dataset with a contamination factor of $\rho = 0.10$ and a Gaussian noise level of $\sigma = 0.05$. We choose three values of $\eta$ within $\{0.0010, 0.0005,0.0001\}$. The results presented in Table~\ref{tab:ablation_eta} indicate that the choice of $\eta$ does not affect the reconstruction results for Gaussian deblurring and has a slight effect on the results for the nonlinear deblurring task.

\begin{table}[!htbp]
\begin{center}
\renewcommand{\arraystretch}{0.9}
\resizebox{0.5\textwidth}{!}{
\setlength{\tabcolsep}{0.85mm}{
\begin{tabular}{c c c c c c c}
\hline
\multicolumn{1}{c}{}
&\multicolumn{3}{c}{\scriptsize{\textbf{Gaussian Deblurring}}}
&\multicolumn{3}{c}{\scriptsize{\textbf{Nonliear Deblurring}}}
\\
\cline{2-7}
\scriptsize{Parameter $\eta$}
&\scriptsize{PSNR$\uparrow$}
&\scriptsize{SSIM$\uparrow$}
&\scriptsize{LPIPS$\downarrow$}
&\scriptsize{PSNR$\uparrow$}
&\scriptsize{SSIM$\uparrow$}
&\scriptsize{LPIPS$\downarrow$}
\\
\cline{1-7}
\scriptsize{$0.0010$}
&\scriptsize{29.38}%
&\scriptsize{0.819}%
&\scriptsize{0.151}%
&\scriptsize{25.97}%
&\scriptsize{0.681}%
&\scriptsize{0.212}%
\\
\scriptsize{$0.0005$}
&\scriptsize{29.38}%
&\scriptsize{0.819}%
&\scriptsize{0.151}%
&\scriptsize{26.37}%
&\scriptsize{0.703}%
&\scriptsize{0.195}%
\\
\scriptsize{$0.0001$}
&\scriptsize{29.38}%
&\scriptsize{0.819}%
&\scriptsize{0.151}%
&\scriptsize{26.80}%
&\scriptsize{0.728}%
&\scriptsize{0.188}%
\\
\hline
\end{tabular}
}
}
\end{center}
\vspace{-1.25em}
\caption{Performance of Robust-CG with different value of finite-difference approximation parameter $\eta$ on \textbf{Gaussian deblurring} and \textbf{nonlinear deblurring} with additive Gaussian noise ($\sigma = 0.05$) and contamination fraction $0.10$.}
\label{tab:ablation_eta}
\vspace{-1.25em}
\end{table}

\section{Conclusion}
In this paper, we present Robust-GD and Robust-CG, two novel DM-based methods to address IPs contaminated with outliers. Numerical results demonstrate the effectiveness of our proposed methods in mitigating the influence of outliers, thereby confirming their superior robustness compared to existing DM-based methods.
{
    \small
    \vspace{-1em}
    \bibliographystyle{ieeenat_fullname}
    \bibliography{main}

@String(CVPR= {IEEE Conf. Comput. Vis. Pattern Recog.})

@String(ICCV= {Int. Conf. Comput. Vis.})

@String(ECCV= {Eur. Conf. Comput. Vis.})

@String(ICASSP=	{ICASSP})

@String(ICLR = {Int. Conf. Learn. Represent.})

@String(AAAI = {AAAI})

@String(CVPR  = {CVPR})

@String(ICCV  = {ICCV})

@String(ECCV  = {ECCV})

@String(ICLR  = {ICLR})

@inproceedings{chung2023parallel,
  title={Parallel diffusion models of operator and image for blind inverse problems},
  author={Chung, Hyungjin and Kim, Jeongsol and Kim, Sehui and Ye, Jong Chul},
  booktitle={CVPR},
  year={2023}
}

@inproceedings{zheng2025integrating,
  title = {Integrating intermediate layer optimization and projected gradient descent for solving inverse problems with diffusion models},
  author={Zheng, Yang and Li, Wen and Liu, Zhaoqiang},
  booktitle={ICML},
  year={2025}
}

@inproceedings{murata2023gibbsddrm,
  title={Gibbs{DDRM}: A partially collapsed {G}ibbs sampler for solving blind inverse problems with denoising diffusion restoration},
  author={Murata, Naoki and Saito, Koichi and Lai, Chieh-Hsin and Takida, Yuhta and Uesaka, Toshimitsu and Mitsufuji, Yuki and Ermon, Stefano},
  booktitle={ICML},
  year={2023},
}

@inproceedings{wang2024dmplug,
  title={{DMP}lug: A plug-in method for solving inverse problems with diffusion models},
  author={Wang, Hengkang and Zhang, Xu and Li, Taihui and Wan, Yuxiang and Chen, Tiancong and Sun, Ju},
  booktitle={NeurIPS},
  year={2024}
}

@inproceedings{zhang2024daps,
      title={Improving diffusion inverse problem solving with decoupled noise annealing}, 
      author={Bingliang Zhang and Wenda Chu and Julius Berner and Chenlin Meng and Anima Anandkumar and Yang Song},
      year={2025},
      booktitle={CVPR}, 
}

@inproceedings{kawar2022ddrm,
  title={Denoising diffusion restoration models},
  author={Kawar, Bahjat and Elad, Michael and Ermon, Stefano and Song, Jiaming},
  booktitle={NeurIPS},
  year={2022}
}

@inproceedings{chung2022mcg,
  title={Improving diffusion models for inverse problems using manifold constraints},
  author={Chung, Hyungjin and Sim, Byeongsu and Ryu, Dohoon and Ye, Jong Chul},
  booktitle={NeurIPS},
  year={2022}
}

@inproceedings{chung2023dps,
  title={Diffusion posterior sampling for general noisy inverse problems},
  author={Chung, Hyungjin and Kim, Jeongsol and Mccann, Michael T and Klasky, Marc L and Ye, Jong Chul},
  booktitle={ICLR},
  year={2023}
}

@inproceedings{song2023pigdm,
  title={Pseudoinverse-guided diffusion models for inverse problems},
  author={Song, Jiaming and Vahdat, Arash and Mardani, Morteza and Kautz, Jan},
  booktitle={ICLR},
  year={2023}
}

@Book{Fou13,
  Title                    = {A mathematical introduction to compressive sensing},
  Author                   = {Simon Foucart and Holger Rauhut},
  Publisher                = {Springer New York},
  Year                     = {2013}
}

@inproceedings{saharia2022palette,
  title={Palette: {I}mage-to-image diffusion models},
  author={Saharia, Chitwan and Chan, William and Chang, Huiwen and Lee, Chris and Ho, Jonathan and Salimans, Tim and Fleet, David and Norouzi, Mohammad},
  booktitle={SIGGRAPH},
  year={2022}
}

@inproceedings{pan2016robust,
  title={Robust kernel estimation with outliers handling for image deblurring},
  author={Pan, Jinshan and Lin, Zhouchen and Su, Zhixun and Yang, Ming-Hsuan},
  booktitle={CVPR},
  year={2016}
}

@inproceedings{dong2017blind,
  title={Blind image deblurring with outlier handling},
  author={Dong, Jiangxin and Pan, Jinshan and Su, Zhixun and Yang, Ming-Hsuan},
  booktitle={ICCV},
  year={2017}
}

@article{song2024robust,
  title={Robust image restoration with an adaptive {H}uber function based fidelity},
  author={Song, Lingfei and Huang, Hua},
  journal={International Journal of Computer Vision},
  year={2024},
}

@inproceedings{mardani2023variational,
      title={{A} variational perspective on solving inverse problems with diffusion models}, 
      author={Mardani, Morteza and Song, Jiaming and Kautz, Jan and Vahdat, Arash},
      year={2023},
      booktitle={ICLR}, 
}

@inproceedings{zhu2023denoising,
  title={Denoising diffusion models for plug-and-play image restoration},
  author={Zhu, Yuanzhi and Zhang, Kai and Liang, Jingyun and Cao, Jiezhang and Wen, Bihan and Timofte, Radu and Van Gool, Luc},
  booktitle={CVPR},
  year={2023}
}

@inproceedings{zhang2018unreasonable,
  title={The unreasonable effectiveness of deep features as a perceptual metric},
  author={Zhang, Richard and Isola, Phillip and Efros, Alexei A and Shechtman, Eli and Wang, Oliver},
  booktitle={CVPR},
  year={2018}
}

@inproceedings{jalal2021robust,
  title={Robust compressed sensing {MRI} with deep generative priors},
  author={Jalal, Ajil and Arvinte, Marius and Daras, Giannis and Price, Eric and Dimakis, Alexandros G and Tamir, Jon},
  booktitle={NeurIPS},
  year={2021}
}

@inproceedings{sohl2015deep,
  title={Deep unsupervised learning using nonequilibrium thermodynamics},
  author={Sohl-Dickstein, Jascha and Weiss, Eric and Maheswaranathan, Niru and Ganguli, Surya},
  booktitle={ICML},
  year={2015}
}

@inproceedings{ho2020denoising,
  title={Denoising diffusion probabilistic models},
  author={Ho, Jonathan and Jain, Ajay and Abbeel, Pieter},
  booktitle={NeurIPS},
  year={2020}
}

@inproceedings{song2020denoising,
  title     = {Denoising diffusion implicit models},
  author    = {Jiaming Song and Chenlin Meng and Stefano Ermon},
  booktitle = {ICLR},
  year      = {2021}
}

@article{lu2022dpmp,
  title={D{PM-S}olver++: Fast solver for guided sampling of diffusion probabilistic models},
  author={Lu, Cheng and Zhou, Yuhao and Bao, Fan and Chen, Jianfei and Li, Chongxuan and Zhu, Jun},
  journal={arXiv preprint arXiv:2211.01095},
  year={2022}
}

@inproceedings{lu2022dpm,
  title={D{PM-S}olver: A fast {ODE} solver for diffusion probabilistic model sampling in around 10 steps},
  author={Lu, Cheng and Zhou, Yuhao and Bao, Fan and Chen, Jianfei and Li, Chongxuan and Zhu, Jun},
  booktitle={NeurIPS},
  year={2022}
}

@inproceedings{zhao2024unipc,
  title={Uni{PC}: {A} unified predictor-corrector framework for fast sampling of diffusion models},
  author={Zhao, Wenliang and Bai, Lujia and Rao, Yongming and Zhou, Jie and Lu, Jiwen},
  booktitle={NeurIPS},
  year={2023}
}

@inproceedings{gao2023implicit,
  title={Implicit diffusion models for continuous super-resolution},
  author={Gao, Sicheng and Liu, Xuhui and Zeng, Bohan and Xu, Sheng and Li, Yanjing and Luo, Xiaoyan and Liu, Jianzhuang and Zhen, Xiantong and Zhang, Baochang},
  booktitle={CVPR},
  year={2023}
}

@inproceedings{shang2024resdiff,
  title={Res{D}iff: Combining {CNN} and diffusion model for image super-resolution},
  author={Shang, Shuyao and Shan, Zhengyang and Liu, Guangxing and Wang, LunQian and Wang, XingHua and Zhang, Zekai and Zhang, Jinglin},
  booktitle={AAAI},
  year={2024}
}

@inproceedings{lugmayr2022repaint,
  title={Repaint: Inpainting using denoising diffusion probabilistic models},
  author={Lugmayr, Andreas and Danelljan, Martin and Romero, Andres and Yu, Fisher and Timofte, Radu and Van Gool, Luc},
  booktitle={CVPR},
  year={2022}
}

@inproceedings{sanghvi2025kernel,
  title={Kernel diffusion: An alternate approach to blind deconvolution},
  author={Sanghvi, Yash and Chi, Yiheng and Chan, Stanley H},
  booktitle={ECCV},
  year={2025},
}

@inproceedings{song2021score,
  title={Score-Based generative modeling through stochastic differential equations},
  author={Song, Yang and Sohl-Dickstein, Jascha and Kingma, Diederik P and Kumar, Abhishek and Ermon, Stefano and Poole, Ben},
  booktitle={ICLR},
  year={2021}
}

@inproceedings{blau2018perception,
  title={The perception-distortion tradeoff},
  author={Blau, Yochai and Michaeli, Tomer},
  booktitle={CVPR},
  year={2018}
}

@inproceedings{karras2019style,
  title={A style-based generator architecture for generative adversarial networks},
  author={Karras, Tero and Laine, Samuli and Aila, Timo},
  booktitle={CVPR},
  year={2019}
}

@inproceedings{deng2009imagenet,
  title={{I}mage{N}et: {A} large-scale hierarchical image database},
  author={Deng, Jia and Dong, Wei and Socher, Richard and Li, Li-Jia and Li, Kai and Fei-Fei, Li},
  booktitle={CVPR},
  year={2009},
}

@inproceedings{liu2015faceattributes,
  title = {Deep learning face attributes in the wild},
  author = {Liu, Ziwei and Luo, Ping and Wang, Xiaogang and Tang, Xiaoou},
  booktitle = {ICCV},
  year = {2015} 
}

@inproceedings{song2021solving,
  title={Solving inverse problems in medical imaging with score-based generative models},
  author={Song, Yang and Shen, Liyue and Xing, Lei and Ermon, Stefano},
  booktitle={ICLR},
  year={2022}
}

@inproceedings{wu2019deep,
  title={Deep compressed sensing},
  author={Wu, Yan and Rosca, Mihaela and Lillicrap, Timothy},
  booktitle={ICML},
  year={2019}
}

@article{candes2005decoding,
  title={Decoding by linear programming},
  author={Cand{\`e}s, Emmanuel J and Tao, Terence},
  journal={IEEE Transactions on Information Theory},
  year={2005},
}

@book{twomey2019introduction,
  title={Introduction to the mathematics of inversion in remote sensing and indirect measurements},
  author={Twomey, Sean},
  year={2019},
  publisher={Courier Dover Publications}
}

@inproceedings{wang2022zero,
  title={Zero-shot image restoration using denoising diffusion null-space model},
  author={Wang, Yinhuai and Yu, Jiwen and Zhang, Jian},
  booktitle={ICLR},
  year={2023}
}

@inproceedings{dou2024diffusion,
  title={Diffusion posterior sampling for linear inverse problem solving: A filtering perspective},
  author={Dou, Zehao and Song, Yang},
  booktitle={ICLR},
  year={2024}
}

@inproceedings{song2024solving,
  title={Solving inverse problems with latent diffusion models via hard data consistency},
  author={Song, Bowen and Kwon, Soo Min and Zhang, Zecheng and Hu, Xinyu and Qu, Qing and Shen, Liyue},
  booktitle={ICLR},
  year={2024}
}

@inproceedings{tran2021explore,
  title={Explore image deblurring via encoded blur kernel space},
  author={Tran, Phong and Tran, Anh Tuan and Phung, Quynh and Hoai, Minh},
  booktitle={CVPR},
  year={2021}
}

@inproceedings{heusel2017gans,
  title={{GAN}s trained by a two time-scale update rule converge to a local {N}ash equilibrium},
  author={Heusel, Martin and Ramsauer, Hubert and Unterthiner, Thomas and Nessler, Bernhard and Hochreiter, Sepp},
  booktitle={NeurIPS},
  year={2017}
}

@article{chen2025robust,
  title={Robust learning of diffusion models with extremely noisy conditions},
  author={Chen, Xin and Dobbie, Gillian and Wang, Xinyu and Liu, Feng and Wang, Di and Zhang, Jingfeng},
  journal={arXiv preprint arXiv:2510.10149},
  year={2025}
}

@inproceedings{moliner2023solving,
  title={Solving audio inverse problems with a diffusion model},
  author={Moliner, Eloi and Lehtinen, Jaakko and V{\"a}lim{\"a}ki, Vesa},
  booktitle={ICASSP},
  year={2023},
}

@inproceedings{saito2023unsupervised,
  title={Unsupervised vocal dereverberation with diffusion-based generative models},
  author={Saito, Koichi and Murata, Naoki and Uesaka, Toshimitsu and Lai, Chieh-Hsin and Takida, Yuhta and Fukui, Takao and Mitsufuji, Yuki},
  booktitle={ICASSP},
  year={2023},
}

@article{chung2022mr,
  title={{MR} image denoising and super-resolution using regularized reverse diffusion},
  author={Chung, Hyungjin and Lee, Eun Sun and Ye, Jong Chul},
  journal={IEEE Transactions on Medical Imaging},
  year={2022},
}

@inproceedings{ren2023multiscale,
  title={Multiscale structure guided diffusion for image deblurring},
  author={Ren, Mengwei and Delbracio, Mauricio and Talebi, Hossein and Gerig, Guido and Milanfar, Peyman},
  booktitle={CVPR},
  year={2023}
}

@inproceedings{corneanu2024latentpaint,
  title={Latent{P}aint: {I}mage inpainting in latent space with diffusion models},
  author={Corneanu, Ciprian and Gadde, Raghudeep and Martinez, Aleix M},
  booktitle={CVPR},
  year={2024}
}

@inproceedings{karras2022elucidating,
  title={Elucidating the design space of diffusion-based generative models},
  author={Karras, Tero and Aittala, Miika and Aila, Timo and Laine, Samuli},
  booktitle={NeurIPS},
  year={2022}
}

@inproceedings{kingma2021variational,
  title={Variational diffusion models},
  author={Kingma, Diederik and Salimans, Tim and Poole, Ben and Ho, Jonathan},
  booktitle={NeurIPS},
  year={2021}
}

@inproceedings{janati2024divide,
  title={Divide-and-conquer posterior sampling for denoising diffusion priors},
  author={Janati, Yazid and Moufad, Badr and Durmus, Alain and Moulines, Eric and Olsson, Jimmy},
  booktitle={NeurIPS},
  year={2024}
}

@inproceedings{moufad2024variational,
  title={Variational diffusion posterior sampling with midpoint guidance},
  author={Moufad, Badr and Janati, Yazid and Bedin, Lisa and Durmus, Alain and Douc, Randal and Moulines, Eric and Olsson, Jimmy},
  booktitle={ICLR},
  year={2025}
}

@article{li2024decoupled,
  title={Decoupled data consistency with diffusion purification for image restoration},
  author={Li, Xiang and Kwon, Soo Min and Liang, Shijun and Alkhouri, Ismail R and Ravishankar, Saiprasad and Qu, Qing},
  journal={arXiv preprint arXiv:2403.06054},
  year={2024}
}

@inproceedings{dou2025hybrid, 
title={Hybird regularization improves diffusion- based inverse problem solving}, author={Dou, Hongkun and Li, Zeyu and Du, Jinyang and Yang, Lijun and Yao, Wen and Deng, Yue},
booktitle={ICLR},
year={2025}, 
}

@article{liu2019high,
  title={High dimensional robust {$M$}-estimation: {A}rbitrary corruption and heavy tails},
  author={Liu, Liu and Li, Tianyang and Caramanis, Constantine},
  journal={arXiv preprint arXiv:1901.08237},
  year={2019}
}

@inproceedings{liu2020high,
  title={High dimensional robust sparse regression},
  author={Liu, Liu and Shen, Yanyao and Li, Tianyang and Caramanis, Constantine},
  booktitle={ICAIS},
  year={2020},
}

@article{dong2021deep,
  title={Deep outlier handling for image deblurring},
  author={Dong, Jiangxin and Pan, Jinshan},
  journal={IEEE Transactions on Image Processing},
  year={2021},
  publisher={IEEE}
}

@inproceedings{chen2020oid,
  title={{OID}: {O}utlier identifying and discarding in blind image deblurring},
  author={Chen, Liang and Fang, Faming and Zhang, Jiawei and Liu, Jun and Zhang, Guixu},
  booktitle={ECCV},
  year={2020},
}

@inproceedings{dalalyan2019outlier,
  title={{O}utlier-robust estimation of a sparse linear model using $\ell_1 $-penalized {H}uber's {$M$}-estimator},
  author={Dalalyan, Arnak and Thompson, Philip},
  booktitle={NeurIPS},
  year={2019}
}

@inproceedings{cho2011handling,
  title={Handling outliers in non-blind image deconvolution},
  author={Cho, Sunghyun and Wang, Jue and Lee, Seungyong},
  booktitle={ICCV},
  year={2011},
}

@inproceedings{wang2020further,
  title={Further analysis of outlier detection with deep generative models},
  author={Wang, Ziyu and Dai, Bin and Wipf, David and Zhu, Jun},
  booktitle={NeurIPS},
  year={2020}
}

@article{loh2015regularized,
  title={Regularized {M}-estimators with nonconvexity: {S}tatistical and algorithmic theory for local optima},
  author={Loh, Po-Ling and Wainwright, Martin J},
  year={2015},
}

@inproceedings{laforgue2021generalization,
  title={Generalization bounds in the presence of outliers: {A} median-of-means study},
  author={Laforgue, Pierre and Staerman, Guillaume and Cl{\'e}men{\c{c}}on, Stephan},
  booktitle={ICML},
  year={2021},
}

@inproceedings{d2021consistent,
  title={Consistent regression when oblivious outliers overwhelm},
  author={d’Orsi, Tommaso and Novikov, Gleb and Steurer, David},
  booktitle={ICML},
  year={2021},
}

@inproceedings{shen2019learning,
  title={Learning with bad training data via iterative trimmed loss minimization},
  author={Shen, Yanyao and Sanghavi, Sujay},
  booktitle={ICML},
  year={2019},
}

@inproceedings{martens2010deep,
  title={Deep learning via hessian-free optimization.},
  author={Martens, James and others},
  booktitle={ICML},
  year={2010}
}

@article{jiang2019improved,
  title={Improved {F}letcher--{R}eeves and {D}ai--{Y}uan conjugate gradient methods with the strong {W}olfe line search},
  author={Jiang, Xianzhen and Jian, Jinbao},
  journal={Journal of Computational and Applied Mathematics},
  year={2019},
}

@book{kelley1995iterative,
  title={Iterative methods for linear and nonlinear equations},
  author={Kelley, Carl T},
  year={1995},
  publisher={SIAM}
}

@inproceedings{shoushtari2025unsupervised,
  title={Unsupervised detection of distribution shift in inverse problems using diffusion models},
  author={Shoushtari, Shirin and Chandler, Edward P and Wang, Yuanhao and Asif, M Salman and Kamilov, Ulugbek S},
  booktitle={arXiv preprint arXiv:2505.11482},
  year={2025}
}

@inproceedings{kummerle2021iteratively,
  title={Iteratively reweighted least squares for basis pursuit with global linear convergence rate},
  author={K{\"u}mmerle, Christian and Mayrink Verdun, Claudio and St{\"o}ger, Dominik},
  booktitle={NeurIPS},
  year={2021}
}

@book{nocedal2006numerical,
  title={Numerical optimization},
  author={Nocedal, Jorge and Wright, Stephen J},
  year={2006},
  publisher={Springer}
}

@inproceedings{li2026integrating,
  title={Integrating Reweighted Least Squares with Plug-and-Play Diffusion Priors for Noisy Image Restoration},
  author={Li, Ji and Wang, Chao},
  booktitle={AAAI},
  year={2026}
}

@article{yang2026outlier,
  title={Outlier-robust Diffusion Posterior Sampling for Bayesian Inverse Problems},
  author={Yang, Yiming and Cheng, Xiaoyuan and He, Yi and Li, Kaiyu and Yuan, Wenxuan and Sun, Zhuo},
  journal={arXiv preprint arXiv:2602.02045},
  year={2026}
}
}


\clearpage
\onecolumn
\appendix
\setcounter{page}{1}
{
\centering
\Large
\textbf{\thetitle}\\
\vspace{0.5em}Supplementary Material \\
\vspace{1.0em}
}

\section{Comparison with IRLS-PnPDP}

Both our methods and the recent work, IRLS-PnPDP~\cite{li2026integrating}, employ the well-established iteratively reweighted least squares (IRLS) strategy to address outlier problems in IPs. Our methods leverage this strategy to mitigate outliers within the Huber loss framework, which differentially penalizes small and large residuals to limit the influence of outliers while preserving all measurement information. IRLS-PnPDP uses the strategy to solve an $\ell_q$-norm minimization problem derived from a generalized Gaussian scale mixture model. While both approaches achieve robustness through iterative reweighting, the Huber loss offers a smooth transition between quadratic and linear penalties controlled by a single threshold parameter, whereas the $\ell_q$-norm with $q < 1$ encourages sparser solutions at the cost of introducing non-convexity. Additionally, we provide a conjugate gradient method as an alternative to gradient descent, avoiding the need for delicate learning rate tuning via an efficient line search strategy.

\section{More experimental results}
In this section, we first present the detailed setup of the hyperparameter set in our proposed Robust-GD and Robust-CG methods for different inverse problems in Section~\ref{app:parameter_setup}. We then present the performance of our methods compared to other approaches on the FFHQ dataset in Section~\ref{app:exp_ffhq}. We also validate the effect of the Huber loss threshold parameter $\delta$ and the learning rate $\eta_{x}$ in Robust-GD, as well as the use of $\bg^{\rm T}\bg$ versus $\bg^{\rm T}\bd$ for calculating $\alpha$ in Robust-CG, in Section~\ref{app:exp_abl}.

\subsection{Parameter Setup}
\label{app:parameter_setup}
We present the detailed setup of the hyperparameters used for Robust-GD and Robust-CG in Table~\ref{app:tab:parameter}. Here, $N$ denotes the total number of sample steps and $J$ denotes the total number of iterations. $\delta$ is the Huber loss threshold parameter, $\eta_{x}$ is the learning rate used in Robust-GD, and $\eta$ is the finite-difference approximation parameter used in Robust-CG.

\begin{table}[!htbp]
\begin{center}
\renewcommand{\arraystretch}{0.9}
\setlength{\tabcolsep}{0.85mm}{
\begin{tabular}{c c c c c c c c c c c}
\hline
&\multicolumn{2}{c}{\scriptsize{\textbf{Super-resolution ($4\times$)}}}
&\multicolumn{2}{c}{\scriptsize{\textbf{Inpainting (random $70\%$)}}}
&\multicolumn{2}{c}{\scriptsize{\textbf{Gaussian Deblurring}}}
&\multicolumn{2}{c}{\scriptsize{\textbf{Motion Deblurring}}}
&\multicolumn{2}{c}{\scriptsize{\textbf{Nonlinear Deblurring}}}
\\
\cline{2-11}
\scriptsize{\textbf{Hyperparameter}}
&\scriptsize{\textbf{Robust-GD}}
&\scriptsize{\textbf{Robust-CG}}
&\scriptsize{\textbf{Robust-GD}}
&\scriptsize{\textbf{Robust-CG}}
&\scriptsize{\textbf{Robust-GD}}
&\scriptsize{\textbf{Robust-CG}}
&\scriptsize{\textbf{Robust-GD}}
&\scriptsize{\textbf{Robust-CG}}
&\scriptsize{\textbf{Robust-GD}}
&\scriptsize{\textbf{Robust-CG}}
\\
\cline{1-11}
\scriptsize{$N$}
&\scriptsize{200}%
&\scriptsize{200}%
&\scriptsize{200}%
&\scriptsize{200}%
&\scriptsize{200}%
&\scriptsize{200}%
&\scriptsize{200}%
&\scriptsize{200}%
&\scriptsize{200}%
&\scriptsize{200}%
\\
\scriptsize{$J$}
&\scriptsize{100}%
&\scriptsize{20}%
&\scriptsize{100}%
&\scriptsize{100}%
&\scriptsize{100}%
&\scriptsize{20}%
&\scriptsize{100}%
&\scriptsize{20}%
&\scriptsize{100}%
&\scriptsize{50}%
\\
\scriptsize{$\delta$} 
&\scriptsize{0.020}%
&\scriptsize{0.005}%
&\scriptsize{0.010}%
&\scriptsize{0.020}%
&\scriptsize{0.020}%
&\scriptsize{0.020}%
&\scriptsize{0.020}%
&\scriptsize{0.020}%
&\scriptsize{0.010}%
&\scriptsize{0.010}%

\\
\scriptsize{$\eta_{x}$} 
&\scriptsize{0.0001}%
&\scriptsize{/}%
&\scriptsize{0.0001}%
&\scriptsize{/}%
&\scriptsize{0.0001}%
&\scriptsize{/}%
&\scriptsize{0.00005}%
&\scriptsize{/}%
&\scriptsize{0.00005}%
&\scriptsize{/}%
\\
\scriptsize{$\eta$} 
&\scriptsize{/}%
&\scriptsize{0.0001}%
&\scriptsize{/}%
&\scriptsize{0.0001}%
&\scriptsize{/}%
&\scriptsize{0.0001}%
&\scriptsize{/}%
&\scriptsize{0.0001}%
&\scriptsize{/}%
&\scriptsize{0.0001}%
\\
\hline
\end{tabular}
}
\end{center}
\vspace{-1.5em}
\caption{Detailed setup of the hyperparameter set in our proposed Robust-GD and Robust-CG methods.}
\vspace{-1.25em}
\label{app:tab:parameter}
\end{table}

\subsection{Experimental results on FFHQ}
\label{app:exp_ffhq}
We validate the performance of our proposed Robust-GD and Robust-CG methods on the FFHQ dataset for five tasks, namely super-resolution ($4\times$), inpainting (random $70\%$), Gaussian deblurring, motion deblurring, and nonlinear deblurring. The experimental settings for all tasks are the same as those in Section~\ref{sec:exp}. The results are presented in Table~\ref{app:linear_ips} for linear IPs and Table~\ref{app:non} for the nonlinear deblurring task. The results indicate that our methods perform the best across nearly all metrics for all tasks compared to recent DM-based methods.

\begin{table*}[htbp]
\begin{center}
\renewcommand{\arraystretch}{0.9}
\resizebox{\textwidth}{!}{
\setlength{\tabcolsep}{0.85mm}{
\begin{tabular}{c c c c c c c c c c c c c c c c c c}
\hline
\multicolumn{2}{c}{}
&\multicolumn{4}{c}{\scriptsize{\textbf{Super-resolution ($4\times$)}}}
&\multicolumn{4}{c}{\scriptsize{\textbf{Inpainting (random $70\%$)}}}
&\multicolumn{4}{c}{\scriptsize{\textbf{Gaussian Deblurring}}}
&\multicolumn{4}{c}{\scriptsize{\textbf{Motion Deblurring}}}
\\
\cline{3-18}
\scriptsize{$\rho$}
&\scriptsize{Methods}
&\scriptsize{PSNR$\uparrow$}
&\scriptsize{SSIM$\uparrow$}
&\scriptsize{LPIPS$\downarrow$}
&\scriptsize{FID$\downarrow$}
&\scriptsize{PSNR$\uparrow$}
&\scriptsize{SSIM$\uparrow$}
&\scriptsize{LPIPS$\downarrow$}
&\scriptsize{FID$\downarrow$}
&\scriptsize{PSNR$\uparrow$}
&\scriptsize{SSIM$\uparrow$}
&\scriptsize{LPIPS$\downarrow$}
&\scriptsize{FID$\downarrow$}
&\scriptsize{PSNR$\uparrow$}
&\scriptsize{SSIM$\uparrow$}
&\scriptsize{LPIPS$\downarrow$}
&\scriptsize{FID$\downarrow$}
\\
\cline{1-18}
\multirow{7}{*}{\scriptsize{0.02}}
&\scriptsize{DPS}
&\scriptsize{22.53 \tiny{$\pm$ 1.79}}%
&\scriptsize{0.612 \tiny{$\pm$ 0.078}}%
&\scriptsize{0.228 \tiny{$\pm$ 0.062}}%
&\scriptsize{95.41}%
&\scriptsize{25.26 \tiny{$\pm$ 1.71}}%
&\scriptsize{0.719 \tiny{$\pm$ 0.060}}%
&\scriptsize{0.182 \tiny{$\pm$ 0.049}}%
&\scriptsize{86.95}%
&\scriptsize{23.76 \tiny{$\pm$ 1.91}}%
&\scriptsize{0.643 \tiny{$\pm$ 0.076}}%
&\scriptsize{\textbf{0.183} \tiny{$\pm$ 0.052}}%
&\scriptsize{\underline{85.60}}%
&\scriptsize{22.00 \tiny{$\pm$ 1.79}}%
&\scriptsize{0.587 \tiny{$\pm$ 0.081}}%
&\scriptsize{0.220 \tiny{$\pm$ 0.058}}%
&\scriptsize{93.82}%
\\
&\scriptsize{DiffPIR}
&\scriptsize{19.98 \tiny{$\pm$ 1.50}}%
&\scriptsize{0.525 \tiny{$\pm$ 0.043}}%
&\scriptsize{0.509 \tiny{$\pm$ 0.068}}%
&\scriptsize{263.99}%
&\scriptsize{25.03 \tiny{$\pm$ 1.39}}%
&\scriptsize{0.691 \tiny{$\pm$ 0.032}}%
&\scriptsize{0.310 \tiny{$\pm$ 0.074}}%
&\scriptsize{146.31}%
&\scriptsize{21.49 \tiny{$\pm$ 0.97}}%
&\scriptsize{0.443 \tiny{$\pm$ 0.062}}%
&\scriptsize{0.573 \tiny{$\pm$ 0.084}}%
&\scriptsize{229.75}%
&\scriptsize{19.13 \tiny{$\pm$ 0.82}}%
&\scriptsize{0.322 \tiny{$\pm$ 0.062}}%
&\scriptsize{0.642 \tiny{$\pm$ 0.082}}%
&\scriptsize{284.05}%
\\
&\scriptsize{DCPS}
&\scriptsize{20.39 \tiny{$\pm$ 2.21}}%
&\scriptsize{0.612 \tiny{$\pm$ 0.084}}%
&\scriptsize{0.348 \tiny{$\pm$ 0.097}}%
&\scriptsize{173.42}%
&\scriptsize{\underline{29.31} \tiny{$\pm$ 2.03}}%
&\scriptsize{\textbf{0.849} \tiny{$\pm$ 0.039}}%
&\scriptsize{0.145 \tiny{$\pm$ 0.048}}%
&\scriptsize{81.91}%
&\scriptsize{21.45 \tiny{$\pm$ 2.05}}%
&\scriptsize{0.548 \tiny{$\pm$ 0.086}}%
&\scriptsize{0.297 \tiny{$\pm$ 0.080}}%
&\scriptsize{158.54}%
&\scriptsize{20.82 \tiny{$\pm$ 3.68}}%
&\scriptsize{0.596 \tiny{$\pm$ 0.150}}%
&\scriptsize{0.315 \tiny{$\pm$ 0.158}}%
&\scriptsize{131.96}%
\\
&\scriptsize{RED-diff} 
&\scriptsize{21.73 \tiny{$\pm$ 1.52}}%
&\scriptsize{0.666 \tiny{$\pm$ 0.031}}%
&\scriptsize{0.480 \tiny{$\pm$ 0.066}}%
&\scriptsize{215.67}%
&\scriptsize{23.86 \tiny{$\pm$ 1.29}}%
&\scriptsize{0.646 \tiny{$\pm$ 0.027}}%
&\scriptsize{0.392 \tiny{$\pm$ 0.072}}%
&\scriptsize{160.64}%
&\scriptsize{26.50 \tiny{$\pm$ 2.02}}%
&\scriptsize{0.727 \tiny{$\pm$ 0.053}}%
&\scriptsize{0.318 \tiny{$\pm$ 0.071}}%
&\scriptsize{103.89}%
&\scriptsize{23.42 \tiny{$\pm$ 1.75}}%
&\scriptsize{0.546 \tiny{$\pm$ 0.060}}%
&\scriptsize{0.410 \tiny{$\pm$ 0.076}}%
&\scriptsize{157.78}%
\\
&\scriptsize{DAPS} 
&\scriptsize{21.92 \tiny{$\pm$ 1.69}}%
&\scriptsize{0.689 \tiny{$\pm$ 0.031}}%
&\scriptsize{0.440 \tiny{$\pm$ 0.075}}%
&\scriptsize{264.15}%
&\scriptsize{24.46 \tiny{$\pm$ 1.67}}%
&\scriptsize{0.694 \tiny{$\pm$ 0.036}}%
&\scriptsize{0.308 \tiny{$\pm$ 0.087}}%
&\scriptsize{203.10}%
&\scriptsize{22.83 \tiny{$\pm$ 2.27}}%
&\scriptsize{0.565 \tiny{$\pm$ 0.086}}%
&\scriptsize{0.467 \tiny{$\pm$ 0.080}}%
&\scriptsize{235.16}%
&\scriptsize{22.35 \tiny{$\pm$ 2.47}}%
&\scriptsize{0.585 \tiny{$\pm$ 0.104}}%
&\scriptsize{0.374 \tiny{$\pm$ 0.118}}%
&\scriptsize{230.03}%
\\
&\scriptsize{\textbf{Robust-GD}} 
&\scriptsize{\underline{27.83} \tiny{$\pm$ 1.93}}%
&\scriptsize{\underline{0.780} \tiny{$\pm$ 0.039}}%
&\scriptsize{\underline{0.226} \tiny{$\pm$ 0.060}}%
&\scriptsize{\underline{91.25}}%
&\scriptsize{28.95 \tiny{$\pm$ 1.77}}%
&\scriptsize{0.808 \tiny{$\pm$ 0.022}}%
&\scriptsize{\underline{0.087} \tiny{$\pm$ 0.023}}%
&\scriptsize{\underline{67.89}}%
&\scriptsize{\textbf{28.63} \tiny{$\pm$ 2.24}}%
&\scriptsize{\textbf{0.812} \tiny{$\pm$ 0.045}}%
&\scriptsize{\underline{0.200} \tiny{$\pm$ 0.056}}%
&\scriptsize{\textbf{78.54}}%
&\scriptsize{\textbf{28.57} \tiny{$\pm$ 2.21}}%
&\scriptsize{\textbf{0.795} \tiny{$\pm$ 0.046}}%
&\scriptsize{\textbf{0.131} \tiny{$\pm$ 0.051}}%
&\scriptsize{\textbf{68.53}}%
\\
&\scriptsize{\textbf{Robust-CG}} 
&\scriptsize{\textbf{28.16} \tiny{$\pm$ 2.21}}%
&\scriptsize{\textbf{0.814} \tiny{$\pm$ 0.045}}%
&\scriptsize{\textbf{0.186} \tiny{$\pm$ 0.050}}%
&\scriptsize{\textbf{75.58}}%
&\scriptsize{\textbf{29.51} \tiny{$\pm$ 2.22}}%
&\scriptsize{\underline{0.847} \tiny{$\pm$ 0.023}}%
&\scriptsize{\textbf{0.070} \tiny{$\pm$ 0.017}}%
&\scriptsize{\textbf{53.15}}%
&\scriptsize{\underline{28.04} \tiny{$\pm$ 2.35}}%
&\scriptsize{\underline{0.804} \tiny{$\pm$ 0.054}}%
&\scriptsize{0.232 \tiny{$\pm$ 0.065}}%
&\scriptsize{87.66}%
&\scriptsize{\underline{27.79} \tiny{$\pm$ 2.61}}%
&\scriptsize{\underline{0.783} \tiny{$\pm$ 0.062}}%
&\scriptsize{\underline{0.162} \tiny{$\pm$ 0.062}}%
&\scriptsize{\underline{77.63}}%
\\
\cline{1-18}
\multirow{7}{*}{\scriptsize{0.10}}
&\scriptsize{DPS}
&\scriptsize{19.89 \tiny{$\pm$ 1.78}}%
&\scriptsize{0.560 \tiny{$\pm$ 0.086}}%
&\scriptsize{\underline{0.262} \tiny{$\pm$ 0.070}}%
&\scriptsize{147.83}%
&\scriptsize{21.36 \tiny{$\pm$ 1.66}}%
&\scriptsize{0.641 \tiny{$\pm$ 0.087}}%
&\scriptsize{0.226 \tiny{$\pm$ 0.076}}%
&\scriptsize{152.70}%
&\scriptsize{21.07 \tiny{$\pm$ 1.55}}%
&\scriptsize{0.602 \tiny{$\pm$ 0.080}}%
&\scriptsize{\underline{0.214} \tiny{$\pm$ 0.056}}%
&\scriptsize{91.95}%
&\scriptsize{19.79 \tiny{$\pm$ 1.52}}%
&\scriptsize{0.542 \tiny{$\pm$ 0.085}}%
&\scriptsize{0.250 \tiny{$\pm$ 0.059}}%
&\scriptsize{100.22}%
\\
&\scriptsize{DiffPIR}
&\scriptsize{13.79 \tiny{$\pm$ 1.43}}%
&\scriptsize{0.244 \tiny{$\pm$ 0.065}}%
&\scriptsize{0.705 \tiny{$\pm$ 0.067}}%
&\scriptsize{335.80}%
&\scriptsize{19.25 \tiny{$\pm$ 1.50}}%
&\scriptsize{0.432 \tiny{$\pm$ 0.076}}%
&\scriptsize{0.599 \tiny{$\pm$ 0.113}}%
&\scriptsize{191.14}%
&\scriptsize{17.30 \tiny{$\pm$ 0.86}}%
&\scriptsize{0.310 \tiny{$\pm$ 0.064}}%
&\scriptsize{0.685 \tiny{$\pm$ 0.070}}%
&\scriptsize{332.72}%
&\scriptsize{15.98 \tiny{$\pm$ 0.73}}%
&\scriptsize{0.221 \tiny{$\pm$ 0.057}}%
&\scriptsize{0.744 \tiny{$\pm$ 0.076}}%
&\scriptsize{329.52}%
\\
&\scriptsize{DCPS}
&\scriptsize{14.96 \tiny{$\pm$ 1.31}}%
&\scriptsize{0.364 \tiny{$\pm$ 0.082}}%
&\scriptsize{0.590 \tiny{$\pm$ 0.076}}%
&\scriptsize{271.28}%
&\scriptsize{\underline{23.58} \tiny{$\pm$ 1.59}}%
&\scriptsize{\underline{0.787} \tiny{$\pm$ 0.048}}%
&\scriptsize{\underline{0.186} \tiny{$\pm$ 0.053}}%
&\scriptsize{\underline{95.24}}%
&\scriptsize{13.94 \tiny{$\pm$ 1.51}}%
&\scriptsize{0.246 \tiny{$\pm$ 0.100}}%
&\scriptsize{0.621 \tiny{$\pm$ 0.093}}%
&\scriptsize{313.75}%
&\scriptsize{11.65 \tiny{$\pm$ 2.81}}%
&\scriptsize{0.155 \tiny{$\pm$ 0.157}}%
&\scriptsize{0.920 \tiny{$\pm$ 0.250}}%
&\scriptsize{332.83}%
\\
&\scriptsize{RED-diff} 
&\scriptsize{15.83 \tiny{$\pm$ 1.61}}%
&\scriptsize{0.436 \tiny{$\pm$ 0.055}}%
&\scriptsize{0.659 \tiny{$\pm$ 0.073}}%
&\scriptsize{281.44}%
&\scriptsize{19.36 \tiny{$\pm$ 1.42}}%
&\scriptsize{0.426 \tiny{$\pm$ 0.066}}%
&\scriptsize{0.631 \tiny{$\pm$ 0.107}}%
&\scriptsize{186.58}%
&\scriptsize{22.17 \tiny{$\pm$ 1.57}}%
&\scriptsize{0.600 \tiny{$\pm$ 0.067}}%
&\scriptsize{0.440 \tiny{$\pm$ 0.082}}%
&\scriptsize{158.14}%
&\scriptsize{17.02 \tiny{$\pm$ 2.06}}%
&\scriptsize{0.296 \tiny{$\pm$ 0.088}}%
&\scriptsize{0.828 \tiny{$\pm$ 0.143}}%
&\scriptsize{335.15}%
\\
&\scriptsize{DAPS} 
&\scriptsize{15.60 \tiny{$\pm$ 1.94}}%
&\scriptsize{0.395 \tiny{$\pm$ 0.075}}%
&\scriptsize{0.650 \tiny{$\pm$ 0.082}}%
&\scriptsize{319.75}%
&\scriptsize{18.20 \tiny{$\pm$ 1.67}}%
&\scriptsize{0.427 \tiny{$\pm$ 0.082}}%
&\scriptsize{0.615 \tiny{$\pm$ 0.115}}%
&\scriptsize{282.18}%
&\scriptsize{14.52 \tiny{$\pm$ 2.17}}%
&\scriptsize{0.287 \tiny{$\pm$ 0.103}}%
&\scriptsize{0.654 \tiny{$\pm$ 0.081}}%
&\scriptsize{348.48}%
&\scriptsize{13.92 \tiny{$\pm$ 2.21}}%
&\scriptsize{0.263 \tiny{$\pm$ 0.116}}%
&\scriptsize{0.766 \tiny{$\pm$ 0.140}}%
&\scriptsize{375.09}%
\\
&\scriptsize{\textbf{Robust-GD}} 
&\scriptsize{\underline{24.15} \tiny{$\pm$ 1.36}}%
&\scriptsize{\underline{0.648} \tiny{$\pm$ 0.040}}%
&\scriptsize{0.417 \tiny{$\pm$ 0.080}}%
&\scriptsize{\underline{141.06}}%
&\scriptsize{20.27 \tiny{$\pm$ 1.08}}%
&\scriptsize{0.494 \tiny{$\pm$ 0.060}}%
&\scriptsize{0.500 \tiny{$\pm$ 0.088}}%
&\scriptsize{197.86}%
&\scriptsize{\textbf{28.37} \tiny{$\pm$ 2.17}}%
&\scriptsize{\textbf{0.804} \tiny{$\pm$ 0.045}}%
&\scriptsize{\textbf{0.209} \tiny{$\pm$ 0.056}}%
&\scriptsize{\textbf{82.02}}%
&\scriptsize{\underline{26.36} \tiny{$\pm$ 2.11}}%
&\scriptsize{\underline{0.710} \tiny{$\pm$ 0.045}}%
&\scriptsize{\textbf{0.153} \tiny{$\pm$ 0.050}}%
&\scriptsize{\textbf{70.98}}%
\\
&\scriptsize{\textbf{Robust-CG}} 
&\scriptsize{\textbf{27.47} \tiny{$\pm$ 2.17}}%
&\scriptsize{\textbf{0.800} \tiny{$\pm$ 0.046}}%
&\scriptsize{\textbf{0.194} \tiny{$\pm$ 0.052}}%
&\scriptsize{\textbf{76.75}}%
&\scriptsize{\textbf{28.24} \tiny{$\pm$ 1.95}}%
&\scriptsize{\textbf{0.799} \tiny{$\pm$ 0.022}}%
&\scriptsize{\textbf{0.087} \tiny{$\pm$ 0.022}}%
&\scriptsize{\textbf{59.16}}%
&\scriptsize{\underline{27.96} \tiny{$\pm$ 2.32}}%
&\scriptsize{\underline{0.802} \tiny{$\pm$ 0.054}}%
&\scriptsize{0.234 \tiny{$\pm$ 0.065}}%
&\scriptsize{\underline{86.58}}%
&\scriptsize{\textbf{27.51} \tiny{$\pm$ 2.42}}%
&\scriptsize{\textbf{0.767} \tiny{$\pm$ 0.060}}%
&\scriptsize{\underline{0.168} \tiny{$\pm$ 0.059}}%
&\scriptsize{\underline{81.94}}%
\\
\hline
\end{tabular}
}
}
\end{center}
\vspace{-1.5em}
\caption{(\textbf{Linear IPs}) \textbf{Super-resolution ($4\times$)}, \textbf{inpainting (random $70\%$)}, \textbf{Gaussian deblurring} and \textbf{motion deblurring} with additive Gaussian noise ($\sigma = 0.05$) and contamination fraction $\rho = 0.02$ or $0.10$.}
\vspace{-1.25em}
\label{app:linear_ips}
\end{table*}

\begin{table}[!htbp]
\begin{center}
\renewcommand{\arraystretch}{0.9}
\setlength{\tabcolsep}{0.85mm}{
\begin{tabular}{c c c c c c c c c }
\hline
\multicolumn{1}{c}{}
&\multicolumn{4}{c}{\scriptsize{\textbf{$\rho = 0.02$}}}
&\multicolumn{4}{c}{\scriptsize{\textbf{$\rho = 0.10$}}}
\\
\cline{2-9}
\scriptsize{Methods}
&\scriptsize{PSNR$\uparrow$}
&\scriptsize{SSIM$\uparrow$}
&\scriptsize{LPIPS$\downarrow$}
&\scriptsize{FID$\downarrow$}
&\scriptsize{PSNR$\uparrow$}
&\scriptsize{SSIM$\uparrow$}
&\scriptsize{LPIPS$\downarrow$}
&\scriptsize{FID$\downarrow$}
\\
\cline{1-9}
\scriptsize{DPS}
\scriptsize{DPS}
&\scriptsize{22.64 \tiny{$\pm$ 1.96}}%
&\scriptsize{0.607 \tiny{$\pm$ 0.089}}%
&\scriptsize{0.227 \tiny{$\pm$ 0.066}}%
&\scriptsize{\underline{94.15}}%
&\scriptsize{20.50 \tiny{$\pm$ 1.58}}%
&\scriptsize{0.575 \tiny{$\pm$ 0.085}}%
&\scriptsize{0.242 \tiny{$\pm$ 0.064}}%
&\scriptsize{\underline{94.60}}%
\\
\scriptsize{RED-diff}
&\scriptsize{21.93 \tiny{$\pm$ 1.29}}%
&\scriptsize{0.487 \tiny{$\pm$ 0.052}}%
&\scriptsize{0.563 \tiny{$\pm$ 0.101}}%
&\scriptsize{155.26}%
&\scriptsize{15.98 \tiny{$\pm$ 1.56}}%
&\scriptsize{0.263 \tiny{$\pm$ 0.071}}%
&\scriptsize{0.923 \tiny{$\pm$ 0.136}}%
&\scriptsize{274.57}%
\\
\scriptsize{DAPS} 
&\scriptsize{20.63 \tiny{$\pm$ 1.56}}%
&\scriptsize{0.447 \tiny{$\pm$ 0.077}}%
&\scriptsize{0.635 \tiny{$\pm$ 0.161}}%
&\scriptsize{300.53}%
&\scriptsize{15.40 \tiny{$\pm$ 1.62}}%
&\scriptsize{0.217 \tiny{$\pm$ 0.068}}%
&\scriptsize{1.044 \tiny{$\pm$ 0.123}}%
&\scriptsize{418.32}%
\\
\scriptsize{\textbf{Robust-GD}} 
&\scriptsize{\textbf{27.57} \tiny{$\pm$ 1.66}}%
&\scriptsize{\textbf{0.768} \tiny{$\pm$ 0.038}}%
&\scriptsize{\textbf{0.137} \tiny{$\pm$ 0.040}}%
&\scriptsize{\textbf{71.36}}%
&\scriptsize{\textbf{26.36} \tiny{$\pm$ 1.44}}%
&\scriptsize{\underline{0.710} \tiny{$\pm$ 0.036}}%
&\scriptsize{\textbf{0.153} \tiny{$\pm$ 0.043}}%
&\scriptsize{\textbf{77.31}}%
\\
\scriptsize{\textbf{Robust-CG}} 
&\scriptsize{\underline{25.90} \tiny{$\pm$ 1.86}}%
&\scriptsize{\underline{0.724} \tiny{$\pm$ 0.056}}%
&\scriptsize{\underline{0.210} \tiny{$\pm$ 0.062}}%
&\scriptsize{95.05}%
&\scriptsize{\underline{25.83} \tiny{$\pm$ 1.80}}%
&\scriptsize{\textbf{0.719} \tiny{$\pm$ 0.052}}%
&\scriptsize{\underline{0.207} \tiny{$\pm$ 0.051}}%
&\scriptsize{95.47}%
\\
\hline
\end{tabular}
}
\end{center}
\vspace{-1.5em}
\caption{\textbf{Nonlinear deblurring} with additive Gaussian noise ($\sigma = 0.05$) and contamination fraction $\rho = 0.02$ or $0.10$.}
\vspace{-1.25em}
\label{app:non}
\end{table}

\subsection{Ablation study}
\label{app:exp_abl}
We validate the effect of the Huber loss threshold parameter $\delta$ and the learning rate $\eta_{x}$ on the Robust-GD method. Additionally, we compare the use of the $\bg^{\rmT}\bg$ term versus the $\bg^{\rmT}\bd$ term for calculating $\alpha$ in Robust-CG (refer to Eq.~\eqref{eq:cg_gg}).
\subsubsection{Huber loss threshold parameter}
To evaluate the effect of the Huber loss threshold parameter $\delta$, we use the same experimental setting as described in Section~\ref{sec:abl_huber}. We tested the Robust-GD method on 100 validation images from the CelebA dataset for the Gaussian and motion deblurring tasks, using a contamination factor of $\rho = 0.10$ and a Gaussian noise level of $\sigma = 0.05$. We chose the value of $\delta$ from the set $\{0.005, 0.01, 0.02, 0.04\}$. The results presented in Table~\ref{app:weight} indicate that Robust-GD performs similarly across these different threshold values, thereby demonstrating the robustness of our methods to the selection of the threshold parameter.

\begin{table}[!htbp]
\begin{center}
\renewcommand{\arraystretch}{0.9}
\setlength{\tabcolsep}{0.85mm}{
\begin{tabular}{c c c c c c c}
\hline
\multicolumn{1}{c}{}
&\multicolumn{3}{c}{\scriptsize{\textbf{Gaussian Deblurring}}}
&\multicolumn{3}{c}{\scriptsize{\textbf{Motion Deblurring}}}
\\
\cline{2-7}
\scriptsize{Threshold $\delta$}
&\scriptsize{PSNR$\uparrow$}
&\scriptsize{SSIM$\uparrow$}
&\scriptsize{LPIPS$\downarrow$}
&\scriptsize{PSNR$\uparrow$}
&\scriptsize{SSIM$\uparrow$}
&\scriptsize{LPIPS$\downarrow$}
\\
\cline{1-7}
\scriptsize{0.005}
&\scriptsize{29.62}%
&\scriptsize{0.826}%
&\scriptsize{0.141}%
&\scriptsize{28.42}%
&\scriptsize{0.796}%
&\scriptsize{0.152}%
\\
\scriptsize{0.010}
&\scriptsize{29.80}%
&\scriptsize{0.827}%
&\scriptsize{0.136}%
&\scriptsize{29.33}%
&\scriptsize{0.806}%
&\scriptsize{0.128}%
\\
\scriptsize{0.020}
&\scriptsize{29.27}%
&\scriptsize{0.817}%
&\scriptsize{0.138}%
&\scriptsize{29.44}%
&\scriptsize{0.789}%
&\scriptsize{0.123}%
\\
\scriptsize{0.040} 
&\scriptsize{29.21}%
&\scriptsize{0.790}%
&\scriptsize{0.160}%
&\scriptsize{28.52}%
&\scriptsize{0.734}%
&\scriptsize{0.156}%
\\
\hline
\end{tabular}
}
\end{center}
\vspace{-1.5em}
\caption{Performance of Robust-CG with different value of the Huber loss threshold $\delta$ on \textbf{Gaussian deblurring} and \textbf{motion deblurring} with additive Gaussian noise ($\sigma = 0.05$) and contamination fraction $0.10$.}
\vspace{-1.25em}
\label{app:weight}
\end{table}

\subsubsection{Learning rate in Robust-GD}
We validate the effect of the learning rate $\eta_{x}$ in the Robust-GD method on five tasks, namely super-resolution ($4\times$), inpainting (random $70\%$), Gaussian deblurring, motion deblurring and nonlinear deblurring tasks. We use 100 validation images from the CelebA dataset with a contamination factor of $\rho = 0.10$ and a Gaussian noise level of $\sigma = 0.05$. We choose five values of $\eta_{x}$ from the set $\{0.0010, 0.0005, 0.0001, 0.00005, 0.00001\}$. The results presented in Table~\ref{app:ablation_eta_x} indicate that the performance of Robust-GD on five tasks is sensitive to the choice of $\eta_{x}$, thereby motivating the proposition of Robust-CG to avoid the delicate fine-tuning of the learning rate required by Robust-GD.

\begin{table}[!htbp]
\begin{center}
\renewcommand{\arraystretch}{0.9}
\resizebox{\textwidth}{!}{
\setlength{\tabcolsep}{0.85mm}{
\begin{tabular}{c c c c c c c c c c c c c c c c}
\hline
\multicolumn{1}{c}{}
&\multicolumn{3}{c}{\scriptsize{\textbf{Super-resolution ($4\times$)}}}
&\multicolumn{3}{c}{\scriptsize{\textbf{Inpainting (random $70\%$)}}}
&\multicolumn{3}{c}{\scriptsize{\textbf{Gaussian Deblurring}}}
&\multicolumn{3}{c}{\scriptsize{\textbf{Motion Deblurring}}}
&\multicolumn{3}{c}{\scriptsize{\textbf{Nonlinear Deblurring}}}
\\
\cline{2-16}
\scriptsize{Learning Rate $\eta_{x}$}
&\scriptsize{PSNR$\uparrow$}
&\scriptsize{SSIM$\uparrow$}
&\scriptsize{LPIPS$\downarrow$}
&\scriptsize{PSNR$\uparrow$}
&\scriptsize{SSIM$\uparrow$}
&\scriptsize{LPIPS$\downarrow$}
&\scriptsize{PSNR$\uparrow$}
&\scriptsize{SSIM$\uparrow$}
&\scriptsize{LPIPS$\downarrow$}
&\scriptsize{PSNR$\uparrow$}
&\scriptsize{SSIM$\uparrow$}
&\scriptsize{LPIPS$\downarrow$}
&\scriptsize{PSNR$\uparrow$}
&\scriptsize{SSIM$\uparrow$}
&\scriptsize{LPIPS$\downarrow$}
\\
\cline{1-16}
\scriptsize{$0.0010$}
&\scriptsize{16.20}%
&\scriptsize{0.413}%
&\scriptsize{0.625}%
&\scriptsize{19.16}%
&\scriptsize{0.438}%
&\scriptsize{0.585}%
&\scriptsize{25.32}%
&\scriptsize{0.697}%
&\scriptsize{0.215}%
&\scriptsize{19.75}%
&\scriptsize{0.357}%
&\scriptsize{0.646}%
&\scriptsize{14.44}%
&\scriptsize{0.173}%
&\scriptsize{1.160}%
\\
\scriptsize{$0.0005$}
&\scriptsize{16.60}%
&\scriptsize{0.429}%
&\scriptsize{0.613}%
&\scriptsize{19.18}%
&\scriptsize{0.445}%
&\scriptsize{0.583}%
&\scriptsize{28.49}%
&\scriptsize{0.756}%
&\scriptsize{0.182}%
&\scriptsize{22.11}%
&\scriptsize{0.431}%
&\scriptsize{0.541}%
&\scriptsize{14.36}%
&\scriptsize{0.174}%
&\scriptsize{1.181}%
\\
\scriptsize{$0.0001$}
&\scriptsize{26.52}%
&\scriptsize{0.716}%
&\scriptsize{0.264}%
&\scriptsize{23.28}%
&\scriptsize{0.575}%
&\scriptsize{0.374}%
&\scriptsize{29.27}%
&\scriptsize{0.817}%
&\scriptsize{0.138}%
&\scriptsize{28.53}%
&\scriptsize{0.734}%
&\scriptsize{0.154}%
&\scriptsize{24.52}%
&\scriptsize{0.582}%
&\scriptsize{0.269}%
\\
\scriptsize{$0.00005$}
&\scriptsize{28.29}%
&\scriptsize{0.799}%
&\scriptsize{0.145}%
&\scriptsize{28.29}%
&\scriptsize{0.750}%
&\scriptsize{0.132}%
&\scriptsize{29.82}%
&\scriptsize{0.826}%
&\scriptsize{0.136}%
&\scriptsize{29.44}%
&\scriptsize{0.789}%
&\scriptsize{0.123}%
&\scriptsize{27.06}%
&\scriptsize{0.711}%
&\scriptsize{0.156}%
\\
\scriptsize{$0.00001$}
&\scriptsize{15.84}%
&\scriptsize{0.546}%
&\scriptsize{0.405}%
&\scriptsize{23.62}%
&\scriptsize{0.776}%
&\scriptsize{0.176}%
&\scriptsize{29.21}%
&\scriptsize{0.814}%
&\scriptsize{0.154}%
&\scriptsize{28.08}%
&\scriptsize{0.789}%
&\scriptsize{0.160}%
&\scriptsize{22.84}%
&\scriptsize{0.674}%
&\scriptsize{0.225}%
\\
\hline
\end{tabular}
}
}
\end{center}
\vspace{-1.5em}
\caption{Performance of Robust-GD with different value of learning rate $\eta_{x}$ on \textbf{super-resolution}, \textbf{inpainting}, \textbf{Gaussian deblurring}, \textbf{motion deblurring} and \textbf{nonlinear deblurring} with Gaussian noise ($\sigma = 0.05$) and contamination fraction $\rho = 0.10$.}
\vspace{-1.25em}
\label{app:ablation_eta_x}
\end{table}

\subsubsection{Conjugate gradient term in Robust-CG}
We validate the performance of Robust-CG using the $\bg^{\rmT}\bg$ term versus the $\bg^{\rmT}\bd$ term for calculating $\alpha$ (refer to Eq.~\eqref{eq:cg_gg}) on 100 validation images from the CelebA dataset. The evaluation covers five tasks, namely super-resolution ($4\times$), inpainting (random $70\%$), Gaussian deblurring, motion deblurring, and nonlinear deblurring, each with a contamination factor of $\rho = 0.10$ and a Gaussian noise level of $\sigma = 0.05$. The results presented in Table~\ref{app:gg_gp} indicate that Robust-CG using $\bg^{\rmT}\bg$ exhibits more stable and superior performance across all tasks compared to using $\bg^{\rmT}\bd$.

\begin{table*}[htbp]
\begin{center}
\renewcommand{\arraystretch}{0.9}
\resizebox{\textwidth}{!}{
\setlength{\tabcolsep}{0.85mm}{
\begin{tabular}{c c c c c c c c c c c c c c c c}
\hline
\multicolumn{1}{c}{}
&\multicolumn{3}{c}{\scriptsize{\textbf{Super-resolution ($4\times$)}}}
&\multicolumn{3}{c}{\scriptsize{\textbf{Inpainting (random $70\%$)}}}
&\multicolumn{3}{c}{\scriptsize{\textbf{Gaussian Deblurring}}}
&\multicolumn{3}{c}{\scriptsize{\textbf{Motion Deblurring}}}
&\multicolumn{3}{c}{\scriptsize{\textbf{Nonlinear Deblurring}}}
\\
\cline{2-16}
\scriptsize{Methods}
&\scriptsize{PSNR$\uparrow$}
&\scriptsize{SSIM$\uparrow$}
&\scriptsize{LPIPS$\downarrow$}
&\scriptsize{PSNR$\uparrow$}
&\scriptsize{SSIM$\uparrow$}
&\scriptsize{LPIPS$\downarrow$}
&\scriptsize{PSNR$\uparrow$}
&\scriptsize{SSIM$\uparrow$}
&\scriptsize{LPIPS$\downarrow$}
&\scriptsize{PSNR$\uparrow$}
&\scriptsize{SSIM$\uparrow$}
&\scriptsize{LPIPS$\downarrow$}
&\scriptsize{PSNR$\uparrow$}
&\scriptsize{SSIM$\uparrow$}
&\scriptsize{LPIPS$\downarrow$}
\\
\cline{1-16}
\scriptsize{$\bg^{\rmT}\bg$}
&\scriptsize{28.96}%
&\scriptsize{0.819}%
&\scriptsize{0.129}%
&\scriptsize{29.74}%
&\scriptsize{0.809}%
&\scriptsize{0.093}%
&\scriptsize{29.38}%
&\scriptsize{0.819}%
&\scriptsize{0.151}%
&\scriptsize{28.91}%
&\scriptsize{0.777}%
&\scriptsize{0.144}%
&\scriptsize{26.80}%
&\scriptsize{0.728}%
&\scriptsize{0.188}%
\\
\scriptsize{$\bg^{\rmT}\bd$}
&\scriptsize{28.07}%
&\scriptsize{0.792}%
&\scriptsize{0.164}%
&\scriptsize{19.15}%
&\scriptsize{0.446}%
&\scriptsize{0.582}%
&\scriptsize{29.51}%
&\scriptsize{0.803}%
&\scriptsize{0.152}%
&\scriptsize{24.50}%
&\scriptsize{0.530}%
&\scriptsize{0.403}%
&\scriptsize{23.75}%
&\scriptsize{0.526}%
&\scriptsize{0.369}%
\\
\hline
\end{tabular}
}
}
\end{center}
\vspace{-1.5em}
\caption{Performance of Robust-CG using the $\bg^{\rmT}\bg$ term versus the $\bg^{\rmT}\bd$ term for calculating $\alpha$ on \textbf{super-resolution ($4\times$)}, \textbf{inpainting (random $70\%$)}, \textbf{Gaussian deblurring}, \textbf{motion deblurring} and \textbf{nonlinear deblurring} with additive Gaussian noise ($\sigma = 0.05$) and contamination fraction $\rho = 0.10$.}
\vspace{-1.5em}
\label{app:gg_gp}
\end{table*}
\subsection{Ablation of core components}
\begin{figure}[htb]
    \centering
    \begin{subfigure}{0.33\textwidth}
        \centering
        \includegraphics[width=\linewidth]{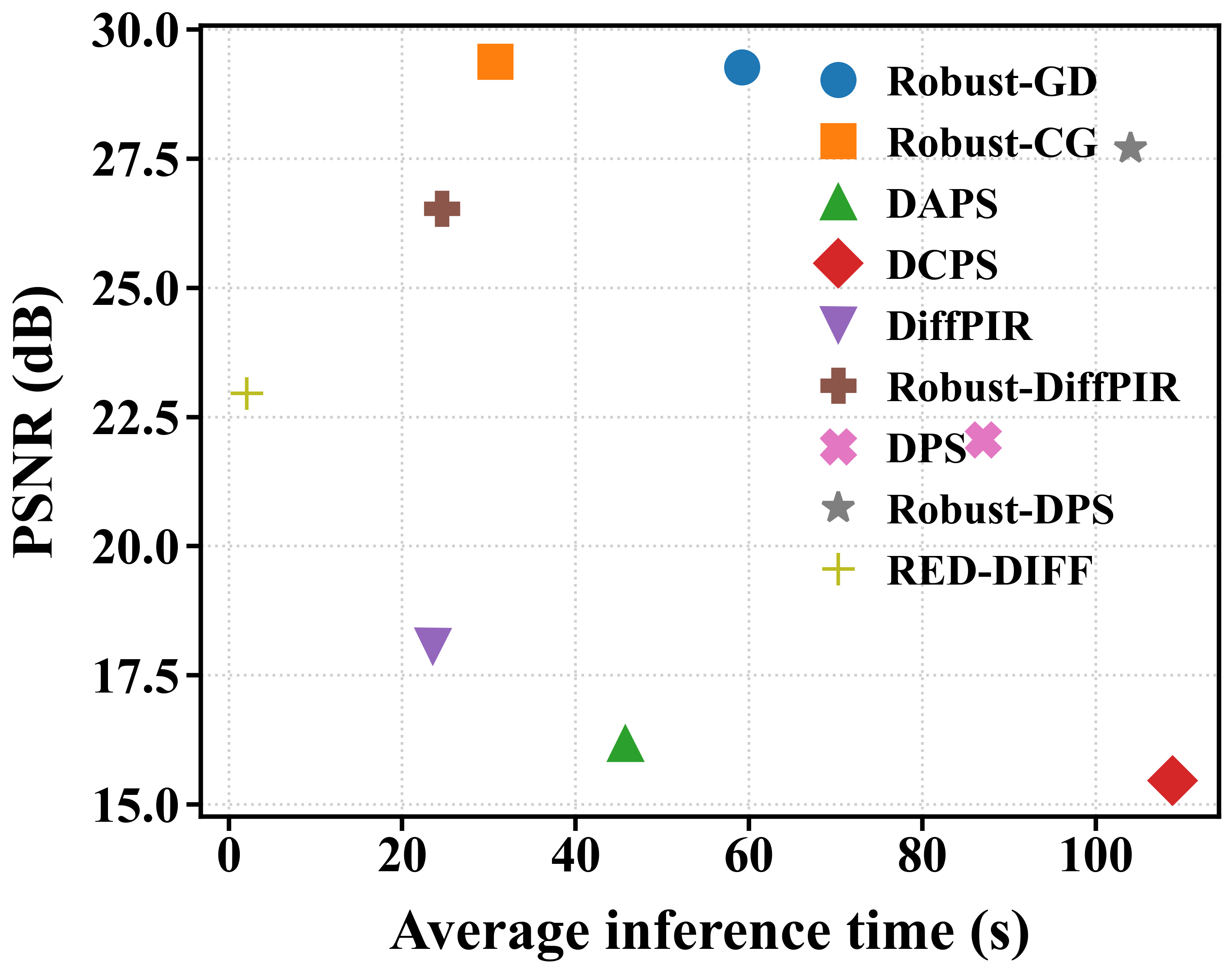}
    \end{subfigure}
    \hfill  
    \begin{subfigure}{0.33\textwidth}
        \centering
        \includegraphics[width=\linewidth]{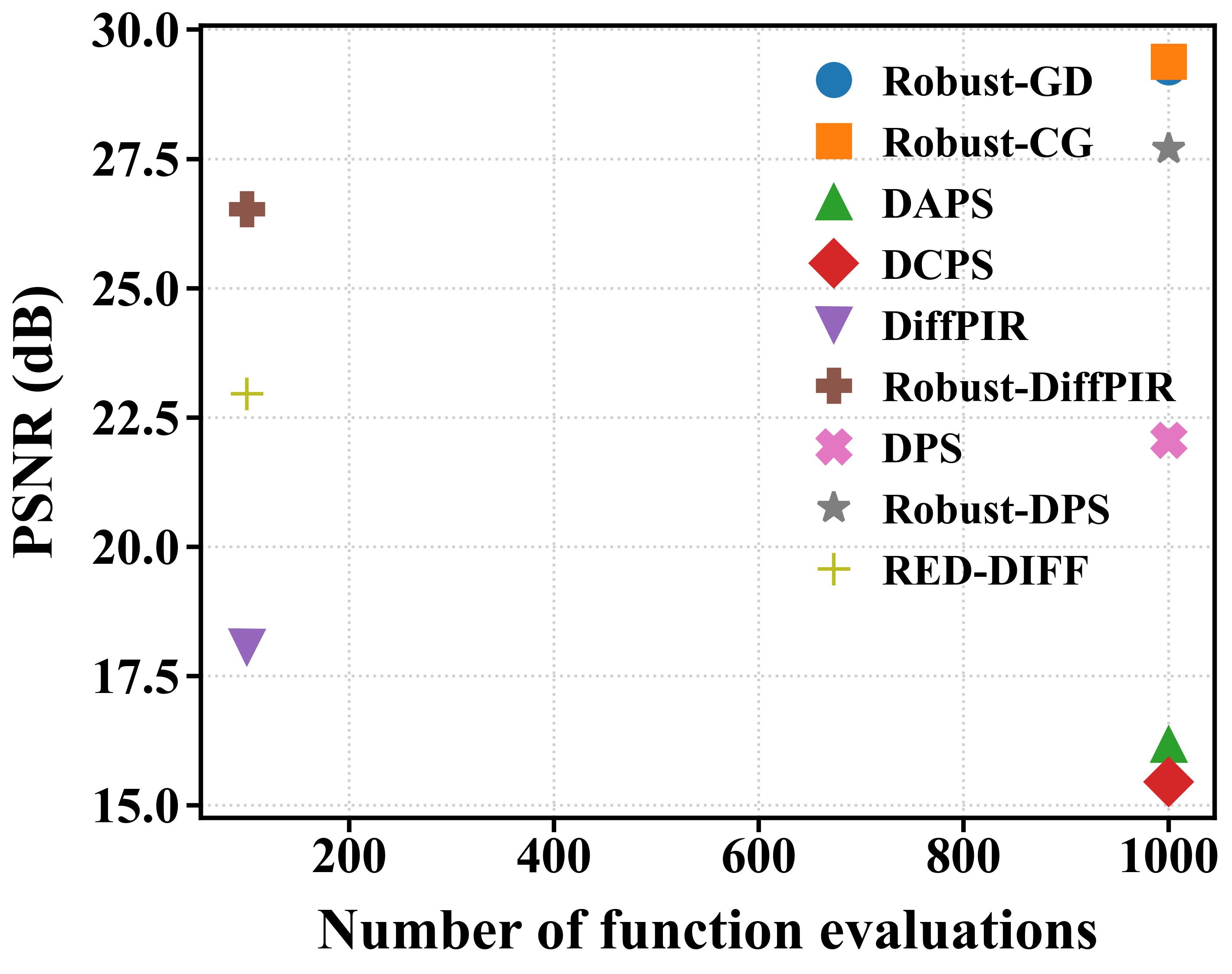}
    \end{subfigure}
    \hfill
    \begin{subfigure}{0.33\textwidth}
        \centering
        \includegraphics[width=\linewidth]{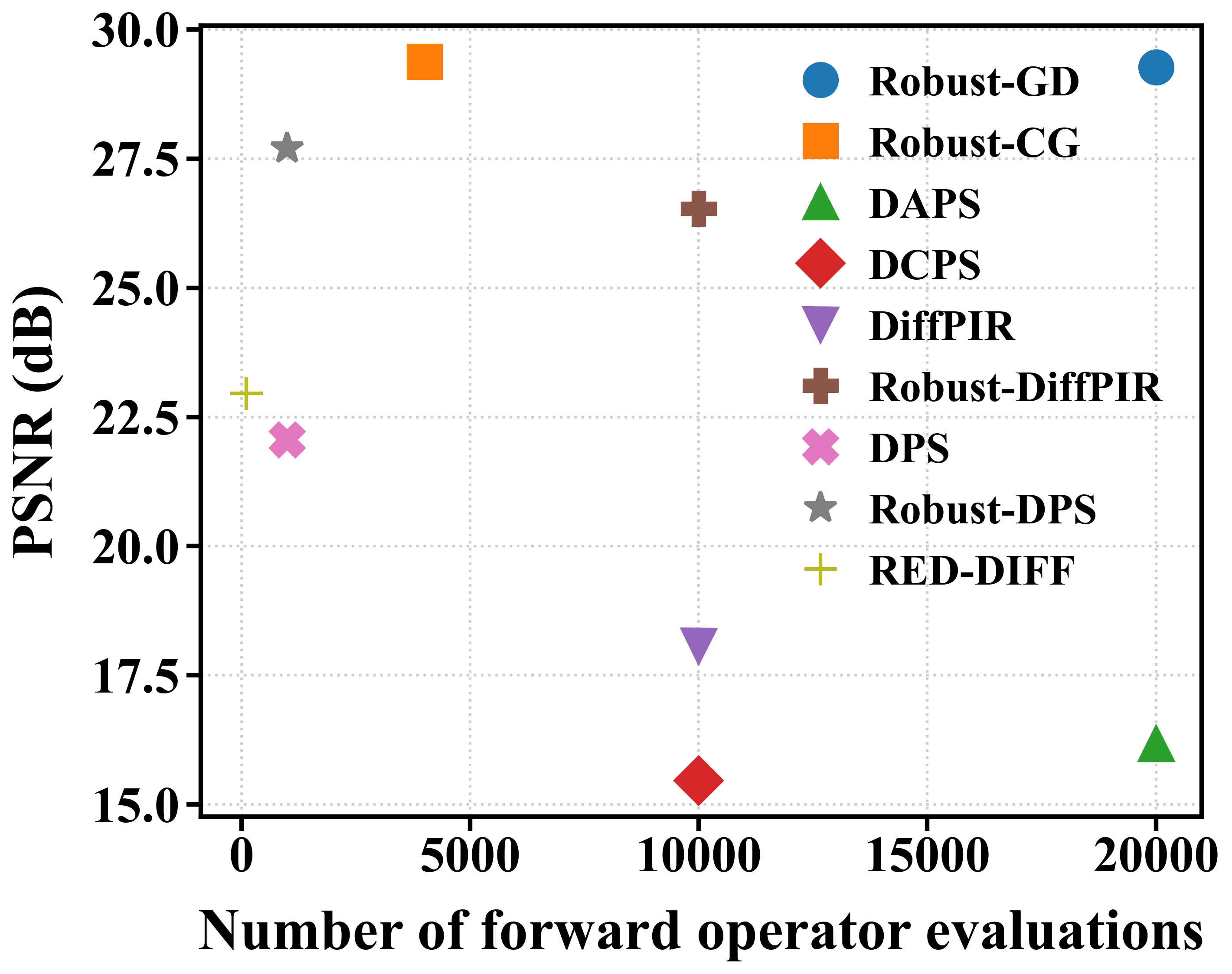}
    \end{subfigure}
    \caption{Visualization of the relationship between the distortion metric PSNR and computational cost for the CelebA Gaussian deblurring task. The computational cost is measured in terms of (left) average inference time over 100 images, (middle) number of function evaluations, and (right) number of forward operator evaluations.} 
    \label{app:fig:time_comparison}
\end{figure}

To validate the effectiveness of measurement refinement and the Huber loss, we incorporate these components into two baseline methods, namely DPS and DiffPIR, and denote the resulting variants as Robust-DPS and Robust-DiffPIR, respectively. The results in Table~\ref{app:tab:linear_deblur} indicate that, with the incorporation of measurement refinement and the Huber loss, both Robust-DPS and Robust-DiffPIR outperform their original counterparts. Additionally, the proposed methods Robust-GD and Robust-CG achieve the best performance across most evaluation metrics, which suggests that their design is more compatible with these components. Furthermore, we visualize the relationship between the distortion metric PSNR and computational cost, including the average inference time over 100 images for the CelebA Gaussian deblurring task, as well as the number of function evaluations and forward operator evaluations for each algorithm. The results in Figure~\ref{app:fig:time_comparison} show that Robust-GD and Robust-CG achieve a favorable trade-off between efficiency and reconstruction performance.

\begin{table*}[ht]
\begin{center}
\renewcommand{\arraystretch}{0.9}
\resizebox{\textwidth}{!}{
\setlength{\tabcolsep}{0.85mm}{
\begin{tabular}{c c c c c c c c c c c c c c c c c}
\hline
\multicolumn{1}{c}{}
&\multicolumn{4}{c}{\scriptsize{\textbf{CelebA ($256 \times 256$)}}}
&\multicolumn{4}{c}{\scriptsize{\textbf{FFHQ ($256 \times 256$)}}}
&\multicolumn{4}{c}{\scriptsize{\textbf{CelebA ($256 \times 256$)}}}
&\multicolumn{4}{c}{\scriptsize{\textbf{FFHQ ($256 \times 256$)}}}
\\
\cmidrule(lr){2-5}
\cmidrule(lr){6-9}
\cmidrule(lr){10-13}
\cmidrule(lr){14-17}
\multicolumn{1}{c}{}
&\scriptsize{PSNR$\uparrow$}
&\scriptsize{SSIM$\uparrow$}
&\scriptsize{LPIPS$\downarrow$}
&\scriptsize{FID$\downarrow$}
&\scriptsize{PSNR$\uparrow$}
&\scriptsize{SSIM$\uparrow$}
&\scriptsize{LPIPS$\downarrow$}
&\scriptsize{FID$\downarrow$}
&\scriptsize{PSNR$\uparrow$}
&\scriptsize{SSIM$\uparrow$}
&\scriptsize{LPIPS$\downarrow$}
&\scriptsize{FID$\downarrow$}
&\scriptsize{PSNR$\uparrow$}
&\scriptsize{SSIM$\uparrow$}
&\scriptsize{LPIPS$\downarrow$}
&\scriptsize{FID$\downarrow$}
\\
\hline
\scriptsize{Methods}
&\multicolumn{8}{c}{\scriptsize{\textbf{Gaussian deblurring}}}
&\multicolumn{8}{c}{\scriptsize{\textbf{Motion deblurring}}}
\\
\cline{1-17}
\scriptsize{DPS}
&\scriptsize{22.06}%
&\scriptsize{0.646}%
&\scriptsize{0.178}%
&\scriptsize{63.49}
&\scriptsize{21.07}%
&\scriptsize{0.602}%
&\scriptsize{0.214}%
&\scriptsize{91.95}
&\scriptsize{20.90}%
&\scriptsize{0.596}%
&\scriptsize{0.208}%
&\scriptsize{\underline{68.96}}
&\scriptsize{19.19}%
&\scriptsize{0.542}%
&\scriptsize{0.250}%
&\scriptsize{100.22}
\\
\scriptsize{\textbf{Robust-DPS}}
&\scriptsize{27.70}%
&\scriptsize{0.765}%
&\scriptsize{\textbf{0.099}}%
&\scriptsize{\textbf{53.82}}%
&\scriptsize{26.23}%
&\scriptsize{0.742}%
&\scriptsize{\textbf{0.115}}%
&\scriptsize{\textbf{62.79}}%
&\scriptsize{27.58}%
&\scriptsize{0.770}%
&\scriptsize{\textbf{0.108}}%
&\scriptsize{\textbf{56.01}}%
&\scriptsize{26.00}%
&\scriptsize{0.744}%
&\scriptsize{\textbf{0.127}}%
&\scriptsize{\textbf{69.63}}%
\\
\scriptsize{DiffPIR}
&\scriptsize{18.07}%
&\scriptsize{0.355}%
&\scriptsize{0.622}%
&\scriptsize{308.93}
&\scriptsize{17.30}%
&\scriptsize{0.310}%
&\scriptsize{0.685}%
&\scriptsize{332.72}
&\scriptsize{17.11}%
&\scriptsize{0.259}%
&\scriptsize{0.712}%
&\scriptsize{340.27}
&\scriptsize{15.98}%
&\scriptsize{0.221}%
&\scriptsize{0.744}%
&\scriptsize{329.52}
\\
\scriptsize{\textbf{Robust-DiffPIR}}
&\scriptsize{26.53}%
&\scriptsize{0.723}%
&\scriptsize{0.166}%
&\scriptsize{78.30}
&\scriptsize{24.80}%
&\scriptsize{0.675}%
&\scriptsize{0.227}%
&\scriptsize{102.18}
&\scriptsize{23.46}%
&\scriptsize{0.621}%
&\scriptsize{0.229}%
&\scriptsize{106.00}
&\scriptsize{21.91}%
&\scriptsize{0.565}%
&\scriptsize{0.285}%
&\scriptsize{132.55}
\\
\scriptsize{DCPS}
&\scriptsize{15.46}%
&\scriptsize{0.302}%
&\scriptsize{0.575}%
&\scriptsize{331.91}
&\scriptsize{13.94}%
&\scriptsize{0.246}%
&\scriptsize{0.621}%
&\scriptsize{313.75}
&\scriptsize{12.60}%
&\scriptsize{0.174}%
&\scriptsize{0.927}%
&\scriptsize{338.22}
&\scriptsize{11.65}%
&\scriptsize{0.155}%
&\scriptsize{0.920}%
&\scriptsize{332.83}
\\
\scriptsize{RED-diff} 
&\scriptsize{22.96}%
&\scriptsize{0.626}%
&\scriptsize{0.391}%
&\scriptsize{139.05}
&\scriptsize{22.17}%
&\scriptsize{0.600}%
&\scriptsize{0.440}%
&\scriptsize{158.14}
&\scriptsize{17.36}%
&\scriptsize{0.311}%
&\scriptsize{0.836}%
&\scriptsize{303.61}
&\scriptsize{17.02}%
&\scriptsize{0.296}%
&\scriptsize{0.828}%
&\scriptsize{335.15}
\\
\scriptsize{DAPS} 
&\scriptsize{16.18}%
&\scriptsize{0.356}%
&\scriptsize{0.561}%
&\scriptsize{297.13}
&\scriptsize{14.52}%
&\scriptsize{0.287}%
&\scriptsize{0.654}%
&\scriptsize{348.48}
&\scriptsize{14.99}%
&\scriptsize{0.331}%
&\scriptsize{0.718}%
&\scriptsize{292.81}
&\scriptsize{13.92}%
&\scriptsize{0.263}%
&\scriptsize{0.766}%
&\scriptsize{375.09}
\\
\scriptsize{\textbf{Robust-GD}} 
&\scriptsize{\underline{29.27}}%
&\scriptsize{\underline{0.817}}%
&\scriptsize{\underline{0.138}}%
&\scriptsize{65.59}
&\scriptsize{\textbf{28.37}}%
&\scriptsize{\textbf{0.804}}%
&\scriptsize{\underline{0.209}}%
&\scriptsize{\underline{82.02}}
&\scriptsize{\textbf{29.44}}%
&\scriptsize{\textbf{0.789}}%
&\scriptsize{\underline{0.123}}%
&\scriptsize{84.79}
&\scriptsize{\underline{26.36}}%
&\scriptsize{\underline{0.710}}%
&\scriptsize{\underline{0.153}}%
&\scriptsize{\underline{70.98}}
\\
\scriptsize{\textbf{Robust-CG}} 
&\scriptsize{\textbf{29.38}}%
&\scriptsize{\textbf{0.819}}%
&\scriptsize{0.151}%
&\scriptsize{\underline{62.52}}
&\scriptsize{\underline{27.96}}%
&\scriptsize{\underline{0.802}}%
&\scriptsize{0.234}%
&\scriptsize{86.58}
&\scriptsize{\underline{28.91}}%
&\scriptsize{\underline{0.777}}%
&\scriptsize{0.144}%
&\scriptsize{91.22}
&\scriptsize{\textbf{27.51}}%
&\scriptsize{\textbf{0.767}}%
&\scriptsize{0.168}%
&\scriptsize{81.94}
\\
\hline
\end{tabular}
}}
\end{center}
\vspace{-1.5em}
\caption{(\textbf{Linear IPs}) \textbf{Gaussian deblurring} and \textbf{motion deblurring} with additive Gaussian noise ($\sigma = 0.05$) and contamination fraction $0.10$ on the CelebA and FFHQ datasets.}
\label{app:tab:linear_deblur}
\vspace{-1.5em}
\end{table*}

\clearpage
\section{Visualization of experimental results}
We visualize our experimental results on five tasks, namely super-resolution ($4\times$), inpainting (random $70\%$), Gaussian deblurring, motion deblurring and nonlinear deblurring tasks across three datasets, namely CelebA, FFHQ and ImageNet. The experiments are conducted with a contamination factor of $\rho = 0.10$ and a Gaussian noise level of $\sigma = 0.05$.

\begin{figure}[htbp]
    \centering
    \includegraphics[width=0.95\linewidth]{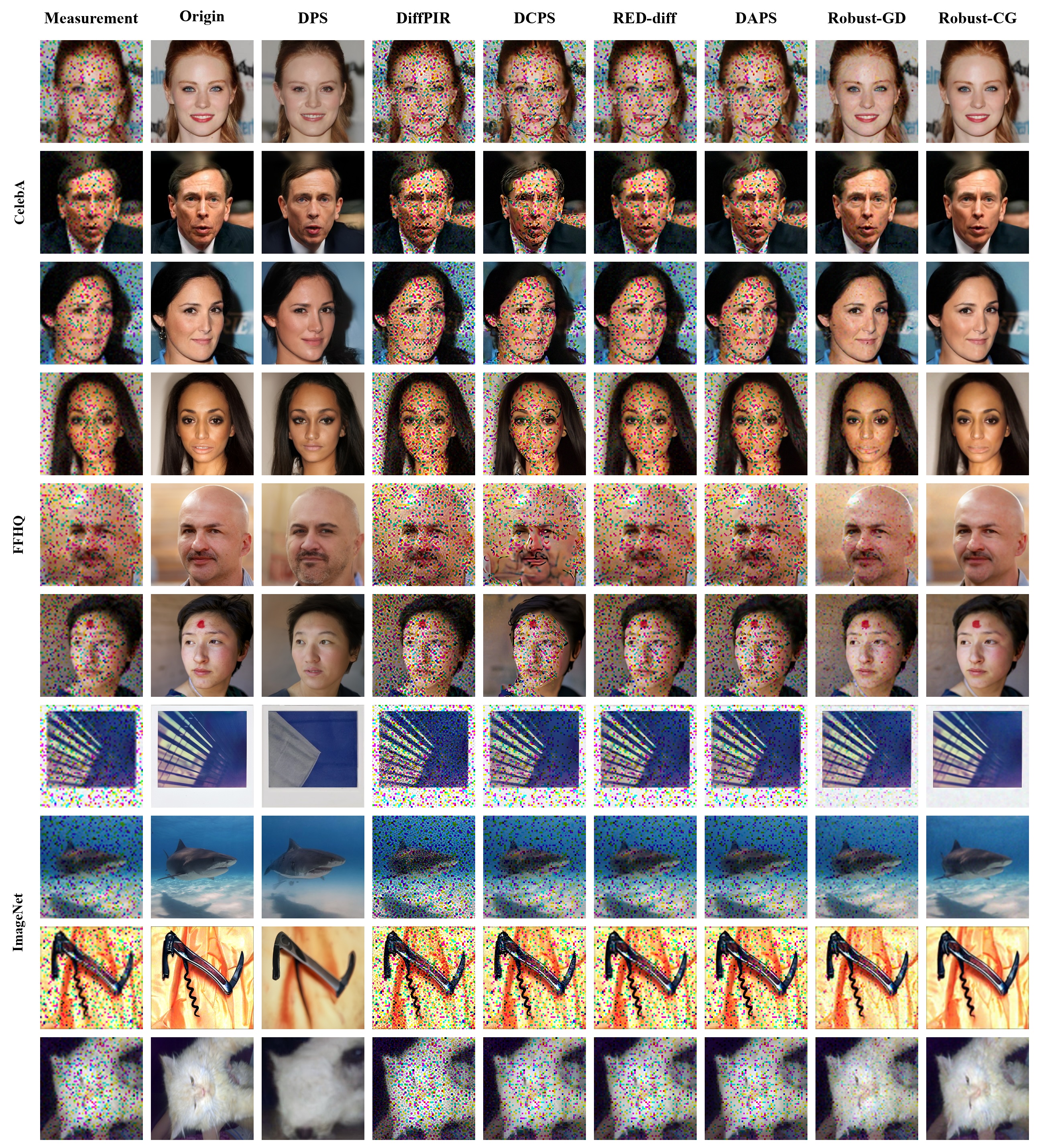}
    \caption{Visualization of the experimental results for the super-resolution ($4\times$) task with a contamination factor of $\rho = 0.10$ and a Gaussian noise level of $\sigma = 0.05$.}
\end{figure}

\begin{figure}[htbp]
    \centering
    \includegraphics[width=0.95\linewidth]{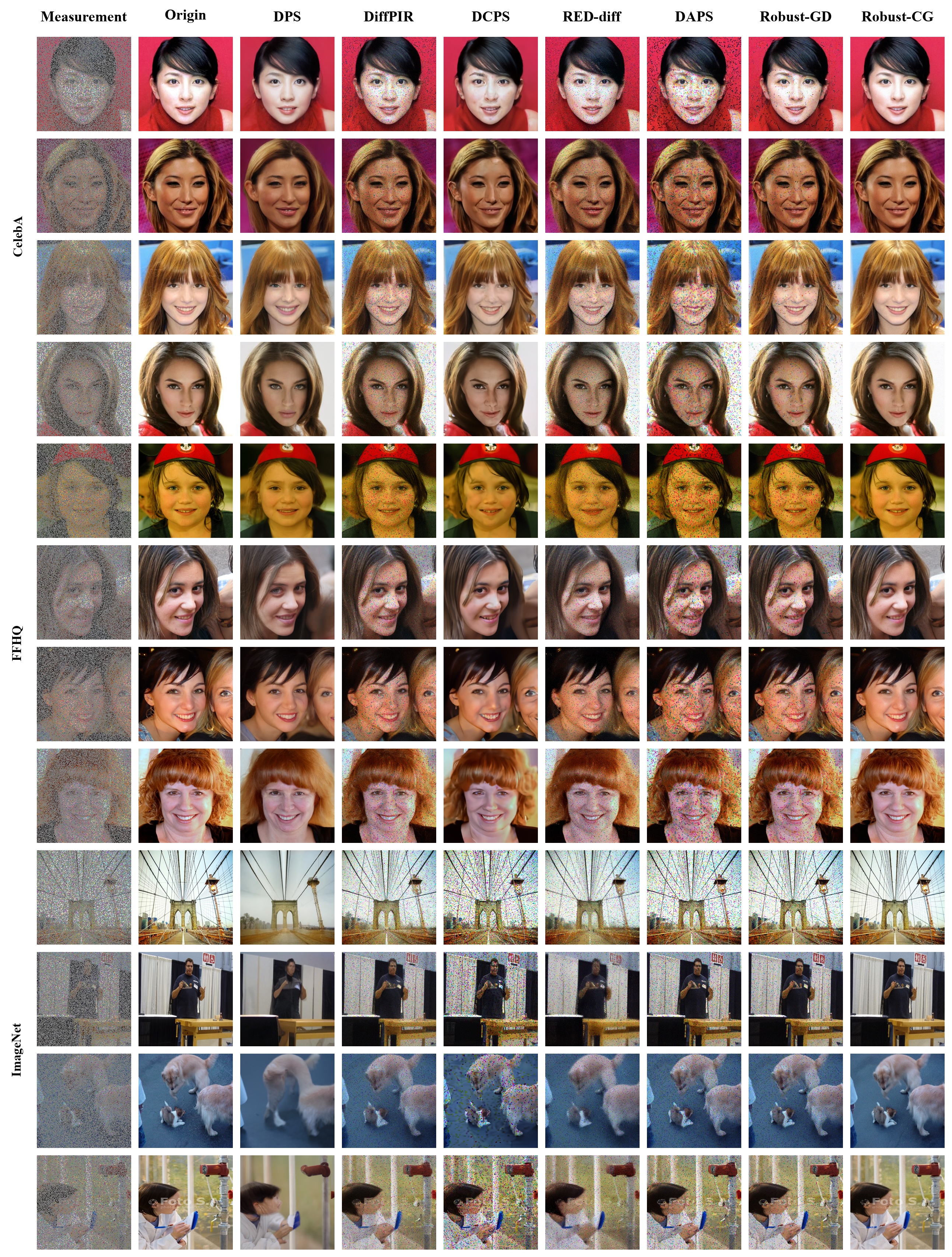}
    \caption{Visualization of the experimental results for the inpainting (random $70\%$) task with a contamination factor of $\rho = 0.10$ and a Gaussian noise level of $\sigma = 0.05$.}
\end{figure}

\begin{figure}[htbp]
    \centering
    \includegraphics[width=0.95\linewidth]{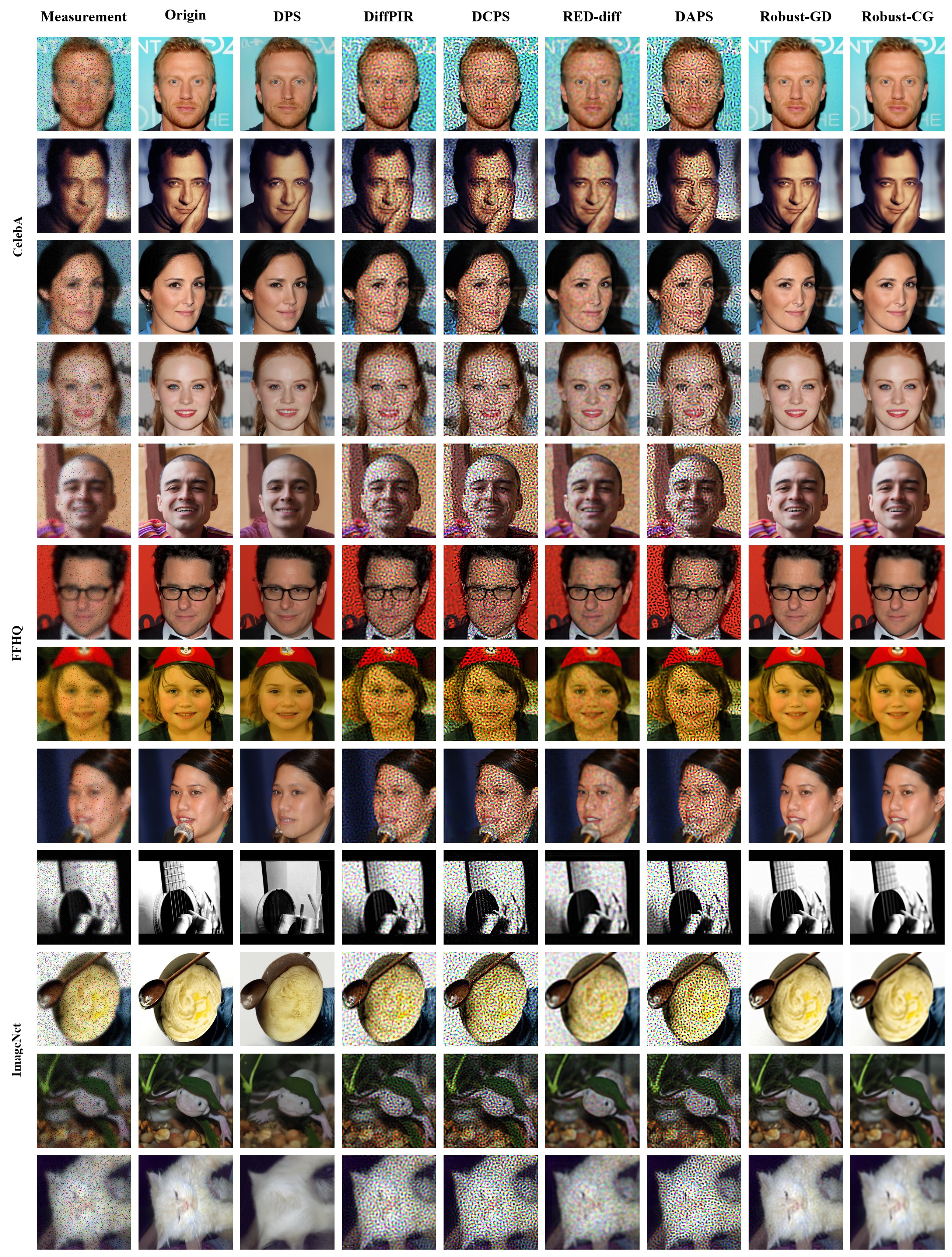}
    \caption{Visualization of the experimental results for the Gaussian deblurring task with a contamination factor of $\rho = 0.10$ and a Gaussian noise level of $\sigma = 0.05$.}
\end{figure}

\begin{figure}[htbp]
    \centering
    \includegraphics[width=0.95\linewidth]{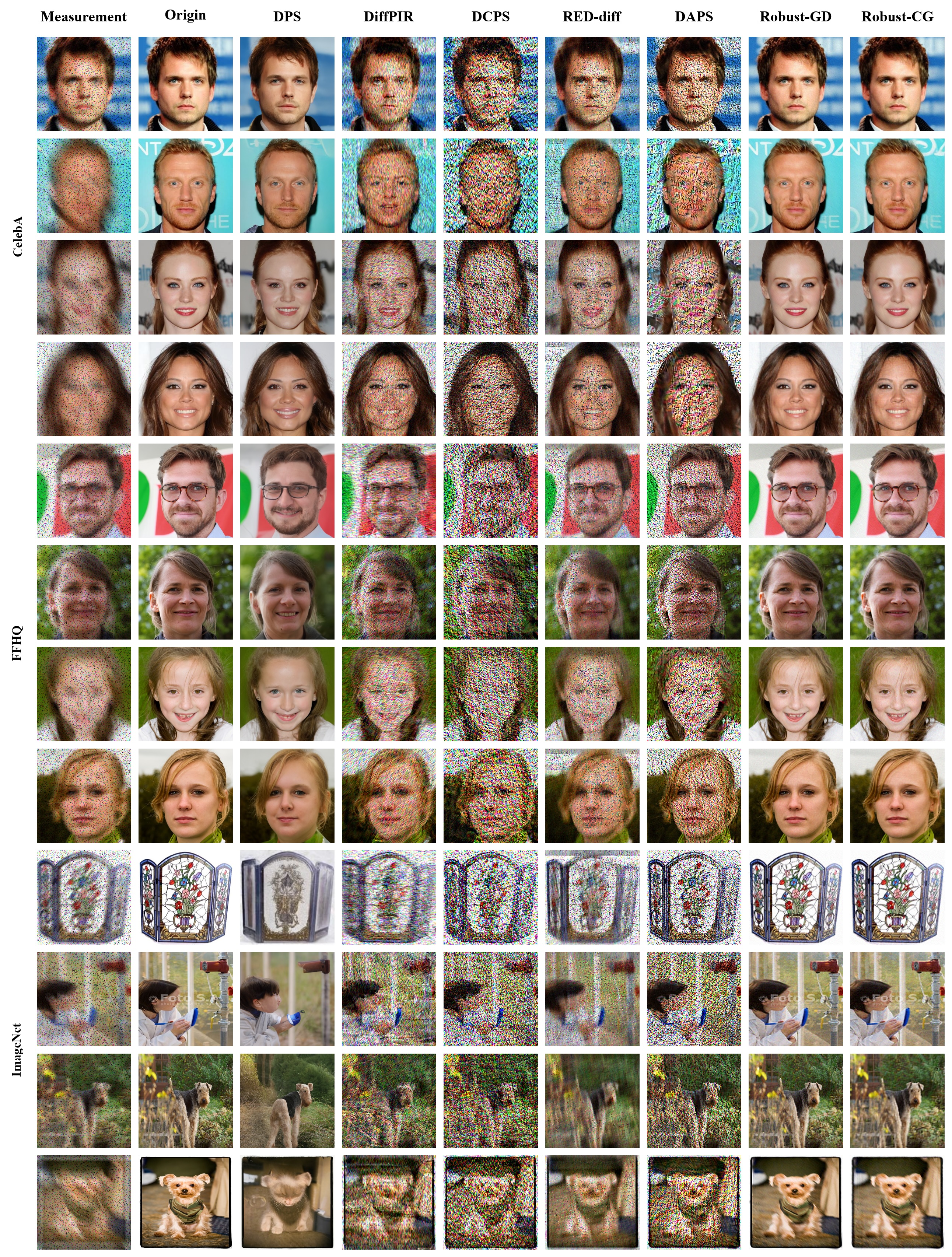}
    \caption{Visualization of the experimental results for the motion deblurring task with a contamination factor of $\rho = 0.10$ and a Gaussian noise level of $\sigma = 0.05$.}
\end{figure}

\begin{figure}[htbp]
    \centering
    \includegraphics[width=0.95\linewidth]{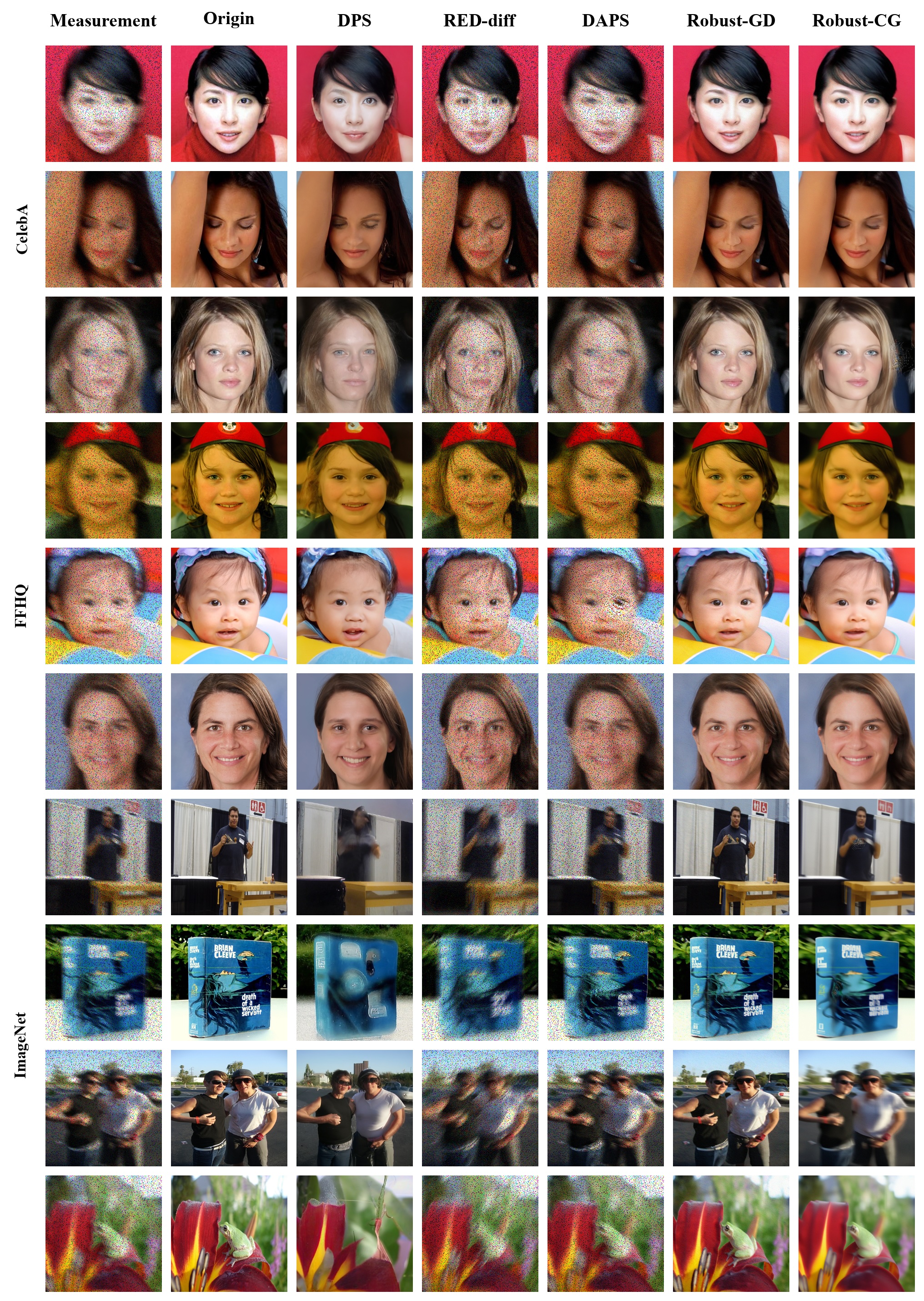}
    \caption{Visualization of the experimental results for the nonlinear deblurring task with a contamination factor of $\rho = 0.10$ and a Gaussian noise level of $\sigma = 0.05$.}
\end{figure}

\end{document}